\documentclass[12pt]{article} % Sets the default font size to 12pt

\usepackage[utf8]{inputenc}    % For Unicode input (necessary for pdfLaTeX)
\usepackage[T1]{fontenc}       % Use T1 encoding to improve font rendering

\usepackage{lmodern}           % Latin Modern fonts (default). 
\usepackage{booktabs}

\usepackage{colortbl}

\usepackage[utf8]{inputenc}

\usepackage{amsfonts}
\usepackage{amssymb}
\usepackage{hyperref}

\usepackage{array}
\usepackage{longtable}
\usepackage{siunitx}
\usepackage{chemformula}

\usepackage{multirow}

\usepackage{setspace}          % Allows control over line spacing
\usepackage[margin=1.00in]{geometry} % Set page margins
\usepackage{enumitem}
\usepackage{authblk}           % Author affiliations etc.
\usepackage{graphicx}          % For including figures
\graphicspath{{./figures/}}    % Figure directory path
\usepackage{subcaption}        % Sub-figures and sub-captions
\usepackage{amsmath}           % Mathematics
\usepackage{lineno}            % Adds line numbers to the document
\usepackage{float}             % Improved floating figures/tables control
\usepackage{hyperref}          % Hyperlinks in the document
\usepackage{listings}          % Code listings
\usepackage{pdflscape}         % Table format
\usepackage[percent]{overpic}  % caption format for (a) and (b) for ex   
\usepackage{adjustbox}
\usepackage{algorithm}
\usepackage{algpseudocode}
\usepackage{rotating}
\usepackage{float}
\usepackage{cleveref}
\usepackage{xcolor}
\definecolor{llmcolor}{RGB}{51, 102, 153}
\definecolor{processcolor}{RGB}{34, 139, 34}
\definecolor{commentcolor}{RGB}{128, 128, 128}

\algrenewcommand\algorithmicrequire{\textbf{Input:}}
\algrenewcommand\algorithmicensure{\textbf{Output:}}
\algrenewcommand\textproc{\textsc}

\usepackage{booktabs}

\definecolor{bgcolor}{rgb}{0.95,0.95,0.95}
\definecolor{commentgray}{rgb}{0.3,0.5,0.3}
\definecolor{keywordblue}{rgb}{0.1,0.1,0.8}
\definecolor{stringred}{rgb}{0.6,0.1,0.1}

\usepackage[style=chem-acs,
            articletitle=true,
            sorting=none,
            maxbibnames=99,
            minbibnames=99,
            maxcitenames=3,
            mincitenames=1,
            uniquelist=false,
            doi=true,
            url=false]{biblatex}
\AtBeginDocument{\setstretch{1.25}}

\title{MolCryst-MLIPs: A Machine-Learned Interatomic Potentials Database for Molecular Crystals}

\author[1]{Adam Lahouari*}
\author[4]{Shen Ai}
\author[1]{Jihye Han}
\author[1]{Jillian Hoffstadt}
\author[1,2,4]{Philipp H{\"o}llmer}
\author[1]{Charlotte Infante}
\author[8]{Pulkita Jain}
\author[1]{Sangram Kadam}
\author[1,4]{Maya M. Martirossyan}
\author[4]{Amara McCune}
\author[2]{Hypatia Newton}
\author[8]{Shlok J. Paul}
\author[1]{Willmor Pena}
\author[4,7]{Jonathan Raghoonanan}
\author[4]{Sumon Sahu}
\author[1]{Oliver Tan}
\author[4]{Andrea Vergara}
\author[3]{Jutta Rogal}
\author[1,2,4,5,6]{Mark E. Tuckerman}

\affil[1]{Department of Chemistry, New York University, 
New York, NY 10003, USA}
\affil[2]{Simons Center for Computational Physical Chemistry, 
New York University, New York, NY 10003, USA}
\affil[3]{Initiative for Computational Catalysis, Flatiron Institute, 
New York, NY 10010, USA}
\affil[4]{Department of Physics, New York University, 
New York, NY 10003, USA}
\affil[5]{Courant Institute of Mathematical Sciences, New York University, 
New York, NY 10012, USA}
\affil[6]{NYU-ECNU Center for Computational Chemistry, 
Shanghai 200062, China}
\affil[7]{New York University Shanghai, 567 West Yangsi Road, Pudong, Shanghai 200126, China}
\affil[8]{Department of Chemical and Biomolecular Engineering, Tandon School of Engineering, New York University, Brooklyn, New York 11201, United States} 
\affil[*]{\textit{Correspondence:} al9500@nyu.edu}

\date{\today}

\begin{document}

\maketitle

\begin{abstract}
We present an open Molecular Crystal (MC) database of Machine-Learned Interatomic Potentials (MLIP) called MolCryst-MLIPs. The first release comprises fine-tuned MACE models for nine molecular crystal systems---Benzamide, Benzoic acid, Coumarin, Durene, Isonicotinamide, Nicotinic acid , Niacinamide, Pyrazinamide, and Resorcinol---developed using the Automated Machine Learning Pipeline (AMLP), which streamlines the entire MLIP development workflow, from reference data generation to model training and validation, into a reproducible and user-friendly pipeline. Models are fine-tuned from the \texttt{MACE-MH-1} foundation model (\texttt{omol} head), yielding a mean energy MAE of 0.141 kJ $\cdot$ mol$^{-1}$$\cdot$ atom$^{-1}$ and a mean force MAE of 0.648 kJ $\cdot$ mol$^{-1}$$\cdot$ \AA$^{-1}$ across all systems.
Benchmarked against three state-of-the-art foundation models on the DFT-labelled polymorph set, only the fine-tuned models resolve the polymorphic energy landscape.
Dynamical stability and structural integrity, as assessed through energy conservation, $P_2$ orientational order parameters, and radial distribution functions, are evaluated using molecular dynamics simulations. The released models and datasets constitute a growing open database of validated MLIPs, ready for production MD simulations of molecular crystal polymorphism across the polymorphic landscape of each target compound under different thermodynamic conditions. 

%The released models and datasets form the foundation of a growing open database of validated MLIPs ready-to-use MLIPs for molecular crystal polymorphism.
\end{abstract}

\section{Introduction}

Polymorphism---the ability of a single chemical compound to crystallize into multiple distinct structures---is a phenomenon of central importance 
in molecular crystals (MC), with far-reaching implications for pharmaceutical development, where polymorphs of the same compound can differ markedly in 
solubility, melting point, and bioavailability.\cite{md-poly1,polymorphisme_md-qc,review_schur,md-poly2} 
Molecular crystals are held together by non-covalent interactions, including $\pi$--$\pi$ stacking, C-H$\cdots\pi$ contacts, and 
hydrogen bonding, which collectively govern the packing arrangement adopted by the crystal. Accurately modelling polymorphism is inherently challenging 
because crystal stability is primarily governed by these intermolecular non-covalent forces, which must be described with sufficient accuracy to 
resolve the subtle interplay between intra- and intermolecular interactions that ultimately determines the relative stability of competing crystal forms. 
Lattice energy differences between polymorphs are typically only a few kJ$\cdot$ mol$^{-1}$, and enantiotropic systems may further exhibit 
stability crossovers as a function of temperature,\cite{freeExpEnergy,freeExpEnergy_rightform} making polymorph 
ranking a particularly sensitive test of potential accuracy. Classical force fields, while computationally efficient, often lack the resolution required 
to discriminate between such energetically close forms or to reliably capture dynamics across varying thermodynamic conditions. High-fidelity structural 
and energetic data can in principle be obtained from quantum mechanical (QM) methods such as density functional theory (DFT); however, the high 
computational cost of DFT---particularly with the tight convergence criteria required for reliable polymorph ranking---restricts its use to small system 
sizes and short timescales in \textit{ab initio} molecular dynamics (AIMD).

Machine-learned interatomic potentials (MLIPs) have emerged as a powerful alternative that bridges this gap: by learning from QM reference data, they achieve accuracy comparable to DFT while enabling molecular dynamics (MD) simulations at a fraction of the computational cost. Three critical choices govern MLIP development: the choice of model 
architecture, the composition and coverage of the training dataset, and whether to train a model from scratch or to fine-tune a pre-existing 
foundation model. These choices are deeply interconnected: the architecture determines both the data requirements and the accuracy ceiling of the 
potential, while the amount and diversity of training data needed depends in turn on whether one trains from scratch or leverages an existing 
foundation model. Training from scratch provides full control over the potential energy surface (PES) but demands large, carefully curated datasets 
and significant computational effort. Fine-tuning leverages representations already learned by a foundation model trained on broad chemical spaces, 
substantially reducing data requirements and training cost while retaining high accuracy for the target system.\cite{fine-tuning-juttafriend} A growing 
number of such foundation models now exist, each targeting different chemical domains and applications.
%
%Two critical choices govern MLIP development: the composition and coverage of the training dataset, and whether to train a model from scratch or to fine-tune a pre-existing foundation model. Training from scratch provides full control over the potential energy surface (PES) but demands large, carefully curated datasets and significant computational effort. Fine-tuning leverages representations already learned by a foundation model trained on broad chemical spaces, substantially reducing data requirements and training cost while retaining high accuracy for the target system.\cite{fine-tuning-juttafriend} A growing number of such foundation models now exist, each targeting different chemical domains and applications. 
Among the most recent, UMA\cite{uma} and Orb\cite{orb25} aim at broad coverage spanning both molecular and materials systems, while the MACE Multi-Head Foundation Model (\texttt{MACE-MH-1})\cite{foundation-mh1} with its \texttt{omol} head---trained on the OMOL dataset of molecular, organic, and organometallic systems\cite{omol}---represents the natural choice for molecular crystal applications, as discussed below.

While MLIPs have demonstrated remarkable success across atomistic and molecular systems, molecular crystals present a distinct challenge: accurate modelling requires simultaneously capturing strong intramolecular interactions and the subtle non-covalent forces governing polymorph stability. Most existing foundation models were trained predominantly on gas-phase molecules or inorganic solids, leaving the periodic crystal environment underrepresented.\cite{omc25} This gap is now being actively addressed: the OMC25 dataset\cite{omc25} provides over 27 million DFT-labelled molecular crystal structures, UMA\cite{uma} incorporates a dedicated molecular crystal training subset derived from it, and \texttt{MACE-MH-1}\cite{foundation-mh1} currently achieves the lowest MAE on the X23 molecular crystal benchmark among available foundation models via cross-domain fine-tuning. These developments signal a turning point for MLIP applicability to molecular crystals. A complementary strategy is to train a domain-specific foundation model directly on organic crystal data: Hafizi \textit{et al.}\cite{hafizi-csp} recently trained MLIPs on crystal structure prediction (CSP) landscapes of the organic molecular solid state, showing that a general model can reproduce DFT energy orderings across large, computationally generated landscapes. The task addressed there, recovering global energetic trends over very large sets of hypothetical structures is, however, distinct from the one considered here, namely resolving the sub-kJ~mol$^{-1}$ ranking of the small number of experimentally realised polymorphs of a given compound; and the same work reports substantial errors for out-of-the-box universal models on organic crystal environments.

Despite these advances, foundation models alone may not reach the accuracy required for demanding tasks such as polymorph ranking or free energy calculations. For instance, in a recent study of the acridine polymorph family, the MACE-MPA foundation model failed to reproduce the experimentally established stability ordering, whereas a system-specifically fine-tuned model recovered the correct ranking.\cite{AMLP} Fine-tuning on system-specific, high-quality DFT data provides a practical route to close this gap, improving both accuracy and transferability across the polymorphic landscape of a target compound.\cite{fine-tuning-juttafriend} However, assembling the necessary training data, selecting appropriate hyperparameters, and validating dynamical stability across multiple polymorphs remains an intensive process that typically requires significant expert knowledge---representing a practical barrier for the broader community. Here, we address this challenge by leveraging the Automated Machine Learning Pipeline (AMLP)\cite{AMLP,amlp_github} to systematically fine-tune \texttt{MACE-MH-1} for nine highly polymorphic molecular crystal systems, demonstrating that the full workflow---from dataset generation to model validation---can be executed in a reproducible and accessible manner.

In this work, we introduce MolCryst-MLIPs\cite{git-MolCryst-MLIPs,huggingface-MolCryst-MLIPs}, an open database of validated, ready-to-use MLIPs for molecular crystals. In contrast to general-purpose foundation models and large training datasets such as OMC25, each model in the database is fine-tuned through the automated active-learning loop of AMLP and explicitly validated for its target compound---including verification of polymorph stability ordering and energy-conserving dynamics---such that it can be deployed directly in downstream free-energy and finite-temperature simulations. The first release comprises fine-tuned \texttt{MACE-MH-1 (omol)} models for nine systems: Benzoic acid\cite{benzac01,benzac02}, Benzamide\cite{bzamid01,bzamid02}, Coumarin\cite{shtukenberg2017powder}, Durene\cite{durene_poly01,durene02,durene03_tkatchenko}, Isonicotinamide\cite{ehowhi01,ehowhi02}, Nicotinic acid\cite{ehowhi_poly2}, Niacinamide\cite{ehowhi_poly2,nicoam_poly3,nicoam_alexandbart_exp}, Pyrazinamide\cite{pyr_poly3,pyrizin02,pyrizin03}, and Resorcinol\cite{resorcinol01,resorcinol02}. These systems were chosen for this first release based on the feasibility of generating high-quality DFT reference data within the active-learning workflow---molecules of moderate size and conformational complexity with tractable unit cells---while exhibiting a sufficient number of experimentally characterized polymorphs to make the energy-ranking task meaningful; these criteria naturally favor small aromatic systems, and the present set should be considered the starting point of a database intended to grow steadily toward larger, more flexible, and chemically more diverse compounds. Application to a chemically distinct compound requires fine-tuning a new model, for which the AMLP workflow and the released datasets provide the protocol and the infrastructure. All models and datasets were developed within the AMLP framework, which automates dataset generation, model training, and validation through a multi-agent architecture, making the development of high-fidelity MLIPs for molecular crystals accessible to a broad research community. By demonstrating consistent accuracy in energies and forces alongside dynamical stability across nine polymorphic systems, this work establishes both AMLP as a practical tool for accelerating MLIP development and MolCryst-MLIPs as a growing community resource for large-scale MD simulations of molecular crystal polymorphism.

\section{Materials and Methods}

All workflows presented in this work---from reference data generation to model training and validation---were orchestrated using AMLP.\cite{AMLP} The pipeline handles DFT input generation, active learning, dataset curation, MACE training configuration, and post-training validation, ensuring reproducibility across diverse molecular systems while significantly reducing the manual effort required for MLIP development.

%\subsection{Reference Data Generation}

%Experimental crystal structures (\texttt{.cif} files) for all polymorphs were obtained from the Cambridge Structural Database (CSD). AMLP automatically parsed these structures and generated input files for geometry optimizations and AIMD simulations using the Vienna Ab Initio Simulation Package (VASP).\cite{vasp1,vasp2,vasp3,vasp4} The Perdew--Burke--Ernzerhof (PBE) exchange-correlation functional was employed with Grimme's D4 dispersion correction.\cite{PBE,grimm-d4} Brillouin-zone sampling was performed using a $\Gamma$-centered Monkhorst--Pack mesh with a density automatically adjusted to the dimensions of each unit cell.\cite{monkhorst}

Experimental crystal structures (\texttt{.cif} files) for all polymorphs were obtained from the Cambridge Structural Database (CSD).\cite{csd_database} AMLP automatically parsed these structures and generated input files for cell or geometry optimizations and AIMD simulations using the Vienna Ab Initio Simulation Package (VASP).\cite{vasp1,vasp2,vasp3,vasp4} The Perdew--Burke--Ernzerhof (PBE) exchange-correlation functional~\cite{PBE} was employed with Grimme's D4 dispersion correction, providing an accurate description of the long-range van der Waals interactions critical for molecular crystal energetics.\cite{grimm-d4} Brillouin-zone sampling was performed using a $\Gamma$-centered Monkhorst--Pack mesh with a density automatically adjusted to the dimensions of each unit cell.\cite{monkhorst} Cell optimizations were conducted with a plane-wave cutoff energy of 650~eV and a tight electronic convergence criterion (EDIFF = $10^{-7}$~eV). PBE+D4 was selected as a pragmatic compromise between accuracy and cost. The AL workflow requires thousands of single-point evaluations on cells of several hundred atoms, in addition to cell relaxations under tight convergence criteria; hybrid functionals such as PBE0+D3/D4 are computationally prohibitive at this scale, and meta-GGA schemes such as SCAN+rVV10, while accurate for many condensed-phase systems, are substantially more demanding and require tighter numerical settings for robust convergence. D4 was chosen over earlier pairwise schemes as a modern, environment-dependent dispersion model that improves the description of long-range correlation at negligible additional cost.\cite{grimm-d4} The accuracy of any MLIP is nevertheless bounded by that of its reference data: dispersion-corrected GGA functionals reproduce the absolute lattice energies of the X23 benchmark with mean absolute errors of typically 4--6~kJ~mol$^{-1}$ per molecule,\cite{d4-periodic} although the relative energies of polymorphs of the same compound benefit from substantial error cancellation, since the molecular species and bonding environment are common to all forms. Residual exchange--correlation error may perturb nonetheless the relative ordering of near-degenerate forms, and the rankings reported below should be understood as reproducing the PBE+D4 reference rather than the exact experimental hierarchy. AMLP is, however, agnostic to the underlying electronic-structure method, and because the complete DFT datasets are released alongside the models, they can be regenerated or benchmarked against improved functionals as these become affordable. AIMD simulations were performed at temperatures ranging from 25~K to 700~K to maximize coverage of the potential energy surface. To further reinforce the models in undersampled regions, an active learning loop was performed: structures that led to simulation failures during NVT runs — using a Berendsen thermostat — were identified, recomputed at the DFT level, and added to the training set. The combined dataset of DFT-optimized structures, AIMD trajectories, and actively learned configurations provides the diverse configurational space essential for training robust and transferable interatomic potentials.

%Cell optimizations were conducted with a plane-wave cutoff energy of 750~eV and a tight electronic convergence criterion (EDIFF = $10^{-7}$~eV) to ensure accurate determination of lattice energies. AIMD simulations were performed at temperatures ranging from 25~K to 700~K to sample off-equilibrium configurations. The combined dataset of DFT-optimized structures (near-equilibrium) and AIMD trajectories (finite-temperature) provides the diverse configurational space essential for training robust and transferable interatomic potentials.

%AMLP here is used to create automatically all the different inputs, automatically extract energies, forces, and structures upon completion to construct training datasets.

%\subsection{MACE Model Training}

%To reduce the computational cost and data requirements for MLIP development, we leverage the MACE Multi-Head Foundation Model (\texttt{mace-mh-1})\cite{foundation-mh1} with its \texttt{omol} head, trained on the OMOL dataset of molecular, organic, and organometallic systems.\cite{omol} This head was selected based on the X23 benchmark, where \texttt{mace-mh-1-omol-1\%} achieved the lowest MAE of 7.41~kJ/mol for formation energies among all available heads for molecular crystals,\cite{foundation-mh1} making it the most suitable starting point for fine-tuning. 

To reduce the computational cost and data requirements for MLIP development, we leverage \texttt{MACE-MH-1}\cite{foundation-mh1} with its \texttt{omol} head, selected based on its lowest MAE of 7.41~kJ/mol on the X23 molecular crystal benchmark among all available heads.\cite{foundation-mh1} Fine-tuning preserves the foundation model architecture, enabling efficient transfer learning with significantly reduced training data.\cite{fine-tuning-juttafriend} Training followed a two-stage protocol combining an initial optimization phase with stochastic weight averaging (SWA), using the Adam optimizer with AMSGrad and early stopping. All models were trained in double precision (\texttt{float64}). Complete hyperparameter configurations are provided in the Supplementary Information (SI).

%\subsection{Model Validation}

To assess the reliability of each MLIP, we designed a systematic validation protocol automated within the AMLP framework, spanning both static and dynamic regimes. Structural fidelity was first evaluated by comparing DFT-optimized and MACE-optimized geometries across all polymorphs in the training set, as well as on unseen polymorphs to probe transferability beyond the training distribution. Moreover, to compare the MLIP-optimized structures directly with experiment, we computed RMSD$_{15}$: the root-mean-square deviation of atomic positions over a cluster of the 15 nearest molecules after optimal rigid-body superposition, following the growing-shell construction introduced by Chisholm and Motherwell\cite{compack} and standard in crystal structure prediction. Because a coordination shell rather than a single unit cell is superimposed, the metric is sensitive to lattice-parameter error as well as to intramolecular and packing deviations. Whole molecules were perceived from the periodic structure using Cordero covalent radii with an additive bonding tolerance of 0.4~\AA, with unwrapping across periodic boundaries; the corresponding cluster was located in the optimized structure by nearest centroid in fractional coordinates, so that a coherent change of cell parameters does not displace the correspondence; atoms were paired within each molecule by species-resolved assignment; and the RMSD was evaluated by a single Kabsch superposition of the complete matched cluster. The implementation is released with the database.

Dynamical robustness was then examined through microcanonical (NVE) simulations performed on the most stable polymorph of each system to monitor energy conservation as a direct measure of PES quality. Canonical (NVT) simulations using the Berendsen thermostat were then conducted as a first-pass screening step within the automated workflow, from room temperature (RT) up to 600~K, to identify models unable to sustain stable dynamics across the polymorphic landscape of each system. Since the experimental melting temperatures of the nine compounds span 344--510~K,\cite{nist-webbook} our simulations deliberately exceed the melting point of all systems, allowing us to probe not only the thermally stable regime but also the onset of structural disordering. As melting temperatures can further vary between polymorphs of the same compound, some forms are expected to lose orientational or structural order before others. Throughout all simulations, radial distribution functions 
(RDFs) and the orientational order parameter $P_2$ (see Eq.\ (\ref{eq:P2}) below) were monitored to characterize structural integrity and molecular orientational order within each polymorph.

Because a fixed cell cannot deform, and may therefore mask deficiencies that would otherwise appear as unphysical lattice relaxation, stability was additionally assessed under constant-pressure (NPT) conditions with a fully flexible cell, using a Parrinello--Rahman barostat coupled to a Nos\'e--Hoover chain thermostat at 300~K and ambient pressure. A timestep of 1~fs was used, with 10~ps of equilibration followed by 100~ps of production dynamics per polymorph (thermostat and barostat relaxation times of 200~fs and 2~ps, respectively). Three systems---benzoic acid, nicotinic acid and resorcinol---were examined, chosen as representative of the distinct hydrogen-bonding motifs spanned by the database: carboxylic-acid dimers, a hydrogen-bonded zwitterionic network, and a polyhydroxy three-dimensional network, respectively.

%As melting temperatures can further vary between polymorphs of the same compound, some forms are expected to lose orientational or structural order before others---a physically correct behaviour that our validation protocol is designed to capture. Throughout all simulations, radial distribution functions (RDFs) and the orientational order parameter $P_2$ were monitored to characterize structural integrity and molecular orientational order within each polymorph.

%All post-processing analyses, including energy drift calculations and RDF extraction, were performed using AMLP's built-in analysis modules.

\section{Result and discussion}

\subsection{Reference Data Generation}

The starting geometries for each compound were taken from experimental crystal structures (\texttt{.cif} files) retrieved from the CSD. 
%The number of polymorphs per system is summarized in Table~\ref{tab:volume_contraction}. 
The dataset was restricted to polymorphs containing at most 8 molecules in the unit cell ($Z \leq 8$), as the computational cost of DFT cell relaxation becomes prohibitive for larger systems. Polymorphs with $Z > 8$ were excluded from the training set but are nonetheless considered in the transferability analysis in Section~\ref{sec:mace_finetuning}, where we assess whether the trained MLIPs can serve as efficient pre-screening tools for geometry/cell optimization prior to costly DFT calculations. Using the AMLP framework, input files for all 65 cell relaxations were generated automatically with a consistent set of DFT parameters. To verify the physical validity of the resulting structures, two metrics were examined: the range of lattice energies across polymorphs, which is not expected to exceed $\sim$10~kJ~mol$^{-1}$, defined as
\begin{equation}
E_{\text{lattice}} = \frac{E_{\text{crystal}}}{Z} - E_{\text{gas}},
\end{equation}
where $Z$ is the number of molecules per unit cell, $E_{\text{crystal}}$ is the total energy of the crystal, and $E_{\text{gas}}$ is the energy of an isolated molecule in the gas phase in its global minimum configuration; and the structural deviation between the DFT-relaxed and experimental unit cells. The volumetric contraction is quantified as
\begin{equation}
    \Delta V = \frac{V_{\text{DFT}} - V_{\text{exp}}}{V_{\text{exp}}} , %\times 100,
\end{equation}
where $V_{\text{DFT}} = \det({{\rm h}}_{\text{DFT}})$ and $V_{\text{exp}} =
\det({\rm h}_{\text{exp}})$, with ${\rm h}_{\text{DFT}}$ and ${\rm h}_{\text{exp}}$
being the $3 \times 3$ matrices whose columns are the lattice vectors of the DFT-optimized and experimental cells, respectively. The full cell deviation, sensitive to both volumetric contraction and angular distortions, is captured by the strain tensor norm
\begin{equation}
    \|{\rm F} - {\rm I}\|_F \equiv \left\|{\rm h}_{\text{exp}}^{-1}{\rm h}_{\text{DFT}}
    - {\rm I}\right\|_F,
\end{equation}
where ${\rm F} \equiv {\rm h}^{-1}_{\rm exp}{\rm h}_{\rm DFT}$, ${\rm I}$ is the $3 \times 3$ identity matrix and $\|\cdot\|_F$ denotes the Frobenius matrix norm. Since DFT calculations are performed at 0~K while experimental structures are typically determined at room temperature ($\sim$298~K), a uniform contraction of the DFT cell is physically expected due to the absence of thermal expansion, corresponding to $\Delta V < 0$ and $\|{\rm F} - {\rm I}\|_F > 0$.

For the nine systems considered, the relative lattice energies between polymorphs span from as low as 0.09~kJ~mol$^{-1}$ for Durene to 4.64~kJ~mol$^{-1}$ for Resorcinol, with a mean of 2.19~kJ~mol$^{-1}$ (Table~\ref{tab:volume_contraction}), well within the physically expected range of below 10~kJ~mol$^{-1}$. DFT optimizations consistently yielded volumetric contractions ranging from $-3.0\%$ to $-6.4\%$ across all compounds (mean $-4.7\%$), with strain tensor norms $\|{\rm F} - {\rm I}\|_F$ ranging from 0.025 to 0.081 (Table~\ref{tab:volume_contraction}), in line with the expected behavior for 0~K calculations against experimental structures. Together, these metrics confirm the physical validity of the DFT-optimized dataset and its suitability as a reference for MLIP training.

\begin{table}[htbp]
\centering
\small
\caption{Mean volumetric contraction $\overline{\Delta V}$ (\%), mean strain tensor norm $\overline{\|{\rm F}-{\rm I}\|}_F$, and maximum relative lattice energy
$\Delta E_{\text{latt.}}^{\text{max}}$ for each compound.}
\label{tab:volume_contraction}
\begin{tabular}{@{}lccc>{\raggedright\arraybackslash}p{3.8cm}@{}}
\toprule
\textbf{Compound}
    & \textbf{$\overline{\Delta V}$ (\%)}
    & \textbf{$\overline{\|{\rm F}-{\rm I}}\|_F$}
    & \textbf{$\Delta E_{\text{latt.}}^{\text{max}}$ (kJ/mol)}
    & \textbf{Experimental References} \\
\midrule
Resorcinol      & $-4.1$ & $0.025$ & 4.64 & \cite{resora_poly01,resora_poly02} \\
Pyrazinamide    & $-3.2$ & $0.028$ & 4.35 & \cite{pyr_poly6,pyr_poly4,pyr_poly5,pyr_poly3,pyr_poly18} \\
Niacinamide     & $-3.0$ & $0.051$ & 3.02 & \cite{nicoam_poly1,nicoam_poly2,nicoam_poly3} \\
Nicotinic acid  & $-5.6$ & $0.046$ & 1.70 & \cite{nicoac_poly1,nicoac_poly2,nicoac_poly3} \\
Isonicotinamide & $-4.1$ & $0.032$ & 1.41 & \cite{ehowhi_poly1,ehowhi_poly2,ehowhi_poly3,ehowhi_poly4} \\
Durene          & $-6.4$ & $0.050$ & 0.09 & \cite{durene_poly01} \\
Coumarin        & $-4.7$ & $0.034$ & 1.93 & \cite{shtukenberg2017powder,zhang2021structural} \\
Benzoic acid    & $-6.3$ & $0.081$ & 1.20 & \cite{benzac_poly01,benzac_poly02} \\
Benzamide       & $-5.0$ & $0.075$ & 1.38 & \cite{bzamid_poly01,bzamid_poly02,bzamid_poly03,bzamid_poly04} \\
\midrule
\textbf{Mean}   & \textbf{$-4.7$} & \textbf{$0.047$} & \textbf{2.19} & \\
\bottomrule
\end{tabular}
\end{table}

To complement the near-equilibrium DFT-optimized structures, extensive off-equilibrium data were generated using AIMD, producing thousands of diverse configurations per polymorph. AMLP was employed to systematically generate
%AIMD inputs across four temperatures (25~K, 300~K, 400~K, and 500~K) for all 65 polymorphs, yielding a total of 260 trajectories. To further reinforce the models in regions of the potential energy surface that remained insufficiently sampled after the initial AIMD runs, an active learning loop was performed: configurations with high model uncertainty were identified, recomputed at the DFT level, and added to the training set.
AIMD inputs across four temperatures (25~K, 300~K, 400~K, and 500~K) for all 65 polymorphs, yielding a total of 260 trajectories, supplemented by an active learning loop to reinforce undersampled regions of the PES. 
%Figure~\ref{fig:energy_force_density} shows the joint distribution of energy per atom versus mean atomic force across all training configurations for each compound, presented as 2D density histograms. The strong linear correlation observed in all nine systems (Pearson $r = 0.907$--$0.949$) reflects the expected physical relationship between structural distortion and potential energy: configurations with larger mean forces correspond to higher-energy, off-equilibrium geometries, while low-force configurations cluster near the DFT-optimized minima. 
The expected physical relationship between structural distortion and potential energy---whereby off-equilibrium configurations with larger atomic forces correspond to higher energies, while low-force configurations cluster near the DFT-optimized minima---should manifest as a strong linear correlation between energies and forces across the training set. This is indeed observed for all nine systems, with Pearson $r = 0.907$--$0.949$ (see Figure~\ref{fig:energy_force_density} in the SI).
The well-populated diagonal in each panel confirms that the training datasets provide balanced coverage across the full energy--force spectrum, from near-equilibrium structures to high-energy configurations sampled during AIMD and active learning. Dataset sizes vary directly with the number of polymorphs per system, ranging from approximately 5,000 configurations for Isonicotinamide ($r = 0.941$) to over 30,000 for Pyrazinamide ($r = 0.907$). In total, 113,953 structures were generated across all nine compounds, combining DFT-optimized geometries, AIMD trajectories, and actively learned configurations, providing the 
MACE models with comprehensive sampling of the configurational space relevant to molecular crystal polymorphism.

%\begin{figure}[H]
%    \centering
%    \includegraphics[width=0.85\linewidth]{plot-Main/force_distribution_all_crystals_multipanel.png}
%    \caption{Distribution of maximum atomic forces across AIMD training datasets for all nine compounds. Each panel shows the histogram of maximum forces per configuration, with dataset size (N) and mean force magnitude indicated.}
%    \label{fig:distribution-forces}
%\end{figure}

The combined dataset was systematically filtered and curated using the AMLP data curation tools.\cite{AMLP} Three filtering criteria were applied: (i) a DBSCAN-based cluster analysis\cite{dbscan} in the joint energy--force space, which identifies and removes configurations that do not belong to the main data distribution, such as structures arising from DFT convergence failures or anomalous geometries; (ii) a force magnitude cutoff of 10.0~eV/\AA\ to remove structures with unphysical atomic forces; and (iii) an energy-per-atom threshold of 0.2~eV/atom ($\approx$19.3~kJ~mol$^{-1}$~atom$^{-1}$) relative to the per-atom reference energies $E_0$ (see Table~\ref{tab:E0s} in the SI), computed as $E_{\text{atom}} = (E_{\text{crystal}} - \sum_i E_{0,i}) / N_{\text{atoms}}$, where the sum runs over all atoms in the unit cell according to their chemical species, to exclude structures with anomalously high energies that could destabilize model training. These filtering steps removed outlier configurations while preserving the chemical diversity necessary for robust transferability across polymorphs and temperatures. The curated datasets were finally split into training and validation sets with a ratio of 85/15.

\subsection{MACE Model Fine-Tuning and Evaluation}
\label{sec:mace_finetuning}

Fine-tuning the \texttt{MACE-MH-1} foundation model yielded highly accurate potentials for all nine systems, as summarized in Table~\ref{tab:training_mae}. Averaged across all compounds, the energy MAE is
%The mean energy MAE across all compounds
0.141~kJ~mol$^{-1}$~atom$^{-1}$, while the forces MAE is 0.648 kJ~mol$^{-1}$~\AA$^{-1}$, both well within the range required for reliable polymorph ranking and stable MD simulations.  All models were trained on a single NVIDIA A100 GPU, with wall-clock training times ranging from approximately 22~h for the smallest datasets to over 100~h for the largest scaling linearly with the number of training configurations
(see Table~\ref{tab:training_times} and Figure \ref{fig:training_time} in the SI). 

\begin{table}[htbp]
\centering
\caption{Training performance of MACE models. Validation set metrics are 
reported for all compounds.}
\label{tab:training_mae}
\begin{tabular}{@{}lcc@{}}
\toprule
\textbf{Compound} & \textbf{Energy MAE} & \textbf{Force MAE} \\
 & \textbf{(kJ~mol$^{-1}$~atom$^{-1}$)} & \textbf{(kJ~mol$^{-1}$~\AA$^{-1}$)} \\
\midrule
Benzoic acid    & 0.128 & 0.762  \\
Benzamide       & 0.069 & 0.848  \\
Coumarin        & 0.161 & 0.414  \\
Durene          & 0.159 & 0.501  \\
Isonicotinamide & 0.184 & 1.043  \\
Nicotinic acid  & 0.116 & 0.562  \\
Niacinamide     & 0.146 & 0.695  \\
Pyrazinamide    & 0.158 & 0.627  \\
Resorcinol      & 0.151 & 0.377  \\
\midrule
\textbf{Mean}   & \textbf{0.141} & \textbf{0.648} \\
\bottomrule
\end{tabular}
\end{table}

\begin{figure}[H]
    \centering
    \includegraphics[width=0.85\linewidth, keepaspectratio]{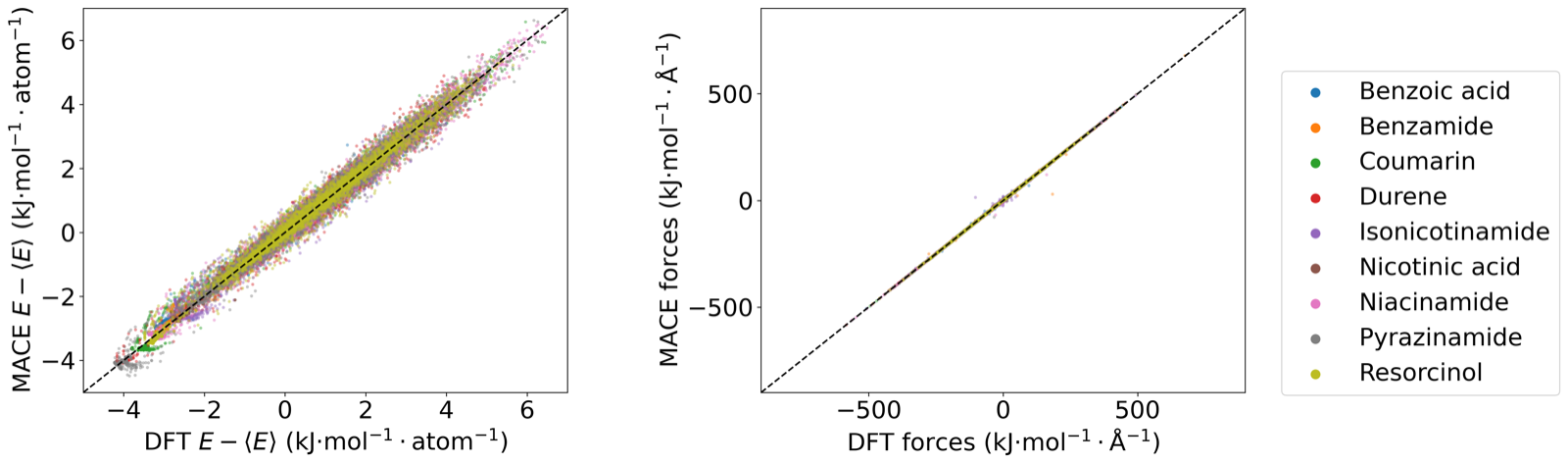}
    \caption{DFT vs.\ MACE correlation plots for (a) energies and (b) forces across all nine fine-tuned models.}
    \label{fig:forces_correlations_all}
\end{figure}

Table~\ref{tab:training_mae} and Figure~\ref{fig:forces_correlations_all} demonstrate consistently low MAE values for both energies and forces across all systems. In the energy correlation plots, $E - \langle E \rangle$ denotes the mean-centred energy per atom, where $\langle E \rangle$ is the mean energy per atom computed independently for each dataset, allowing a direct comparison of the PES shape across systems with different absolute reference energies. Individual per-system correlation plots are provided in Figures~\ref{fig:forces_correlations_01} and~\ref{fig:forces_correlations_02} of the Supporting Information.

\begin{figure}[H]
    \centering
    \includegraphics[width=0.8\linewidth]{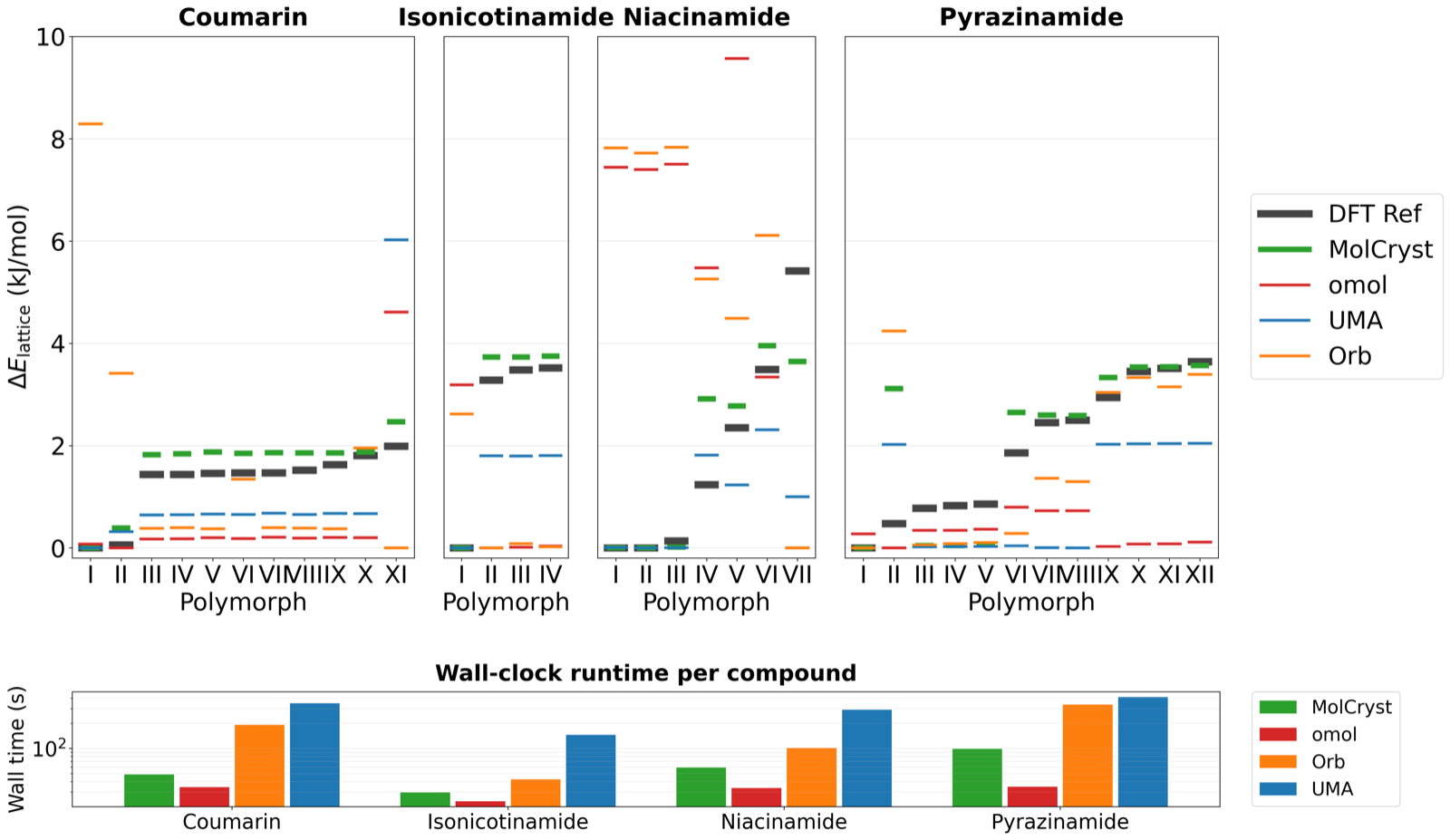}
    \caption{
    Relative lattice energies ($\Delta E_{\mathrm{latt}}$, kJ~mol$^{-1}$) for four molecular crystal systems: Coumarin, Isonicotinamide, Niacinamide, and Pyrazinamide. Five methods are shown: the DFT reference (black bar), the MolCryst-MLIPs fine-tuned MACE model (green colored bar), the MACE-MH-1 foundation model (\texttt{omol} head, red bar), UMA (\texttt{omc} head, solid blue bar), and orb (\texttt{d3-v2)} head, solid orange bar). The reference zero is set independently for each method to the most stable polymorph within the common evaluation set. Bottom: wall-clock runtime per compound for the fine-tuned MolCryst-MLIPs, \texttt{omol}, orb, and UMA models.
    }
    \label{fig:elatt_mace_omol_dft}
\end{figure}

In order to assess whether fine-tuned models can generate a more accurate polymorphic energy landscape than the original foundation model, we compute the relative lattice energy $\Delta E_{\mathrm{latt}}$ with respect to the most stable polymorph. The relative lattice energy for polymorph $i$ is defined as:
\begin{equation}
    %\Delta E_{\mathrm{latt}} = \frac{E_{\mathrm{crystal}}}{Z} - 
    %\min_{j} \frac{E_{\mathrm{crystal},j}}{Z_j}
    \Delta E_{\mathrm{latt},i} = E_{\mathrm{latt},i} - \min_j E_{\mathrm{latt},j}
    \label{eq:delta_elatt}
\end{equation}
where 
%$E_{\mathrm{crystal}}$ is the total cell energy, $Z$ is the number of molecules per unit cell, and 
the minimum is taken over all polymorphs within the common evaluation set. The reference zero is set independently for each method, so that Eq.~\eqref{eq:delta_elatt} measures the relative stability ranking predicted by each MLIP rather than absolute energetic agreement with DFT. Starting from the DFT cell-optimised structures, geometries are fully relaxed with each MLIP model. 
Optimizations were carried out using the \texttt{amlpa.py} module from AMLP, based on the ASE framework\cite{ase-paper} with the LBFGS optimizer and a force convergence criterion of $f_{\text{max}} = 0.001$~eV/\AA. In Figure~\ref{fig:elatt_mace_omol_dft}, polymorphs are labeled with Roman numerals (I, II, \ldots) in order of increasing DFT stability for clarity; the corresponding CSD reference codes are provided in Table~\ref{tab:roman_to_csd} in the SI.

%Figure~\ref{fig:elatt_mace_omol_dft} presents the results for four representative systems. 
The relative lattice energies for four representative systems are shown in Figure~\ref{fig:elatt_mace_omol_dft}.
The MACE-MH-1 foundation model with the \texttt{omol} head consistently fails to discriminate between polymorphs, predicting either a nearly flat energy landscape — as seen for Coumarin and Pyrazinamide — or a qualitatively incorrect stability ordering, as in Niacinamide where the foundation model places the low-energy forms at over 7~kJ~mol$^{-1}$ above the true minimum. This failure is expected: the sub-kJ~mol$^{-1}$ energy differences that govern polymorph stability lie well below the resolution of a model trained on broad chemical space without system-specific data. We further evaluated two additional state-of-the-art foundation models, UMA (using the \texttt{omc} head) and orb (using the \texttt{d3-v2} head), under the identical relaxation and $\Delta E_{\mathrm{latt}}$ protocol. Orb does not reproduce the DFT stability ordering in most of the systems, as shown in Figure~2, yielding a ranking that disagree qualitatively with the DFT reference and closer to that of the \texttt{omol} model, as seen for Isonicotinamide. UMA, on the other hand, recovers the correct ordering for most systems, but like the \texttt{omol} model it collapses several polymorphs within its own MAE, making them not truly distinguishable, limiting its resolution in precisely the regime that governs polymorph stability.

In contrast, the fine-tuned MolCryst-MLIPs models recover the correct DFT stability ranking in all four of these systems. For Coumarin, the fine-tuned model correctly identifies the near-degenerate cluster of polymorphs at $\sim$1.5~kJ~mol$^{-1}$ above the global minimum. For
Niacinamide and Pyrazinamide, the relative ordering across the full polymorph set is in good qualitative agreement with DFT, with energy spreads of the correct magnitude. For Isonicotinamide, the \texttt{omol} foundation model predicts a stability ranking that is inverted relative to DFT, incorrectly stabilising forms~II--IV near zero while placing the true global minimum (form~I) at $\sim$3.2~kJ~mol$^{-1}$. Fine-tuning with the MolCryst-MLIPs dataset fully corrects this inversion, recovering the DFT ranking with form~I as the most stable polymorph and the remaining forms correctly placed at $\sim$3.5~kJ~mol$^{-1}$ above it. Beyond accuracy, the fine-tuned models are also substantially cheaper to evaluate. Under our relaxation protocol, UMA is approximately an order of magnitude more expensive in wall time than the MolCryst-MLIPs models (Figure~\ref{fig:elatt_mace_omol_dft}, bottom), a difference that becomes significant when the potential is deployed in demanding applications such as free-energy calculations.
These results demonstrate that targeted fine-tuning on system-specific DFT data is a prerequisite for reliable polymorph discrimination with MLIPs. The MolCryst-MLIPs fine-tuned models recover a capability that is inaccessible to foundation models, correctly ranking polymorphs in systems where the foundation model predicts a completely flat or inverted energy landscape.

Table~\ref{tab:deltaElatt_summary} aggregates the performance of each model across all nine systems; the complete per-system breakdown, including the penalty of each model on each compound, is provided in Table~\ref{tab:deltaElatt_all9} of the SI. The fine-tuned models attain a mean $\Delta E_{\mathrm{latt}}$ MAE of 0.98~kJ~mol$^{-1}$ and a mean Kendall $\tau$ of 0.40, against 1.21 and 0.15 for the best foundation model (UMA) and negative rank correlations for \texttt{omol} and orb. The most informative measure, however, is the energetic penalty incurred when a model does not select the true minimum: for the fine-tuned models this penalty averages 0.51~kJ~mol$^{-1}$---half that of UMA (1.08~kJ~mol$^{-1}$) and a quarter that of orb---and in five of the nine systems it is exactly zero. Where the models do misrank polymorphs, the selected form lies within $\sim$1~kJ~mol$^{-1}$ of the true minimum for benzamide and nicotinic acid; for durene the entire polymorphic landscape spans only 1.5~kJ~mol$^{-1}$, so that all forms are near-degenerate by construction; and for resorcinol the selected form lies 1.86~kJ~mol$^{-1}$ above the minimum on a landscape spanning 5.64~kJ~mol$^{-1}$, the largest penalty of the fine-tuned set but still well below the qualitative failures discussed next. The foundation models, by contrast, fail qualitatively: for benzoic acid, \texttt{omol}, UMA and orb all incur a penalty of 2.53~kJ~mol$^{-1}$---the full width of the DFT landscape---each therefore placing the least stable polymorph at the bottom of its own energy ordering. Isonicotinamide shows the same pattern, with \texttt{omol} and orb incurring penalties of 3.28~kJ~mol$^{-1}$ on a landscape spanning 3.52~kJ~mol$^{-1}$. Fine-tuning therefore does more than improving the ranking accuracy: it is what separates a landscape that is qualitatively correct from one that is different.

\begin{table}[htbp]
\centering
\small
\caption{Mean over the nine systems of the per-system $\Delta E_\mathrm{latt}$ accuracy and ranking metrics. MAE, RMSE and pairwise MAE in kJ\,mol$^{-1}$ per molecule; $\tau$, $\rho$ and $r$ are mean Kendall, Spearman and Pearson correlations. ``Penalty'' is the mean $\Delta E_\mathrm{latt}$ of the predicted most stable polymorph above the true DFT minimum (zero when the DFT minimum, or a form degenerate with it, is selected). ``Most stable'' counts the systems whose most stable polymorph is identified within 2~kJ\,mol$^{-1}$ of the DFT minimum; the tolerance reflects the near-degeneracy of several forms in the DFT reference itself.}
\label{tab:deltaElatt_summary}
\begin{tabular}{lcccccccc}
\toprule
Model & MAE & RMSE & MAE$_\mathrm{pair}$ & $\tau$ & $\rho$ & $r$ & Penalty & Most stable \\
\midrule
\textbf{MolCryst-MLIPs} & \textbf{0.98} & \textbf{1.22} & \textbf{0.99} & \textbf{0.397} & \textbf{0.460} & \textbf{0.490} & \textbf{0.51} & \textbf{9/9} \\
UMA                     & 1.21 & 1.54 & 1.26 & 0.145 & 0.220 & 0.365 & 1.08 & 6/9 \\
\texttt{omol}           & 1.99 & 2.54 & 2.02 & $-0.044$ & $-0.044$ & $-0.066$ & 1.80 & 5/9 \\
Orb$^{a}$            & 3.10 & 3.86 & 2.58 & $-0.250$ & $-0.309$ & $-0.415$ & 2.01 & 4/7 \\
\bottomrule
\multicolumn{9}{l}{\footnotesize $^{a}$ Durene and benzamide are excluded from the Orb averages. For durene only two polymorphs converged,}\\
\multicolumn{9}{l}{\footnotesize \phantom{$^{a}$} for which rank correlations are undefined; for benzamide, three relaxations failed to converge, oscillating}\\
\multicolumn{9}{l}{\footnotesize \phantom{$^{a}$} at $f_\mathrm{max}\approx0.02$~eV/\AA\ for more than 15{,}000 steps without reaching the $10^{-3}$~eV/\AA\ criterion.}
\end{tabular}
\end{table}

These results also bear on the question of whether a single potential, trained jointly across several molecular crystal systems, could replace the per-compound models released here. The relevant experiment has in effect already been performed by the community, and at a scale far beyond what nine systems could provide: UMA's \texttt{omc} task head is trained on the OMC25 dataset of more than 27 million DFT-labelled molecular crystal structures\cite{omc25,uma} and is therefore a unified molecular crystal model in precisely that sense. Nevertheless, as Table~3 shows, broadening the chemical scope of a potential degrades exactly the sub-kJ~mol$^{-1}$ resolution on which polymorph ranking---and any downstream free-energy estimate---depends. It attains a mean Kendall $\tau$ of only 0.15, and---for benzoic acid---selects the least stable DFT polymorph as its global minimum. A model trained jointly on the nine compounds of the present release would be strictly poorer in chemical coverage than UMA-\texttt{omc} while forfeiting the per-system resolution that Table~\ref{tab:deltaElatt_summary} shows to be decisive. The trade-off between specialization and generalization is therefore not hypothetical here but measured: broadening the chemical scope of a potential degrades exactly the sub-kJ~mol$^{-1}$ resolution on which polymorph ranking---and any downstream free-energy estimate---depends. This does not imply that a general-purpose molecular crystal potential is unattainable; training foundation models directly on organic crystal landscapes\cite{hafizi-csp} is a promising route, but it constitutes a distinct research programme from the validated, per-compound library presented here.

\begin{figure}[H]
    \centering
    \includegraphics[width=0.65\linewidth]{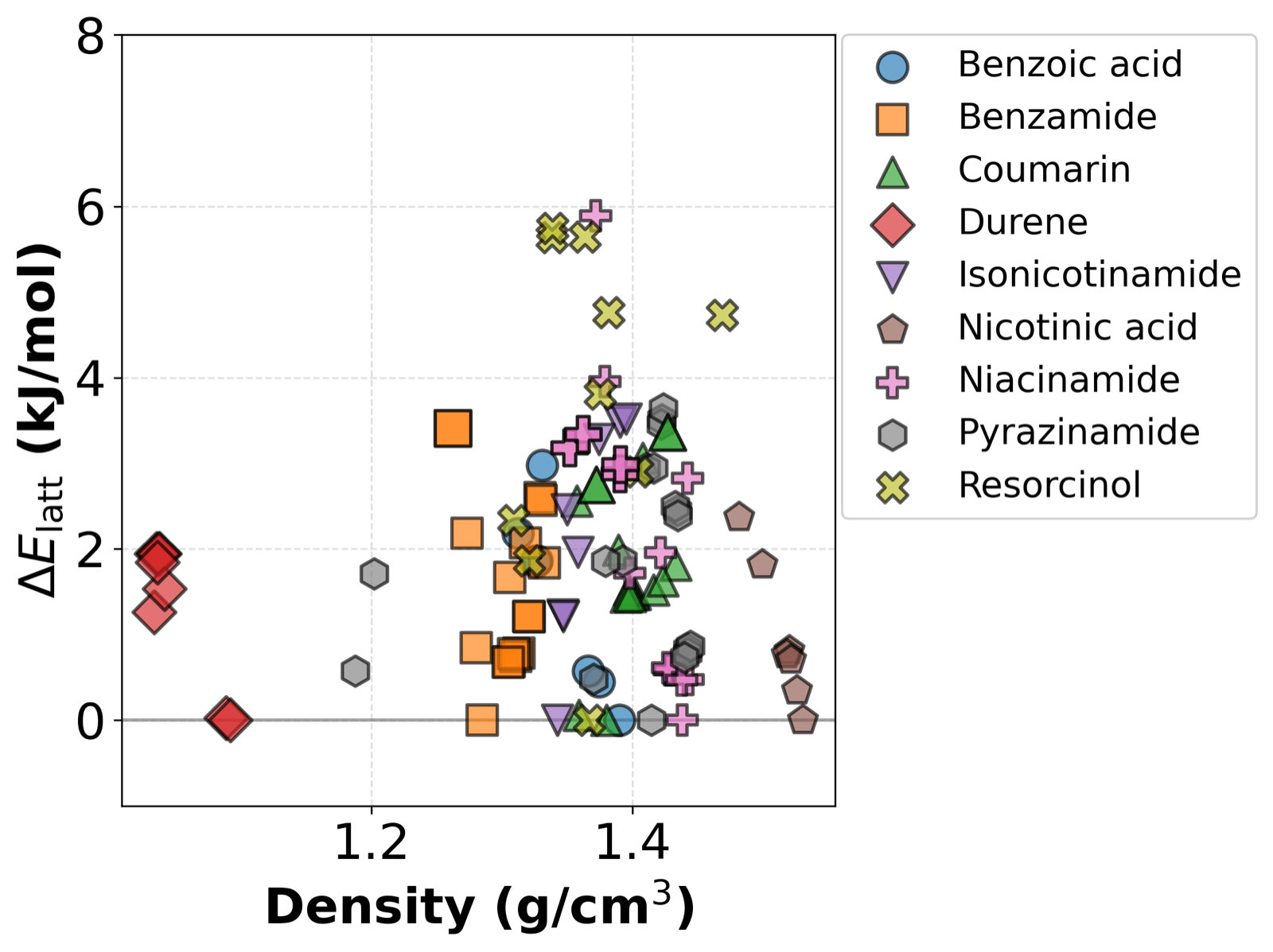}
    \caption{Relative lattice energy $\Delta E_{\text{latt}}$ as a function 
    of density for optimized crystal structures across nine polymorphic 
    systems. Each point represents a distinct polymorph, including structures 
    excluded from the training set due to large unit cells.}
    \label{fig:density_deltaE}
\end{figure}

To further assess the robustness of the trained models, we performed cell optimizations on all experimental \texttt{.cif} structures available in the CSD, including polymorphs that were excluded from the training set due to the high number of molecules in their unit cell---for example, one benzamide polymorph contains 32 molecules and one Niacinamide form contains 40 molecules, making QM calculations too expensive. Optimizations were carried out using the same procedure as above. Figure~\ref{fig:density_deltaE} shows the relative lattice energy $\Delta E_{\text{latt}}$ as a function of density for the nine systems. 
%where $Z$ is the number of molecules per unit cell and the reference is the lowest-energy polymorph for each compound. 
For all systems shown, structures converged to physically meaningful geometries, with densities and relative lattice energy ranges consistent with experimental expectations, though the spread varies across systems. These results confirm that the trained models are sufficiently robust to serve as a computationally efficient starting point for structural refinement prior to DFT calculations, or to generate diverse candidate structures for  crystal structure exploration. Detailed information about the relative lattice energies for all systems can be found in Table~\ref{tab:polymorphs} in the SI.

While Figure~\ref{fig:density_deltaE} gives a qualitative overview of the energy--density landscape, a direct comparison of atomic coordinates against the experimental structures is a more informative test of structural fidelity. We therefore computed the RMSD$_{15}$ packing similarity between each experimentally determined structure and its MLIP cell-optimized counterpart for all 105 CSD polymorphs of the nine systems, including the forms excluded from training. Across the full set the mean RMSD$_{15}$ is 0.16~\AA, with 94\% of structures below 0.3~\AA\ (Table~\ref{tab:rmsd15}), confirming that the models reproduce not merely plausible densities but the experimental packing itself. Agreement is uniform across systems, with per-system means between 0.12~\AA\ (coumarin) and 0.19~\AA\ (nicotinic acid); pyrazinamide, with 18 polymorphs, is the sole outlier at 0.26~\AA. Part of the residual deviation is expected on physical grounds: the MLIP optimizations are performed at 0~K and inherit the systematic cell contraction of the DFT reference relative to room-temperature experimental structures (mean $-4.7\%$, Table~\ref{tab:volume_contraction}), which propagates directly into a metric sensitive to lattice-parameter error. The NPT simulations (see section~\ref{sec:npt}) indicate that thermal expansion accounts for only part of this offset: at 300~K benzoic acid expands by $\sim$3\%, while nicotinic acid and resorcinol remain close to their 0~K volumes, so thermal expansion accounts for only part of the offset from the RT experimental cells.
 
\begin{table}[htbp]
\centering
\small
\caption{Structural agreement between the experimentally determined and MLIP cell-optimized structures, quantified by RMSD$_{15}$ (root-mean-square deviation of atomic positions over a cluster of the 15 nearest molecules after optimal rigid-body superposition). $N$ is the number of polymorphs compared, including forms excluded from the training set.}
\label{tab:rmsd15}
\begin{tabular}{lcc}
\toprule
System & $N$ & RMSD$_{15}$ (\AA) \\
\midrule
Benzoic acid    & 7  & 0.18 \\
Benzamide       & 16 & 0.15 \\
Coumarin        & 16 & 0.12 \\
Durene          & 8  & 0.14 \\
Isonicotinamide & 8  & 0.13 \\
Nicotinic acid  & 7  & 0.19 \\
Niacinamide     & 15 & 0.17 \\
Pyrazinamide    & 18 & 0.26 \\
Resorcinol      & 10 & 0.14 \\
\midrule
\textbf{Total / mean} & \textbf{105} & \textbf{0.16} \\
\bottomrule
\end{tabular}
\end{table}

\subsection{Dynamical Stability for Molecular Dynamics}

To assess whether the trained models support stable 
%and accurate 
MD simulations, a series of dynamical benchmarks were performed. To limit finite-size effects, all unit cells were replicated to reach a minimum dimension of 25~\AA\ in each direction prior to any simulation. The most stable polymorph of each system, as determined by DFT lattice energy, was selected for microcanonical (NVE) ensemble simulations to evaluate energy conservation. Energetic stability was quantified through the cumulative drift of the instantaneous total energy, defined as:
\begin{equation}
    \Delta E = \frac{1}{N_{\text{step}}} \sum_{k=1}^{N_{\text{step}}} 
    \frac{|E_k - E(0)|}{|E(0)|} \quad ,
\end{equation}
where $E(0)$ is the reference energy at the beginning of the analysis period (after equilibration), $E_k$ is the total energy at step $k$, and $N_{\text{step}}$ is the number of simulation steps.\cite{marksbook} Energies were recorded every 10 steps, corresponding to a logging interval of 5~fs. Simulations were performed using the \texttt{amlpa.py} module from AMLP, which enables configurable MD runs through a single \texttt{config.yaml} input file. All NVE simulations ran for 25~ps with a timestep of 0.5~fs. All nine models demonstrate excellent energy conservation, with the cumulative drift remaining in the $10^{-7}$ range throughout the simulations.

\begin{figure}[H]
    \centering
    \includegraphics[width=0.8\linewidth]{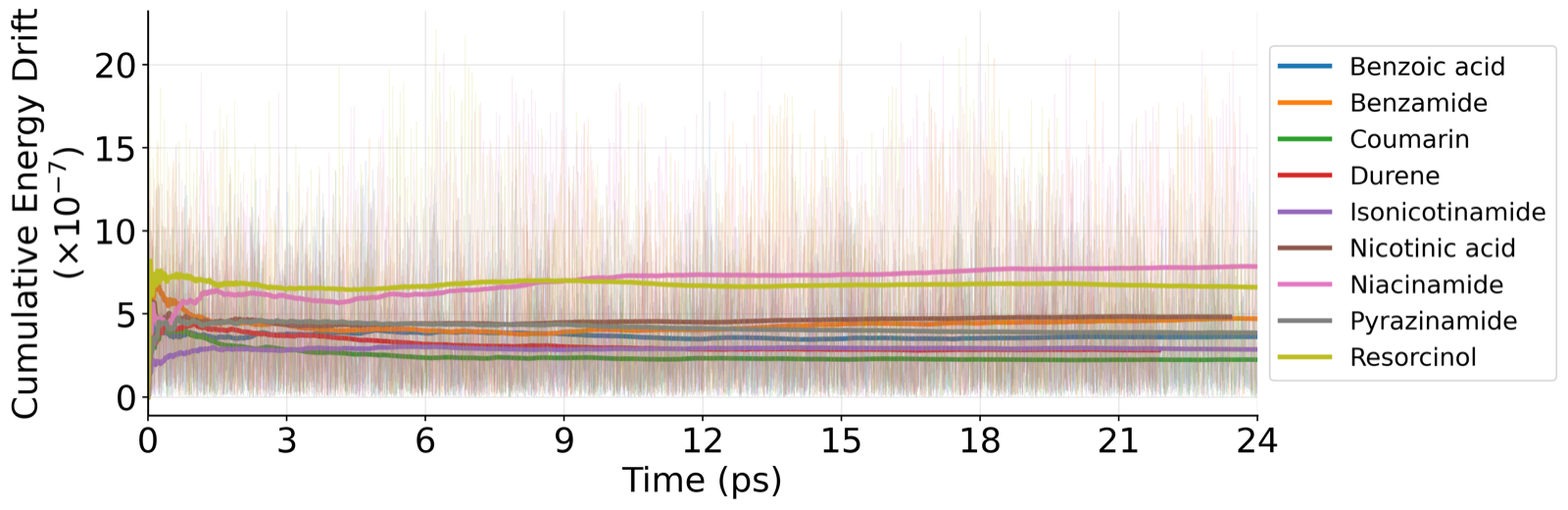}
    \caption{Cumulative energy drift $\Delta E$ from NVE simulations following 1~ps of equilibration, shown for the most stable polymorph of nine systems. Solid lines represent the cumulative drift, while shaded regions reflect instantaneous deviations.}
    \label{fig:energy_drift}
\end{figure}

%NEED TO ADD real simu for BZAMID and RESORA based on the GPU simulations (running since 03/02/2026) CPUs simulations gave only 5ps in 48 hours...

Following the NVE energy conservation tests, the thermal stability of each model was assessed through canonical (NVT) MD simulations using the Berendsen thermostat, with a timestep of 0.5~fs and a total simulation time of 25~ps, again using the \texttt{amlpa.py} module from AMLP. One simulation was performed per polymorph at each of the three target temperatures (300~K, 500~K, and 600~K), and structural integrity was monitored by tracking the temperature stability throughout each trajectory. 
%Figure~\ref{fig:temperature_NVT} shows the temperature evolution averaged over all polymorphs for each system. The absence of any abrupt temperature divergence across all simulations confirms that the trained models capture the structural physics of each polymorph and maintain stable dynamics up to 25~ps. As expected, temperature fluctuations increase with simulation temperature, consistent with the equipartition of energy, but remain well controlled in all cases.

%\begin{figure}[H]
%    \centering
%    \includegraphics[width=0.8\linewidth]{plot-Main/temperature_NVT.png}
%    \caption{Temperature evolution during NVT MD simulations at 300~K, 500~K, and 600~K for seven polymorphic systems. Each line represents the instantaneous temperature averaged over all polymorphs of a given system.}
%    \label{fig:temperature_NVT}
%\end{figure}

To assess the structural integrity of each polymorph throughout the NVT simulations, an orientational order parameter $P_2$ analysis was performed. This parameter is defined as
\begin{equation}
P_2 = \left\langle C_N\sum_{i>j} p_2(\cos(\theta_{ij}))\right\rangle = 
\left\langle {C_N \over 2}\sum_{i>j}\left(3\cos^2\theta_{ij} - 1\right)\right\rangle
\label{eq:P2}
\end{equation}
where $\theta_{ij}$ is the angle between molecular plane normal vectors for molecules $i$ and $j$, $C_N = 2/N(N-1)$, $p_2(x) = (3x^2-1)/2$ is the second-order Legendre polynomial, and the angular brackets $\langle \cdots \rangle$ denote an NVT ensemble average. A value of $P_2 \approx 1$ indicates a highly ordered crystalline arrangement, while a significant drop towards zero signals either a conformational change or a loss of crystal integrity. Pyrazinamide is used here as an illustrative example, owing to its large polymorphic landscape of 14 distinct forms, which provides a comprehensive test of the model's ability to maintain the structural identity of each polymorph across the full range of simulated temperatures.

\begin{figure}
    \centering
    \includegraphics[width=1.0\linewidth]{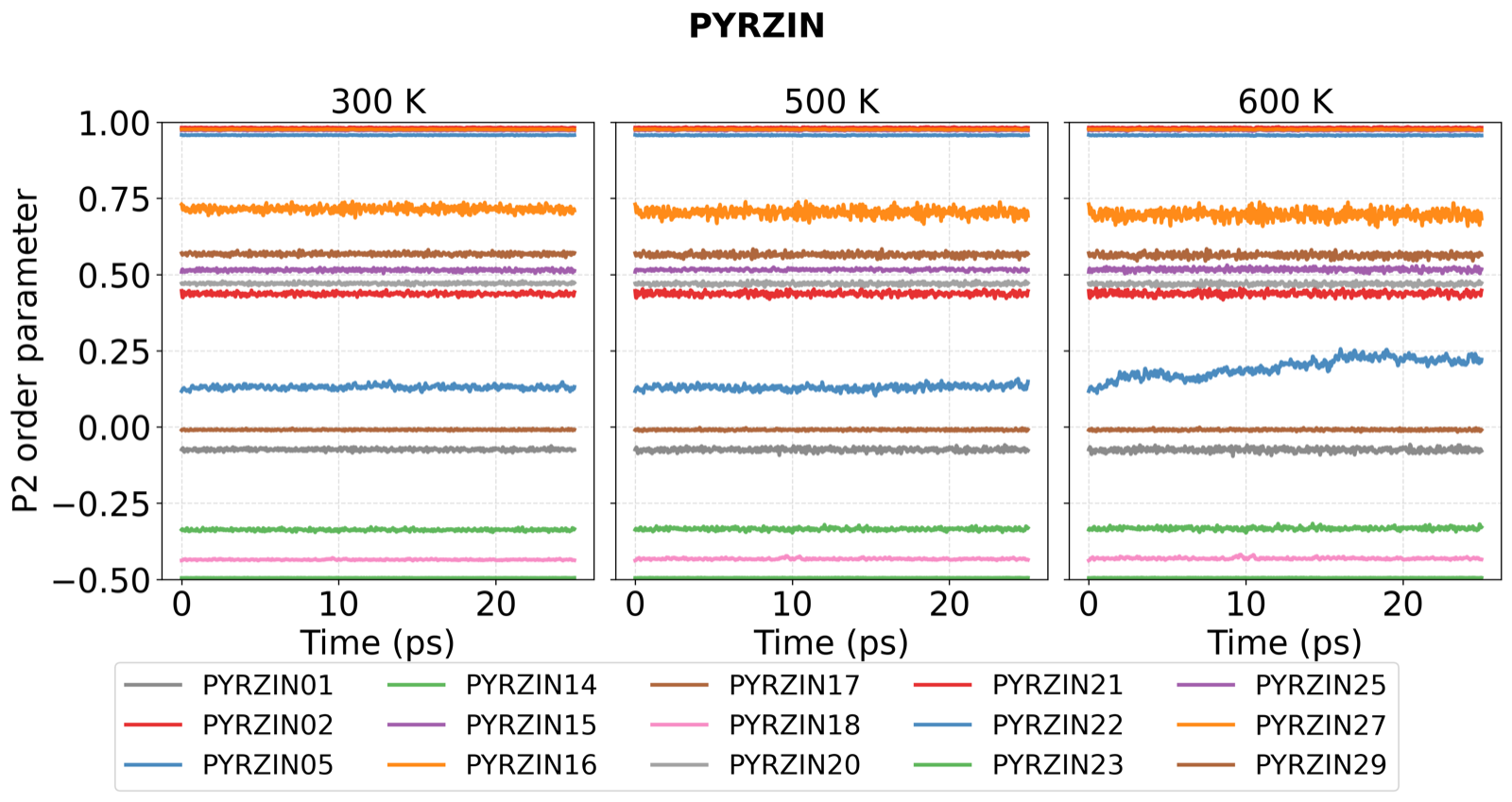}
    \caption{$P_2$ orientational order parameter as a function of simulation time for all Pyrazinamide polymorphs at 300~K, 500~K, and 600~K. Each line corresponds to a distinct polymorph.}
    \label{fig:p2_PYRZIN}
\end{figure}

As depicted in Figure~\ref{fig:p2_PYRZIN}, which shows the instantaneous values of $P_2$ over several NVT trajectories, all polymorphs of Pyrazinamide (PYRZIN) maintain their orientational order throughout the simulations. An average value of $P_2 = -0.5$ corresponds to perfect perpendicular alignment, characteristic of PYRZIN14, PYRZIN18, and PYRZIN23. A value near zero indicates random orientational disorder, as observed for PYRZIN17, whose general packing cannot be assigned to a well-defined motif. Values between 0.1 and 0.2 are indicative of a herringbone packing arrangement, as seen for PYRZIN05, while values above 0.4 correspond to parallel packing, which characterises the remaining polymorphs.

\begin{figure}
    \centering
    \includegraphics[width=1.0\linewidth]{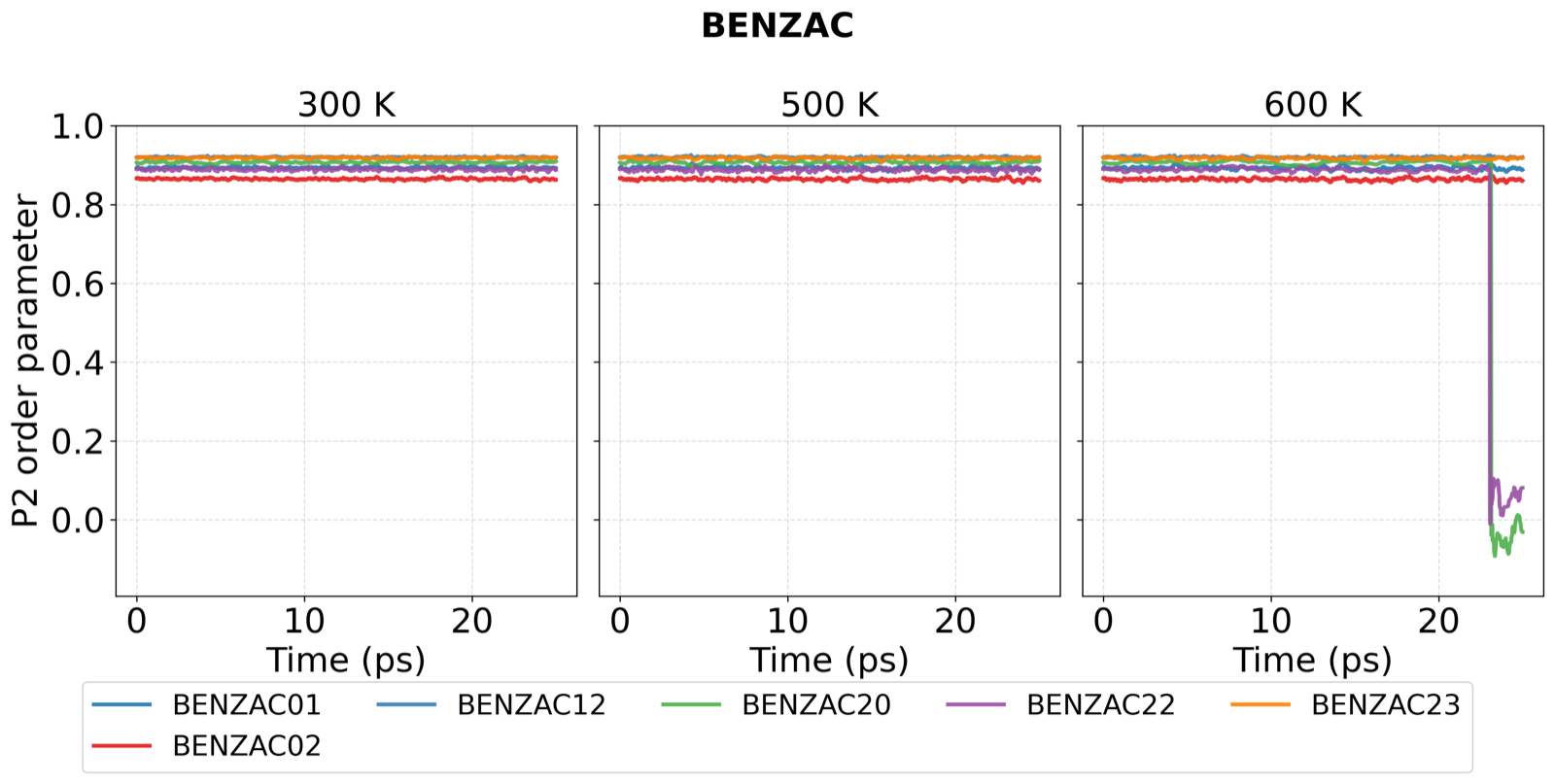}
    \caption{$P_2$ orientational order parameter as a function of simulation time for all Benzoic acid (BENZAC) polymorphs at 300~K, 500~K, and 600~K. Each line corresponds to a distinct polymorph.}
    \label{fig:p2_BENZAC}
\end{figure}

Benzoic acid exhibits parallel stacking across all its polymorphs as depicted in Figure \ref{fig:p2_BENZAC}, with instantaneous $P_2$ values ranging from 0.86 to 0.92. At 300~K and 500~K, all forms remain orientationally stable. At 600~K, however, BENZAC20 and BENZAC22 lose their structural integrity, consistent with these polymorphs reaching their thermal stability limit near or below 600~K. The remaining polymorphs retain well-defined $P_2$ values over the NVT trajectory at this temperature.  Complete $P_2$ analyses for all nine systems are provided in the Supporting Information (see \Cref{,fig:p2_bzamid,fig:p2_coumar,fig:p2_durene,fig:p2_ehowih,fig:p2_nicoam,fig:p2_resora}).

Coumarin (COUMAR) shows predominantly perpendicular alignment with a subset of parallel-stacked forms; notably, at 600~K COUMAR14 undergoes a reorientation to a distinct packing motif, consistent with this polymorph approaching its thermal stability limit. Durene (DURENE) presents a similar bimodal picture, with DURENE01 and DURENE05 adopting perpendicular stacking (average $P_2 \approx -0.5$) and DURENE06 a parallel arrangement. For Isonicotinamide (EHOWIH), a herringbone packing is observed for EHOWIH, EHOWIH01, and EHOWIH07 (average $P_2 \approx 0.35$), while EHOWIH03 adopts parallel stacking. EHOWIH shows larger $P_2$ fluctuations already at 300~K, suggesting an incipient thermal effect on orientational order at this temperature. Niacinamide (NICOAM) remains stable across most of its polymorphs, which predominantly adopt perpendicular stacking arrangements. For Benzamide (BZAMID), the majority of polymorphs lose their orientational order at 600~K, with only BZAMID16 retaining a well-defined instantaneous $P_2$ values  over the NVT trajectory at this temperature. 

%This behaviour reflects the fact that 600~K likely exceeds the thermodynamic stability limit of most Benzamide polymorphs: the model correctly captures the onset of structural disorder as the system approaches and surpasses the melting threshold.

\begin{figure}
    \centering
    \includegraphics[width=1.0\linewidth]{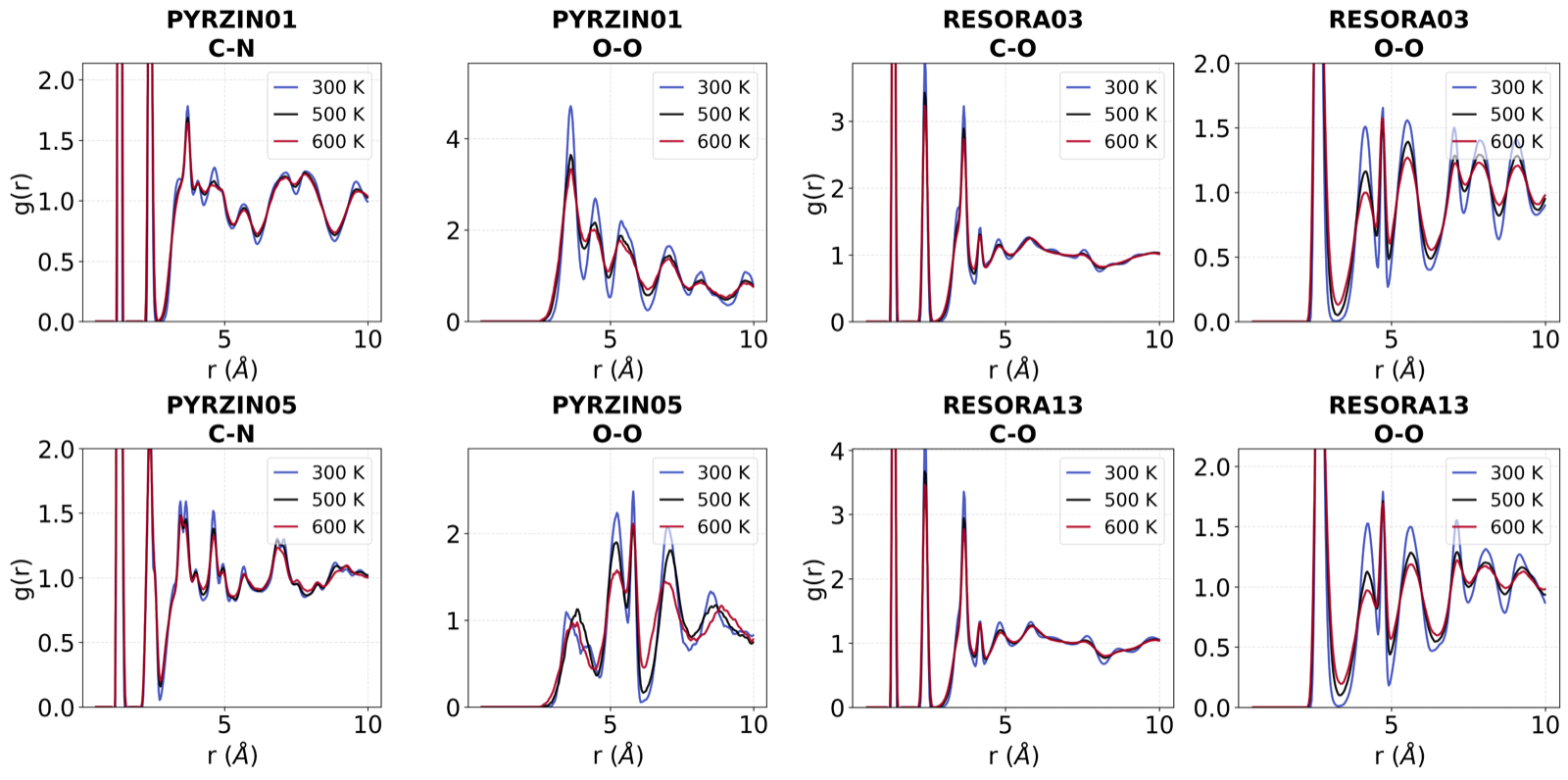}
    \caption{Radial distribution functions for selected atomic pairs of 
    Pyrazinamide (PYRZIN01, PYRZIN05) and Resorcinol (RESORA03, RESORA13) 
    at 300~K, 500~K, and 600~K. C--N and C--O pairs probe intramolecular 
    bonding integrity; O--O pairs probe intermolecular bonding and 
    packing.}
    \label{fig:rdf_pyzi_resor}
\end{figure}

RDFs were computed for each polymorph, with atomic pairs elected to probe both intramolecular bonding integrity and 
intermolecular packing independently. Figure~\ref{fig:rdf_pyzi_resor} illustrates this for two representative systems: Pyrazinamide and Resorcinol. 
For both compounds, intramolecular integrity is assessed through C--N and C--O pair correlations respectively, whose sharp first peaks confirm that covalent bonding geometry is preserved throughout the simulations. Intermolecular packing is probed through O--O correlations, which are sensitive to arrangements between molecules. As expected, RDF peaks broaden progressively with increasing temperature, reflecting thermal fluctuations, while the overall peak positions and structural motifs of each polymorph are preserved---confirming that the models keep structural integrity well within the simulated temperature range.

The complete RDF analyses for all nine systems are provided in the Supporting Information (see \Cref{fig:rdf_benzac,fig:rdf_bzamid_1,fig:rdf_bzamid_2,fig:rdf_coumar_1,fig:rdf_coumar_2,fig:rdf_durene,fig:rdf_ehowih,fig:rdf_nicoac,fig:rdf_nicoam,fig:rdf_pyrzin_1,fig:rdf_pyrzin_2,fig:rdf_resora}), and are consistent with the structural picture established by the $P_2$ analysis. 
Overall, all trained models demonstrate stable dynamics across the full range of simulated temperatures and polymorphs, confirming their suitability as reliable starting points for production MD simulations. Where orientational order is lost at elevated temperatures, this reflects the known thermal stability limits of the corresponding crystal forms.

\subsection{Constant-Pressure Stability}
\label{sec:npt}
 
The NVE and NVT benchmarks above probe energy conservation and orientational order at fixed volume. A fixed cell cannot deform, however, and may therefore conceal deficiencies in the potential that would otherwise appear as unphysical lattice relaxation; constant-pressure dynamics is the more stringent test. We therefore performed NPT simulations with a fully flexible cell (Parrinello--Rahman barostat, Nos\'e--Hoover chain thermostat) at 300~K and ambient pressure for every polymorph of three representative systems---benzoic acid, nicotinic acid and resorcinol---each for 100~ps of production dynamics.
 
\begin{figure}[H]
    \centering
    \includegraphics[width=0.9\linewidth]{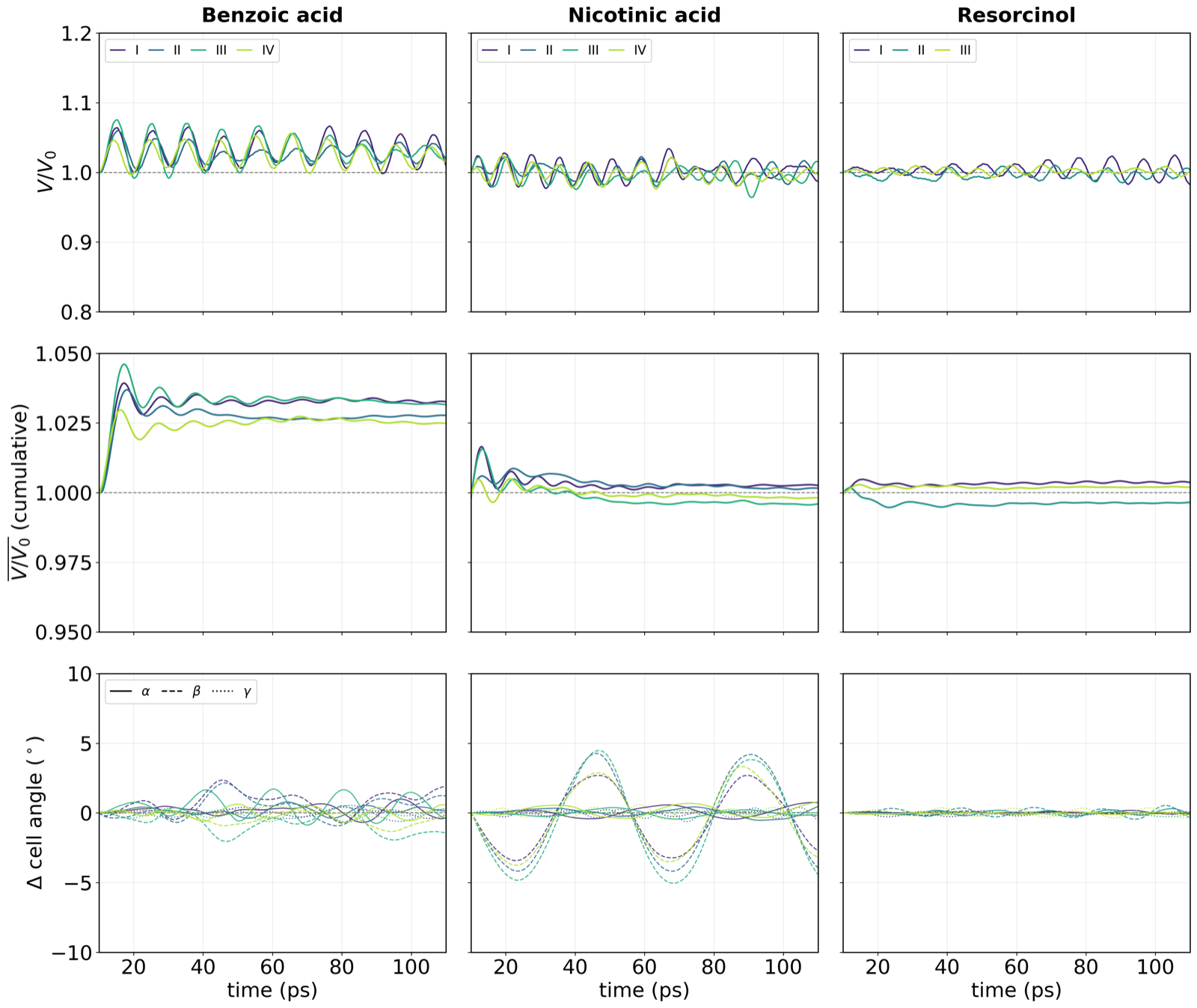}
    \caption{NPT molecular dynamics stability of the fine-tuned MolCryst-MLIPs models at 300~K and ambient pressure (fully flexible cell, Parrinello--Rahman barostat, Nos\'e--Hoover chain thermostat). Top row: cell volume relative to its initial value, $V/V_0$, for every polymorph (labelled by Roman numeral) of benzoic acid, nicotinic acid and resorcinol over 100~ps of production dynamics. Middle row: the corresponding cumulative average $\langle V/V_0\rangle_{0\to t}$. Bottom row: deviation of the three cell angles from their initial values.}
    \label{fig:npt_stability}
\end{figure}
 
As shown in Figure~\ref{fig:npt_stability}, all structures remain stable over the full trajectory. Cell volumes fluctuate about a stable mean with no systematic drift. The mean $V/V_0$ over the first and second halves of the production run differs by at most 0.5 \% for every polymorph, with a block-averaged standard error of $\sim$0.001. The three cell angles remain within a few degrees of their initial values, with no polymorph undergoing a shear instability or a transition to a different packing. The time-averaged volumes stay close to the 0~K reference. Benzoic acid expands by 2.5--3.3\%, while nicotinic acid and resorcinol remain within 0.5 \% of $V_0$. These expansions are smaller than the mean contraction of the DFT cells relative to RT experimental structures (see Table 1), so the 300~K
cells remain somewhat denser than experiment. The present test shows that the lattice, when free to deform, neither drifts nor destabilizes too much — a behavior to which fixed-cell simulations are, by construction, blind.

\section{Conclusion}
We have presented a set of fine-tuned MACE-based MLIPs for nine polymorphic molecular crystals, developed within the AMLP framework using high-quality DFT reference data. The accuracy of the underlying DFT dataset was validated through systematic analysis of unit cell volume contractions, yielding a mean 
contraction of $-4.7\%$ consistent with the 0~K nature of DFT optimizations, and through relative lattice energy differences between polymorphs, which remain within the physically expected range of below 10~kJ~mol$^{-1}$.

Fine-tuning of foundation models is essential for system-specific applications in molecular crystal polymorphism: while foundation models lack the resolution to discriminate between polymorphs of the same compound, targeted fine-tuning on DFT data recovers this capability and yields potentials that correctly rank polymorph stabilities. Beyond energy discrimination, the fine-tuned models transfer reliably across the polymorphic landscape of each target compound, successfully optimizing the geometry and unit cell of experimental crystal structures not included during training---including large-cell polymorphs that would be prohibitively expensive to relax at the DFT level---and reproducing the experimental packing with a mean RMSD$_{15}$ of 0.16~\AA\ across 105 polymorphs. We emphasise that this is transferability within, and not across, chemical systems: deployment on a chemically distinct compound requires fine-tuning a new model, for which AMLP supplies the protocol. The resulting structures exhibit physically consistent densities and relative lattice energy rankings 
in good agreement with the DFT reference, suggesting that these potentials can serve as efficient pre-relaxation tools that significantly reduce the computational cost of subsequent DFT calculations.

Dynamical validation through NVE simulations confirmed excellent energy conservation, with cumulative drift values on the order of $10^{-7}$ across all systems. Canonical NVT simulations demonstrated that all models sustain stable dynamics across most of the polymorphic landscape of each compound, as corroborated by $P_2$ orientational order parameter analysis and RDFs, and constant-pressure NPT simulations with a fully flexible cell confirmed that the potentials maintain the crystal structure, without lattice drift or shear instability, when the cell is free to relax.

Taken together, these results establish the MolCryst-MLIPs database~\cite{git-MolCryst-MLIPs,huggingface-MolCryst-MLIPs} as a first release of 
validated, transferable potentials for large-scale MD simulations of organic crystal polymorphism. The models represent strong starting points for further fine-tuning on system-specific datasets, for accelerating DFT geometry convergence, and for finite-temperature free-energy calculations, where the reduced computational cost relative to AIMD offers a decisive practical advantage. We note in this context that the harmonic and quasi-harmonic approximations, although inexpensive, prove insufficient to reproduce the experimentally observed free-energy ordering of several of these polymorphs; fully anharmonic methods are required, and an efficient workflow for such calculations built on the present potentials is under development and will be reported separately. A residual limitation is that the accuracy of the potentials is bounded by that of the PBE+D4 reference: for near-degenerate forms the uncertainty of the exchange--correlation functional is commensurate with the energy differences being resolved, so the models reproduce the DFT hierarchy rather than the exact experimental one. Crucially, the accompanying DFT datasets are released alongside the potentials: given the rapid pace of development in the MLIP community, where new foundation models with broader chemical coverage and improved accuracy continue to emerge, these curated datasets provide a ready-to-use resource for fine-tuning any future model to the molecular crystal domain without the need to regenerate expensive DFT reference data 
from scratch. By continuously integrating new compounds and polymorphic 
systems within the AMLP framework, the database will provide the community with a growing library of high-fidelity potentials and training data spanning a broad chemical space, substantially lowering the barrier to entry for researchers seeking to perform crystal structure exploration or large-scale MD simulations --- and ensuring that the investment in DFT data generation remains reusable as the landscape of available foundation models continues to evolve.

\section*{Acknowledgments}

\subsection*{Author Contributions}
Adam Lahouari (ORCID: 0000-0001-5857-1066) conceived, designed, led the study and drafted the manuscript. Jutta Rogal (ORCID: 0000-0002-6268-380X) contributed to the design of the study, writing and revision of the manuscript. Mark E. Tuckerman (ORCID: 0000-0003-2194-9955) contributed to the writing and revision of the manuscript. All authors contributed to the computational experiments and analyzed the results.

\subsection*{Funding}
This work was supported by the National Science Foundation under grant DMR-2118890. This work was supported by a grant from the Simons Foundation [MPS-T-MPS-00839534, MET]. Computational resources were provided in part by the NYU IT High Performance Computing facility. The Flatiron Institute is a division of the Simons Foundation.

\subsection*{Conflicts of Interest}
The authors declare no competing interests.

\subsection*{Data Availability}
All models and datasets are publicly available on GitHub at 
\href{https://github.com/adamlaho/MolCryst}{https://github.com/adamlaho/MolCryst} 
and on Hugging Face at 
\href{https://huggingface.co/adamlaho/MolCryst}{https://huggingface.co/adamlaho/MolCryst}.

\section*{Supporting Information}

The Supporting Information includes: DFT parameters for cell optimization and ab initio molecular dynamics; DFT-optimized structural data (lattice parameters and relative energies) for all polymorphs across all nine systems; dataset energy--force distributions; MACE model training hyperparameters, isolated atom reference energies, and force/energy correlation plots; training dataset sizes and wall-clock training times for all nine MolCryst-MLIPs models; MACE-optimized polymorph energetics ($\Delta E_{\mathrm{latt}}$ versus density) including out-of-sample polymorph validation; and radial distribution functions and P2 orientational order parameters.

\printbibliography

\newpage

\textbf{\Large TOC Graphic}

\begin{figure}[H]
    \centering
    \includegraphics[width=0.75\linewidth]{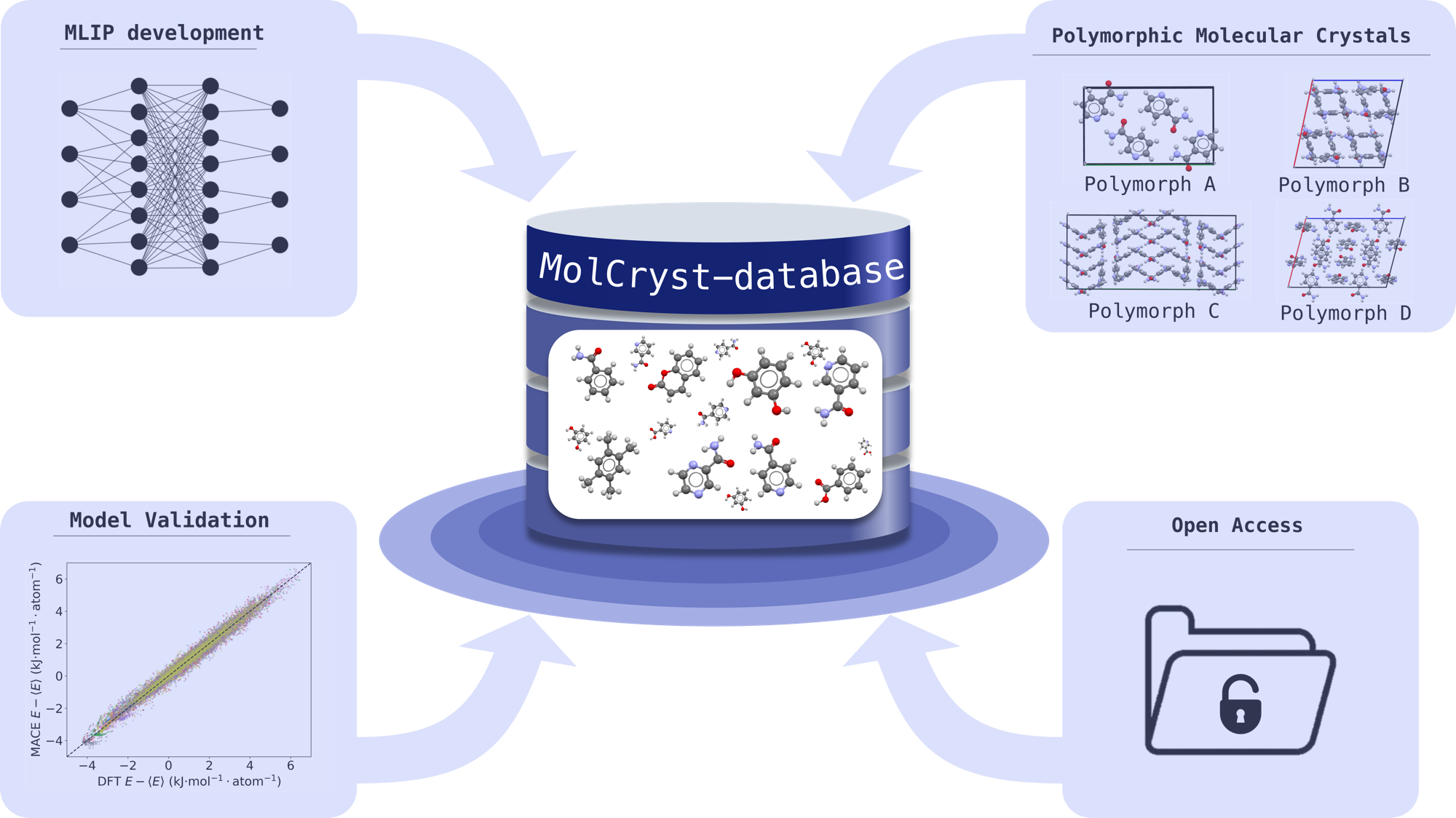}
    \label{fig:TOC}
\end{figure}

\newpage

\appendix

% Supporting Information Content
% To be included in main document with \input{SI.tex}
\DeclareSIUnit\angstrom{\text{Å}}
% Set up supplementary table numbering
\renewcommand{\thetable}{S\arabic{table}}
\setcounter{table}{0}
\renewcommand{\thefigure}{S\arabic{figure}}
\setcounter{figure}{0}
\DeclareSIUnit\angstrom{\text{Å}}

% Configure page numbering for SI
% Simple page numbering configuration
\renewcommand{\thepage}{S-\arabic{page}}
\setcounter{page}{1}

\begin{center}
    {\large\textbf{Supporting Information for:}}\\[0.4cm]
    {\large MolCryst-MLIPs: A Machine-Learned Interatomic Potentials Database for Molecular Crystals}\\[0.4cm]
    Adam Lahouari,$^{1,*}$ Shen Ai,$^{4}$ Jihye Han,$^{1}$ Jillian Hoffstadt,$^{1}$ Philipp H{\"o}llmer,$^{1,2,4}$
    Charlotte Infante,$^{1}$ Pulkita Jain,$^{8}$ Sangram Kadam,$^{1}$ Maya M. Martirossyan,$^{1,4}$
    Amara McCune,$^{4}$ Hypatia Newton,$^{2}$ Shlok J. Paul,$^{8}$ Willmor Pena,$^{1}$
    Jonathan Raghoonanan,$^{4,7}$ Sumon Sahu,$^{4}$ Oliver Tan,$^{1}$ Andrea Vergara,$^{4}$
    Jutta Rogal,$^{3}$ and Mark E. Tuckerman$^{1,2,4,5,6}$\\[0.3cm]
    {\small $^{1}$Department of Chemistry, New York University, New York, NY 10003, USA}\\
    {\small $^{2}$Simons Center for Computational Physical Chemistry, New York University, New York, NY 10003, USA}\\
    {\small $^{3}$Initiative for Computational Catalysis, Flatiron Institute, New York, NY 10010, USA}\\
    {\small $^{4}$Department of Physics, New York University, New York, NY 10003, USA}\\
    {\small $^{5}$Courant Institute of Mathematical Sciences, New York University, New York, NY 10012, USA}\\
    {\small $^{6}$NYU-ECNU Center for Computational Chemistry, Shanghai 200062, China}\\
    {\small $^{7}$New York University Shanghai, 567 West Yangsi Road, Pudong, Shanghai 200126, China}\\
    {\small $^{8}$Department of Chemical and Biomolecular Engineering, Tandon School of Engineering, New York University, Brooklyn, New York 11201, United States}\\
    {\small $^{*}$\textit{Correspondence:} al9500@nyu.edu}
\end{center}
\vspace{0.5cm}

%\tableofcontents
\newpage

\section{Computational Details}

\subsection{DFT Calculations}

All \textit{ab initio} calculations were performed using the Vienna Ab Initio Simulation Package (VASP) version 6.x with projector-augmented wave (PAW) pseudopotentials. The Perdew-Burke-Ernzerhof (PBE) exchange-correlation functional was employed in conjunction with the Grimme D4 dispersion correction to accurately capture van der Waals interactions critical for molecular crystal systems.

\subsubsection{Geometry Optimization Parameters}

DFT geometry optimizations (0~K) were performed with the following settings:

\begin{table}[H]
\centering
\caption{DFT Cell Optimization Parameters}
\begin{tabular}{@{}ll@{}}
\toprule
\textbf{Parameter} & \textbf{Value} \\
\midrule
Exchange-correlation functional & PBE \\
Dispersion correction & Grimme D4 (IVDW = 13) \\
Plane-wave cutoff energy (ENCUT) & 650 eV \\
Electronic convergence (EDIFF) & $10^{-7}$ eV \\
Ionic relaxation algorithm (IBRION) & 2 (Conjugate gradient) \\
Stress/cell optimization (ISIF) & 3 (relax ions + cell shape + volume) \\
Maximum ionic steps (NSW) & 150 \\
Smearing method (ISMEAR) & $-1$ (Fermi smearing) \\
Smearing width (SIGMA) & 0.05 eV \\
Precision (PREC) & Accurate \\
Real-space projection (LREAL) & Auto \\
Maximum electronic steps (NELM) & 150 \\
\bottomrule
\end{tabular}
\end{table}

\subsubsection{\textit{Ab Initio} Molecular Dynamics Parameters}

AIMD simulations were conducted in the canonical (NVT) ensemble using the Langevin thermostat with the following parameters:

\begin{table}[H]
\centering
\caption{AIMD Simulation Parameters}
\begin{tabular}{@{}ll@{}}
\toprule
\textbf{Parameter} & \textbf{Value} \\
\midrule
Exchange-correlation functional & PBE \\
Dispersion correction & Grimme D4 (IVDW = 13) \\
Plane-wave cutoff energy (ENCUT) & 650 eV \\
Electronic convergence (EDIFF) & $10^{-7}$ eV \\
MD algorithm (IBRION) & 0 (Molecular dynamics) \\
MD thermostat (MDALGO) & 3 (Langevin) \\
Timestep (POTIM) & 0.75 fs \\
Temperature range & 25--600 K \\
Friction coefficients (LANGEVIN\_GAMMA) & 10.0 ps$^{-1}$ (all species) \\
Lattice friction (LANGEVIN\_GAMMA\_L) & 1.0 \\
Total MD steps (NSW) & 50,000 \\
Stress/cell control (ISIF) & 2 (relax ions only) \\
Smearing method (ISMEAR) & $-1$ (Fermi smearing) \\
Smearing width (SIGMA) & 0.05 eV \\
Precision (PREC) & High \\
Real-space projection (LREAL) & Auto \\
Symmetry (ISYM) & 0 (off for MD) \\
Output frequency (NBLOCK) & 1 \\
k-point blocking (KBLOCK) & 10 \\
\bottomrule
\end{tabular}
\end{table}

\begin{landscape}
\begin{longtable}{@{}llccccccccccc@{}}
\caption{DFT-Optimized and Experimental Structures for All Polymorphs} \\
\toprule
\textbf{Compound} & \textbf{Polymorph} & \textbf{Type} & \textbf{$\Delta E_{\text{lattice}}^{relative}$} & \textbf{$E_{\text{per mol}}$} & \textbf{$a$} & \textbf{$b$} & \textbf{$c$} & \textbf{$\alpha$} & \textbf{$\beta$} & \textbf{$\gamma$} & \textbf{Volume} & \textbf{$\Delta V$} \\
 & & & \textbf{(kJ/mol)} & \textbf{(eV)} & \textbf{(\AA)} & \textbf{(\AA)} & \textbf{(\AA)} & \textbf{($^\circ$)} & \textbf{($^\circ$)} & \textbf{($^\circ$)} & \textbf{(\AA$^3$)} & \textbf{(\%)} \\
\midrule
\endfirsthead

\multicolumn{13}{c}{\tablename\ \thetable\ -- \textit{Continued from previous page}} \\
\toprule
\textbf{Compound} & \textbf{Polymorph} & \textbf{Type} & \textbf{$\Delta E_{\text{lattice}}^{relative}$} & \textbf{$E_{\text{per mol}}$} & \textbf{$a$} & \textbf{$b$} & \textbf{$c$} & \textbf{$\alpha$} & \textbf{$\beta$} & \textbf{$\gamma$} & \textbf{Volume} & \textbf{$\Delta V$} \\
 & & & \textbf{(kJ/mol)} & \textbf{(eV)} & \textbf{(\AA)} & \textbf{(\AA)} & \textbf{(\AA)} & \textbf{($^\circ$)} & \textbf{($^\circ$)} & \textbf{($^\circ$)} & \textbf{(\AA$^3$)} & \textbf{(\%)} \\
\midrule
\endhead

\midrule
\multicolumn{13}{r}{\textit{Continued on next page}} \\
\endfoot

\bottomrule
\endlastfoot

% ----------------------------------------------------------------
% Resorcinol (RESORA) -- 4 polymorphs, \multirow{8}
% ----------------------------------------------------------------
\multirow{8}{*}{\shortstack[l]{\textit{Resorcinol} \\ \cite{resora_poly01,resora_poly02}}}
 & resora03 & DFT-Opt & 1.55 & $-90.946$ & 10.333 & 9.266 & 5.562 & 90.00 & 90.00 & 90.00 & 532.55 & $-4.6$ \\
 &          & Exp     & ---  & ---        & 10.470 & 9.406 & 5.666 & 90.00 & 90.00 & 90.00 & 557.95 & ---    \\
\cmidrule(lr){2-13}
 & resora08 & DFT-Opt & 4.64 & $-90.914$ & 7.785  & 12.467 & 5.360 & 90.00 & 90.00 & 90.00 & 520.26 & $-5.6$ \\
 &          & Exp     & ---  & ---        & 7.934  & 12.606 & 5.511 & 90.00 & 90.00 & 90.00 & 551.19 & ---    \\
\cmidrule(lr){2-13}
 & resora13 & DFT-Opt & 0.00 & $-90.962$ & 10.339 & 9.289 & 5.555 & 90.00 & 90.00 & 90.00 & 533.52 & $-5.1$ \\
 &          & Exp     & ---  & ---        & 10.530 & 9.530 & 5.600 & 90.00 & 90.00 & 90.00 & 561.97 & ---    \\
\cmidrule(lr){2-13}
 & resora16 & DFT-Opt & 1.27 & $-90.948$ & 10.327 & 9.250 & 5.574 & 90.00 & 90.00 & 90.00 & 532.48 & $-1.0$ \\
 &          & Exp     & ---  & ---        & 10.344 & 9.271 & 5.611 & 90.00 & 90.00 & 90.00 & 538.10 & ---    \\
\midrule

% ----------------------------------------------------------------
% Pyrazinamide (PYRZIN) -- 14 polymorphs, \multirow{28}
% ----------------------------------------------------------------
\multirow{28}{*}{\shortstack[l]{\textit{Pyrazinamide} \\ \cite{pyr_poly6,pyr_poly4,pyr_poly5,pyr_poly3,pyr_poly18}}}
 & pyrzin01 & DFT-Opt & 0.75 & $-96.262$ & 14.182 & 3.620 & 10.476 & 90.00 & 100.76 & 90.00 & 528.41 & $-5.6$ \\
 &          & Exp     & ---  & ---        & 14.372 & 3.711 & 10.726 & 90.00 & 101.92 & 90.00 & 559.73 & ---    \\
\cmidrule(lr){2-13}
 & pyrzin02 & DFT-Opt & 3.11 & $-96.237$ & 5.655 & 5.118 & 9.764 & 98.08 & 97.62 & 106.94 & 263.13 & $-5.9$ \\
 &          & Exp     & ---  & ---        & 5.728 & 5.221 & 9.945 & 96.81 & 97.27 & 106.22 & 279.56 & ---    \\
\cmidrule(lr){2-13}
 & pyrzin14 & DFT-Opt & 3.31 & $-96.235$ & 22.647 & 6.704 & 3.595 & 90.00 & 100.82 & 90.00 & 536.12 & $-7.2$ \\
 &          & Exp     & ---  & ---        & 23.200 & 6.770 & 3.750 & 90.00 & 101.20 & 90.00 & 577.77 & ---    \\
\cmidrule(lr){2-13}
 & pyrzin15 & DFT-Opt & 3.40 & $-96.234$ & 3.596 & 6.701 & 22.246 & 90.00 & 92.32 & 90.00 & 535.72 & $-2.0$ \\
 &          & Exp     & ---  & ---        & 3.615 & 6.738 & 22.464 & 90.00 & 92.50 & 90.00 & 546.64 & ---    \\
\cmidrule(lr){2-13}
 & pyrzin16 & DFT-Opt & 2.47 & $-96.244$ & 5.112 & 5.655 & 9.780 & 97.61 & 98.03 & 106.97 & 263.24 & $-2.1$ \\
 &          & Exp     & ---  & ---        & 5.119 & 5.705 & 9.857 & 97.46 & 98.17 & 106.47 & 268.82 & ---    \\
\cmidrule(lr){2-13}
 & pyrzin17 & DFT-Opt & 4.32 & $-96.225$ & 7.124 & 3.664 & 10.457 & 90.00 & 106.00 & 90.00 & 262.37 & $-4.9$ \\
 &          & Exp     & ---  & ---        & 7.183 & 3.729 & 10.758 & 90.00 & 106.77 & 90.00 & 275.92 & ---    \\
\cmidrule(lr){2-13}
 & pyrzin18 & DFT-Opt & 0.81 & $-96.261$ & 14.188 & 3.618 & 10.473 & 90.00 & 100.72 & 90.00 & 528.17 & $-2.3$ \\
 &          & Exp     & ---  & ---        & 14.315 & 3.624 & 10.616 & 90.00 & 101.12 & 90.00 & 540.35 & ---    \\
\cmidrule(lr){2-13}
 & pyrzin20 & DFT-Opt & 4.35 & $-96.225$ & 7.124 & 3.663 & 10.458 & 90.00 & 106.02 & 90.00 & 262.34 & $-1.8$ \\
 &          & Exp     & ---  & ---        & 7.170 & 3.648 & 10.648 & 90.00 & 106.35 & 90.00 & 267.23 & ---    \\
\cmidrule(lr){2-13}
 & pyrzin21 & DFT-Opt & 3.37 & $-96.235$ & 3.597 & 6.701 & 22.248 & 90.00 & 91.95 & 90.00 & 535.97 & $-3.3$ \\
 &          & Exp     & ---  & ---        & 3.654 & 6.736 & 22.537 & 90.00 & 92.22 & 90.00 & 554.29 & ---    \\
\cmidrule(lr){2-13}
 & pyrzin22 & DFT-Opt & 3.40 & $-96.234$ & 3.596 & 6.701 & 22.252 & 90.00 & 92.47 & 90.00 & 535.75 & $-2.1$ \\
 &          & Exp     & ---  & ---        & 3.617 & 6.741 & 22.462 & 90.00 & 92.39 & 90.00 & 547.28 & ---    \\
\cmidrule(lr){2-13}
 & pyrzin23 & DFT-Opt & 0.00 & $-96.269$ & 14.190 & 3.619 & 10.468 & 90.00 & 100.69 & 90.00 & 528.20 & $-2.3$ \\
 &          & Exp     & ---  & ---        & 14.340 & 3.621 & 10.613 & 90.00 & 101.04 & 90.00 & 540.88 & ---    \\
\cmidrule(lr){2-13}
 & pyrzin25 & DFT-Opt & 3.16 & $-96.237$ & 5.111 & 5.654 & 9.786 & 97.63 & 98.01 & 107.04 & 263.24 & $-1.5$ \\
 &          & Exp     & ---  & ---        & 5.103 & 5.708 & 9.849 & 97.53 & 98.50 & 106.56 & 267.37 & ---    \\
\cmidrule(lr){2-13}
 & pyrzin27 & DFT-Opt & 3.42 & $-96.234$ & 3.596 & 6.701 & 22.251 & 90.00 & 92.67 & 90.00 & 535.58 & $-1.5$ \\
 &          & Exp     & ---  & ---        & 3.588 & 6.761 & 22.434 & 90.00 & 92.75 & 90.00 & 543.62 & ---    \\
\cmidrule(lr){2-13}
 & pyrzin29 & DFT-Opt & 0.82 & $-96.261$ & 14.193 & 3.619 & 10.464 & 90.00 & 100.67 & 90.00 & 528.13 & $-2.0$ \\
 &          & Exp     & ---  & ---        & 14.343 & 3.609 & 10.602 & 90.00 & 100.98 & 90.00 & 538.72 & ---    \\
\midrule

% ----------------------------------------------------------------
% Niacinamide (NICOAM) -- 7 polymorphs, \multirow{14}
% ----------------------------------------------------------------
\multirow{14}{*}{\shortstack[l]{\textit{Niacinamide} \\ \cite{nicoam_poly1,nicoam_poly2,nicoam_poly3}}}
 & nicoam01 & DFT-Opt & 0.03 & $-101.085$ & 3.856 & 15.472 & 9.226 & 90.00 & 97.38 & 90.00 & 545.80 & $-2.7$ \\
 &          & Exp     & ---  & ---         & 3.877 & 15.600 & 9.375 & 90.00 & 98.45 & 90.00 & 560.86 & ---    \\
\cmidrule(lr){2-13}
 & nicoam05 & DFT-Opt & 0.00 & $-101.086$ & 3.859 & 15.444 & 9.228 & 90.00 & 97.32 & 90.00 & 545.47 & $-5.8$ \\
 &          & Exp     & ---  & ---         & 3.974 & 15.642 & 9.430 & 90.00 & 99.02 & 90.00 & 578.98 & ---    \\
\cmidrule(lr){2-13}
 & nicoam07 & DFT-Opt & 1.47 & $-101.070$ & 3.770 & 14.287 & 5.076 & 90.00 & 94.21 & 90.00 & 272.63 & $-2.6$ \\
 &          & Exp     & ---  & ---         & 3.812 & 14.388 & 5.119 & 90.00 & 94.26 & 90.00 & 280.03 & ---    \\
\cmidrule(lr){2-13}
 & nicoam08 & DFT-Opt & 0.99 & $-101.075$ & 3.724 & 12.274 & 12.960 & 71.11 & 84.77 & 84.54 & 556.74 & $-2.3$ \\
 &          & Exp     & ---  & ---         & 3.752 & 12.323 & 13.062 & 71.50 & 85.68 & 85.20 & 570.01 & ---    \\
\cmidrule(lr){2-13}
 & nicoam13 & DFT-Opt & 0.03 & $-101.085$ & 3.857 & 15.481 & 9.219 & 90.00 & 97.31 & 90.00 & 545.97 & $-3.2$ \\
 &          & Exp     & ---  & ---         & 3.883 & 15.645 & 9.384 & 90.00 & 98.39 & 90.00 & 563.90 & ---    \\
\cmidrule(lr){2-13}
 & nicoam17 & DFT-Opt & 2.47 & $-101.060$ & 10.077 & 5.960 & 9.736 & 90.00 & 100.00 & 90.00 & 575.83 & $-2.2$ \\
 &          & Exp     & ---  & ---         & 9.901  & 5.877 & 10.278 & 90.00 & 100.00 & 90.00 & 589.03 & ---    \\
\cmidrule(lr){2-13}
 & nicoam18 & DFT-Opt & 3.02 & $-101.054$ & 7.429 & 7.905 & 10.694 & 108.02 & 102.72 & 96.55 & 571.24 & $-2.4$ \\
 &          & Exp     & ---  & ---         & 7.556 & 7.941 & 10.797 & 108.10 & 102.60 & 98.29 & 585.14 & ---    \\
\midrule

% ----------------------------------------------------------------
% Nicotinic acid (NICOAC) -- 4 polymorphs, \multirow{8}
% ----------------------------------------------------------------
\multirow{8}{*}{\shortstack[l]{\textit{Nicotinic acid} \\ \cite{nicoac_poly1,nicoac_poly2,nicoac_poly3}}}
 & nicoac01 & DFT-Opt & 0.00 & $-95.781$ & 7.128 & 11.585 & 7.001 & 90.00 & 114.54 & 90.00 & 525.96 & $-5.7$ \\
 &          & Exp     & ---  & ---        & 7.162 & 11.703 & 7.242 & 90.00 & 113.20 & 90.00 & 557.92 & ---    \\
\cmidrule(lr){2-13}
 & nicoac05 & DFT-Opt & 1.63 & $-95.764$ & 7.134 & 11.597 & 6.998 & 90.00 & 114.97 & 90.00 & 524.85 & $-8.4$ \\
 &          & Exp     & ---  & ---        & 7.303 & 11.693 & 7.330 & 90.00 & 113.68 & 90.00 & 573.24 & ---    \\
\cmidrule(lr){2-13}
 & nicoac07 & DFT-Opt & 0.89 & $-95.772$ & 7.141 & 11.593 & 7.002 & 90.00 & 115.13 & 90.00 & 524.70 & $-2.8$ \\
 &          & Exp     & ---  & ---        & 7.167 & 11.671 & 7.106 & 90.00 & 114.78 & 90.00 & 539.63 & ---    \\
\cmidrule(lr){2-13}
 & nicoac08 & DFT-Opt & 1.70 & $-95.763$ & 7.129 & 11.589 & 6.995 & 90.00 & 114.77 & 90.00 & 524.67 & $-5.5$ \\
 &          & Exp     & ---  & ---        & 7.176 & 11.670 & 7.227 & 90.00 & 113.52 & 90.00 & 554.93 & ---    \\
\midrule

% ----------------------------------------------------------------
% Isonicotinamide (EHOWIH) -- 5 polymorphs, \multirow{10}
% ----------------------------------------------------------------
\multirow{10}{*}{\shortstack[l]{\textit{Isonicotinamide} \\ \cite{ehowhi_poly1,ehowhi_poly2,ehowhi_poly3,ehowhi_poly4}}}
 & ehowih    & DFT-Opt & 1.40 & $-101.057$ & 10.088 & 5.692 & 9.899 & 90.00 & 98.00 & 90.00 & 562.92 & $-2.7$ \\
 &           & Exp     & ---  & ---         & 10.166 & 5.730 & 10.033 & 90.00 & 98.17 & 90.00 & 578.55 & ---    \\
\cmidrule(lr){2-13}
 & ehowih01  & DFT-Opt & 1.41 & $-101.057$ & 10.098 & 5.687 & 9.900 & 90.00 & 98.02 & 90.00 & 563.01 & $-2.8$ \\
 &           & Exp     & ---  & ---         & 10.176 & 5.732 & 10.034 & 90.00 & 98.04 & 90.00 & 579.48 & ---    \\
\cmidrule(lr){2-13}
 & ehowih03  & DFT-Opt & 0.69 & $-101.065$ & 10.002 & 7.228 & 15.689 & 90.00 & 90.00 & 90.00 & 1134.33 & $-3.9$ \\
 &           & Exp     & ---  & ---         & 10.160 & 7.323 & 15.872 & 90.00 & 90.00 & 90.00 & 1180.95 & ---     \\
\cmidrule(lr){2-13}
 & ehowih05  & DFT-Opt & 0.00 & $-101.072$ & 5.053 & 9.286 & 12.019 & 90.00 & 89.67 & 90.00 & 563.97 & $-6.4$ \\
 &           & Exp     & ---  & ---         & 5.192 & 9.466 & 12.259 & 90.00 & 91.22 & 90.00 & 602.40 & ---    \\
\cmidrule(lr){2-13}
 & ehowih07  & DFT-Opt & 1.38 & $-101.058$ & 10.088 & 5.693 & 9.900 & 90.00 & 97.92 & 90.00 & 563.10 & $-4.5$ \\
 &           & Exp     & ---  & ---         & 10.229 & 5.754 & 10.095 & 90.00 & 97.28 & 90.00 & 589.36 & ---    \\
\midrule

% ----------------------------------------------------------------
% Durene (DURENE) -- 3 polymorphs, \multirow{6}
% ----------------------------------------------------------------
\multirow{6}{*}{\shortstack[l]{\textit{Durene} \\ \cite{durene_poly01}}}
 & durene01 & DFT-Opt & 0.00 & $-144.014$ & 11.583 & 5.493 & 6.753 & 90.00 & 112.56 & 90.00 & 396.82 & $-8.2$ \\
 &          & Exp     & ---  & ---         & 11.570 & 5.770 & 7.030 & 90.00 & 112.93 & 90.00 & 432.23 & ---    \\
\cmidrule(lr){2-13}
 & durene05 & DFT-Opt & 0.05 & $-144.013$ & 6.759 & 11.581 & 5.492 & 90.00 & 90.00 & 112.71 & 396.55 & $-7.6$ \\
 &          & Exp     & ---  & ---         & 7.023 & 11.570 & 5.733 & 90.00 & 90.00 & 112.87 & 429.21 & ---    \\
\cmidrule(lr){2-13}
 & durene06 & DFT-Opt & 0.09 & $-144.013$ & 6.770 & 5.493 & 10.912 & 90.00 & 102.10 & 90.00 & 396.75 & $-3.4$ \\
 &          & Exp     & ---  & ---         & 6.890 & 5.620 & 10.873 & 90.00 & 102.72 & 90.00 & 410.69 & ---    \\
\midrule

% ----------------------------------------------------------------
% Coumarin (COUMAR) -- 12 polymorphs, \multirow{24}
% ----------------------------------------------------------------
\multirow{24}{*}{\shortstack[l]{\textit{Coumarin} \\ \cite{shtukenberg2017powder,zhang2021structural}}}
 & coumar01 & DFT-Opt & 1.42 & $-117.840$ & 15.277 & 5.491 & 7.824 & 90.00 & 90.00 & 90.00 & 656.31 & $-5.5$ \\
 &          & Exp     & ---  & ---         & 15.500 & 5.664 & 7.915 & 90.00 & 90.00 & 90.00 & 694.87 & ---    \\
\cmidrule(lr){2-13}
 & coumar02 & DFT-Opt & 1.45 & $-117.841$ & 15.277 & 5.491 & 7.822 & 90.00 & 90.00 & 90.00 & 656.17 & $-5.7$ \\
 &          & Exp     & ---  & ---         & 15.503 & 5.666 & 7.918 & 90.00 & 90.00 & 90.00 & 695.52 & ---    \\
\cmidrule(lr){2-13}
 & coumar11 & DFT-Opt & 1.57 & $-117.839$ & 5.489 & 7.832 & 15.265 & 90.00 & 90.00 & 90.00 & 656.14 & $-2.3$ \\
 &          & Exp     & ---  & ---         & 5.609 & 7.734 & 15.478 & 90.00 & 90.00 & 90.00 & 671.47 & ---    \\
\cmidrule(lr){2-13}
 & coumar12 & DFT-Opt & 1.51 & $-117.840$ & 15.259 & 5.492 & 7.830 & 90.00 & 90.00 & 90.00 & 656.15 & $-5.5$ \\
 &          & Exp     & ---  & ---         & 15.502 & 5.663 & 7.910 & 90.00 & 90.00 & 90.00 & 694.43 & ---    \\
\cmidrule(lr){2-13}
 & coumar13 & DFT-Opt & 0.00 & $-117.856$ & 3.761 & 15.360 & 5.749 & 90.00 & 94.82 & 90.00 & 330.93 & $-6.9$ \\
 &          & Exp     & ---  & ---         & 3.980 & 15.291 & 5.858 & 90.00 & 94.24 & 90.00 & 355.50 & ---    \\
\cmidrule(lr){2-13}
 & coumar14 & DFT-Opt & 1.38 & $-117.841$ & 16.759 & 5.860 & 13.652 & 90.00 & 90.00 & 90.00 & 1340.69 & $-2.6$ \\
 &          & Exp     & ---  & ---         & 16.782 & 5.921 & 13.852 & 90.00 & 90.00 & 90.00 & 1376.56 & ---     \\
\cmidrule(lr){2-13}
 & coumar18 & DFT-Opt & 1.93 & $-117.836$ & 4.757 & 6.792 & 20.335 & 90.00 & 90.00 & 90.00 & 657.05 & $-5.9$ \\
 &          & Exp     & ---  & ---         & 4.868 & 6.882 & 20.851 & 90.00 & 90.00 & 90.00 & 698.47 & ---    \\
\cmidrule(lr){2-13}
 & coumar19 & DFT-Opt & 1.54 & $-117.840$ & 15.268 & 5.493 & 7.825 & 90.00 & 90.00 & 90.00 & 656.24 & $-4.0$ \\
 &          & Exp     & ---  & ---         & 15.500 & 5.636 & 7.822 & 90.00 & 90.00 & 90.00 & 683.35 & ---    \\
\cmidrule(lr){2-13}
 & coumar20 & DFT-Opt & 0.12 & $-117.854$ & 3.759 & 15.312 & 5.766 & 90.00 & 94.17 & 90.00 & 331.04 & $-3.7$ \\
 &          & Exp     & ---  & ---         & 3.870 & 15.284 & 5.821 & 90.00 & 93.61 & 90.00 & 343.63 & ---    \\
\cmidrule(lr){2-13}
 & coumar21 & DFT-Opt & 1.54 & $-117.840$ & 15.290 & 5.495 & 7.812 & 90.00 & 90.00 & 90.00 & 656.44 & $-5.7$ \\
 &          & Exp     & ---  & ---         & 15.526 & 5.669 & 7.912 & 90.00 & 90.00 & 90.00 & 696.37 & ---    \\
\cmidrule(lr){2-13}
 & coumar22 & DFT-Opt & 1.54 & $-117.840$ & 15.283 & 5.491 & 7.821 & 90.00 & 90.00 & 90.00 & 656.37 & $-5.7$ \\
 &          & Exp     & ---  & ---         & 15.521 & 5.663 & 7.916 & 90.00 & 90.00 & 90.00 & 695.78 & ---    \\
\cmidrule(lr){2-13}
 & coumar23 & DFT-Opt & 1.51 & $-117.840$ & 15.278 & 5.491 & 7.822 & 90.00 & 90.00 & 90.00 & 656.19 & $-3.2$ \\
 &          & Exp     & ---  & ---         & 15.496 & 5.617 & 7.785 & 90.00 & 90.00 & 90.00 & 677.69 & ---    \\
\midrule

% ----------------------------------------------------------------
% Benzoic acid (BENZAC) -- 6 polymorphs, \multirow{12}
% ----------------------------------------------------------------
\multirow{12}{*}{\shortstack[l]{\textit{Benzoic acid} \\ \cite{benzac_poly01,benzac_poly02}}}
 & benzac01 & DFT-Opt & 0.94 & $-100.387$ & 5.346 & 4.989 & 21.604 & 90.00 & 98.61 & 90.00 & 569.73 & $-8.0$ \\
 &          & Exp     & ---  & ---         & 5.510 & 5.157 & 21.973 & 90.00 & 97.41 & 90.00 & 619.15 & ---    \\
\cmidrule(lr){2-13}
 & benzac02 & DFT-Opt & 1.04 & $-100.386$ & 5.381 & 5.003 & 21.426 & 90.00 & 98.69 & 90.00 & 570.11 & $-7.1$ \\
 &          & Exp     & ---  & ---         & 5.500 & 5.128 & 21.950 & 90.00 & 97.37 & 90.00 & 613.96 & ---    \\
\cmidrule(lr){2-13}
 & benzac12 & DFT-Opt & 0.00 & $-100.396$ & 5.348 & 4.985 & 21.466 & 90.00 & 95.63 & 90.00 & 569.56 & $-2.9$ \\
 &          & Exp     & ---  & ---         & 5.415 & 5.039 & 21.630 & 90.00 & 96.14 & 90.00 & 586.82 & ---    \\
\cmidrule(lr){2-13}
 & benzac20 & DFT-Opt & 0.92 & $-100.387$ & 5.345 & 4.989 & 21.476 & 90.00 & 95.64 & 90.00 & 569.96 & $-4.0$ \\
 &          & Exp     & ---  & ---         & 5.438 & 5.060 & 21.706 & 90.00 & 96.18 & 90.00 & 593.74 & ---    \\
\cmidrule(lr){2-13}
 & benzac22 & DFT-Opt & 1.20 & $-100.383$ & 5.391 & 4.996 & 21.289 & 90.00 & 95.85 & 90.00 & 570.49 & $-7.2$ \\
 &          & Exp     & ---  & ---         & 5.502 & 5.137 & 21.927 & 90.00 & 97.05 & 90.00 & 615.06 & ---    \\
\cmidrule(lr){2-13}
 & benzac23 & DFT-Opt & 1.10 & $-100.385$ & 4.992 & 5.367 & 21.371 & 95.72 & 89.98 & 89.99 & 569.79 & $-8.5$ \\
 &          & Exp     & ---  & ---         & 5.156 & 5.521 & 22.049 & 97.19 & 90.00 & 90.00 & 622.72 & ---    \\
\midrule

% ----------------------------------------------------------------
% Benzamide (BZAMID) -- 10 polymorphs, \multirow{20}
% ----------------------------------------------------------------
\multirow{20}{*}{\shortstack[l]{\textit{Benzamide} \\ \cite{bzamid_poly01,bzamid_poly02,bzamid_poly03,bzamid_poly04}}}
 & bzamid01 & DFT-Opt & 0.30 & $-104.609$ & 5.496 & 4.991 & 21.167 & 90.00 & 88.50 & 90.00 & 580.42 & $-7.0$ \\
 &          & Exp     & ---  & ---         & 5.607 & 5.046 & 22.053 & 90.00 & 90.66 & 90.00 & 623.90 & ---    \\
\cmidrule(lr){2-13}
 & bzamid07 & DFT-Opt & 1.38 & $-104.598$ & 5.551 & 5.072 & 20.933 & 90.00 & 88.75 & 90.00 & 589.19 & $-5.8$ \\
 &          & Exp     & ---  & ---         & 5.609 & 5.040 & 22.117 & 90.00 & 90.64 & 90.00 & 625.23 & ---    \\
\cmidrule(lr){2-13}
 & bzamid08 & DFT-Opt & 0.11 & $-104.611$ & 5.012 & 5.362 & 22.172 & 90.00 & 102.36 & 90.00 & 581.92 & $-7.3$ \\
 &          & Exp     & ---  & ---         & 5.055 & 5.514 & 22.956 & 90.00 & 101.29 & 90.00 & 627.50 & ---    \\
\cmidrule(lr){2-13}
 & bzamid11 & DFT-Opt & 0.08 & $-104.612$ & 5.010 & 5.363 & 21.669 & 90.00 & 90.72 & 90.00 & 582.27 & $-3.2$ \\
 &          & Exp     & ---  & ---         & 5.052 & 5.424 & 21.949 & 90.00 & 90.91 & 90.00 & 601.41 & ---    \\
\cmidrule(lr){2-13}
 & bzamid12 & DFT-Opt & 0.02 & $-104.612$ & 5.010 & 5.366 & 22.295 & 90.00 & 103.71 & 90.00 & 582.31 & $-4.9$ \\
 &          & Exp     & ---  & ---         & 5.059 & 5.461 & 22.833 & 90.00 & 103.85 & 90.00 & 612.46 & ---    \\
\cmidrule(lr){2-13}
 & bzamid14 & DFT-Opt & 0.40 & $-104.608$ & 5.498 & 4.992 & 21.145 & 90.00 & 91.52 & 90.00 & 580.16 & $-4.8$ \\
 &          & Exp     & ---  & ---         & 5.574 & 5.036 & 21.702 & 90.00 & 90.50 & 90.00 & 609.20 & ---    \\
\cmidrule(lr){2-13}
 & bzamid15 & DFT-Opt & 0.34 & $-104.609$ & 5.495 & 4.990 & 21.175 & 90.00 & 91.49 & 90.00 & 580.35 & $-4.8$ \\
 &          & Exp     & ---  & ---         & 5.567 & 5.032 & 21.754 & 90.00 & 90.18 & 90.00 & 609.32 & ---    \\
\cmidrule(lr){2-13}
 & bzamid16 & DFT-Opt & 0.00 & $-104.612$ & 5.010 & 5.364 & 22.310 & 90.00 & 103.72 & 90.00 & 582.42 & $-4.1$ \\
 &          & Exp     & ---  & ---         & 5.048 & 5.447 & 22.759 & 90.00 & 103.86 & 90.00 & 607.54 & ---    \\
\cmidrule(lr){2-13}
 & bzamid17 & DFT-Opt & 0.16 & $-104.611$ & 5.011 & 5.364 & 22.284 & 90.00 & 103.71 & 90.00 & 581.93 & $-3.3$ \\
 &          & Exp     & ---  & ---         & 5.042 & 5.430 & 22.639 & 90.00 & 103.80 & 90.00 & 601.95 & ---    \\
\cmidrule(lr){2-13}
 & bzamid18 & DFT-Opt & 0.36 & $-104.609$ & 5.492 & 4.989 & 21.185 & 90.00 & 91.48 & 90.00 & 580.27 & $-4.7$ \\
 &          & Exp     & ---  & ---         & 5.560 & 5.040 & 21.720 & 90.00 & 90.10 & 90.00 & 608.65 & ---    \\

\end{longtable}
\end{landscape}

\subsection{Dataset Energy-Force Distributions}

\begin{figure}[H]
\centering
\includegraphics[width=\textwidth]{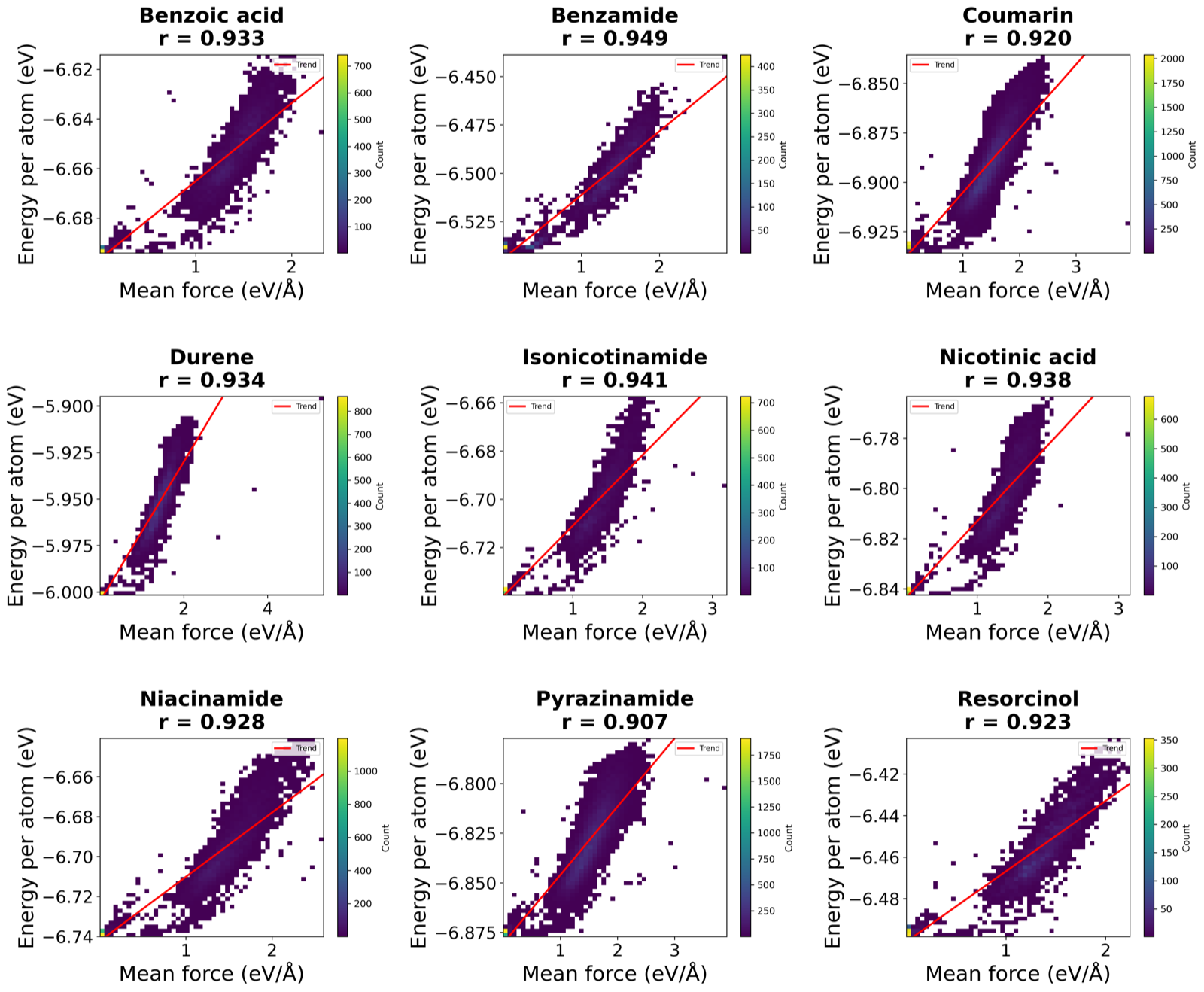}
\caption{Energy-force density distributions for all nine molecular crystal datasets. Each panel shows the 2D histogram of energy per atom versus mean atomic force magnitude, with color indicating configuration count. Red lines show linear fits, and Pearson correlation coefficients ($r$) are indicated.}
\label{fig:energy_force_density}
\end{figure}

Figure \ref{fig:energy_force_density} shows 2D density plots of energy per atom versus mean atomic force magnitude for each compound. All datasets exhibit strong positive correlations ($r > 0.9$), consistent with the expected physical relationship between configuration energy and atomic forces, confirming the internal consistency of the DFT reference data across both near- and off-equilibrium regions.

\subsection{MACE Model Training}

\subsubsection{Foundation Model and Architecture}

\subsubsection{Training Hyperparameters}
\begin{table}[H]
\centering
\caption{MACE fine-tuning hyperparameters used for all nine systems.}
\begin{tabular}{@{}ll@{}}
\toprule
\textbf{Parameter} & \textbf{Value} \\
\midrule
\multicolumn{2}{l}{\textit{Architecture}} \\
Foundation model & MACE-MH-1 (\texttt{omol} head) \\
Hidden irreducible representations & $512 \times 0e + 512 \times 1o$ \\
Correlation order & 3 (body order: 4) \\
Spherical harmonics up to & $\ell = 3$ \\
Radial cutoff ($r_{\max}$) & 6.0~\AA \\
Total receptive field & 12.0~\AA \\
Number of interaction layers & 2 \\
Radial basis transform & Agnesi \\
\midrule
\multicolumn{2}{l}{\textit{Training Protocol}} \\
Batch size & 10 \\
Optimizer & Adam \\
Initial learning rate & $1 \times 10^{-3}$ \\
Weight decay & $5 \times 10^{-7}$ \\
Exponential moving average (EMA) decay & 0.99 \\
\midrule
\multicolumn{2}{l}{\textit{Loss Function Weights (Epochs 0--100)}} \\
Energy weight & 100.0 \\
Forces weight & 100.0 \\
\midrule
\multicolumn{2}{l}{\textit{Stage Two (Epochs 100--300)}} \\
Stage two learning rate & $1 \times 10^{-4}$ \\
Energy weight & 100.0 \\
Forces weight & 100.0 \\
\midrule
\multicolumn{2}{l}{\textit{Data Processing}} \\
Train/validation split & 85/15 \\
Data format & HDF5 \\
\bottomrule
\end{tabular}
\label{tab:hyperparameters}
\end{table}
\subsubsection{Isolated Atom Energies}

The following isolated atom reference energies (E0s) were used for all systems containing H, C, N, and O. Each was computed with VASP for a single atom in a large cubic box, using the PBE functional (ENCUT = 850~eV, $\Gamma$-point only, \texttt{PREC = High}, EDIFF = $10^{-6}$~eV), with spin polarization enabled for open-shell atoms (C, N, O) following Hund's first rule.

\begin{table}[H]
\centering
\caption{Isolated Atom Reference Energies}
\begin{tabular}{@{}cc@{}}
\toprule
\textbf{Element} & \textbf{Energy (eV)} \\
\midrule
H (Z=1) & $-0.00292689$ \\
C (Z=6) & $-1.24732389$ \\
N (Z=7) & $-3.12548797$ \\
O (Z=8) & $-1.5314622$ \\
\bottomrule
\end{tabular}
\label{tab:E0s}
\end{table}

\subsection{Correlation Plots and training time}

\begin{figure}[H]
    \centering
    \includegraphics[width=\linewidth, keepaspectratio]{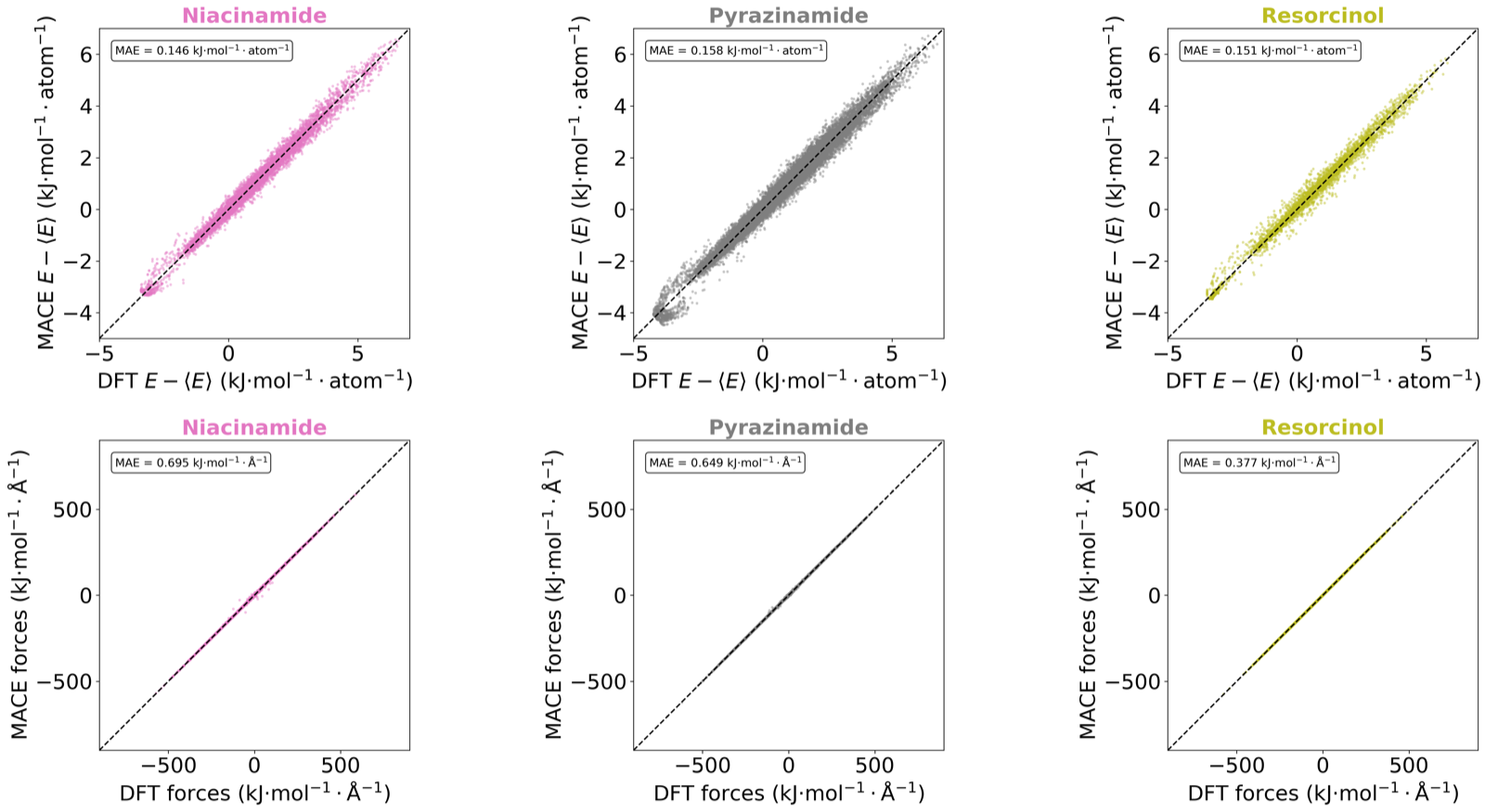}
    \caption{Force and energy correlation plots for the second set of fine-tuned MACE models (EHOWIH, NICOAC, NICOAM, PYRIZIN). Each panel shows MACE-predicted values against DFT reference values. The dashed line indicates perfect agreement.}
    \label{fig:forces_correlations_02}
\end{figure}

\begin{figure}[H]
    \centering
    \includegraphics[width=\linewidth, keepaspectratio]{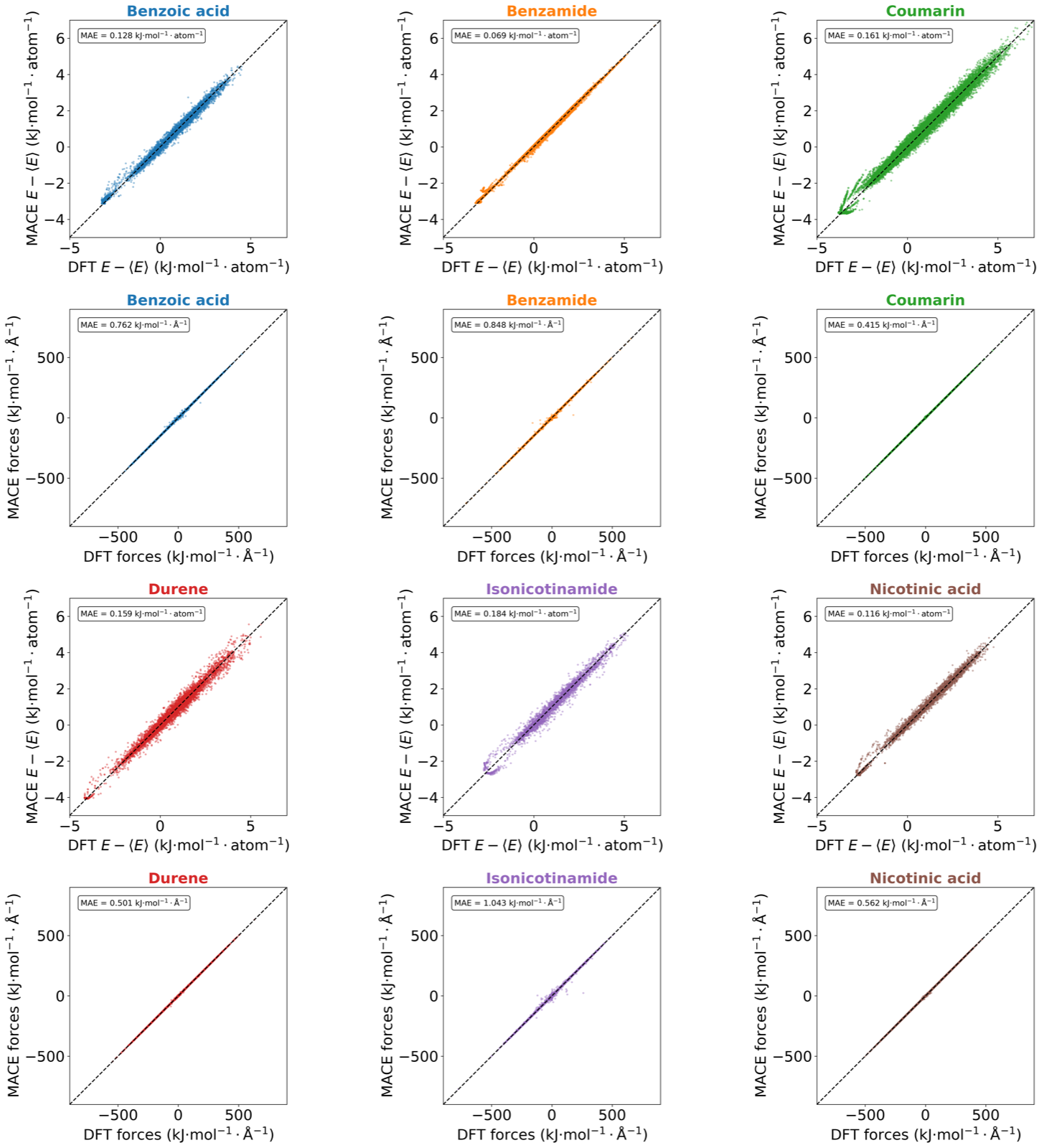}
    \caption{Force and energy correlation plots for the first set of fine-tuned MACE models (BENZAC, BZAMID, COUMAR, DURENE). Each panel shows MACE-predicted values against DFT reference values. The dashed line indicates perfect agreement.}
    \label{fig:forces_correlations_01}
\end{figure}

\begin{table}[htbp]
\centering
\caption{Training dataset sizes and approximate wall-clock training times for all nine MC-MLIP models. Training times are estimated from MACE training log timestamps. All models were trained on a single NVIDIA A100 GPU.}
\label{tab:training_times}
\begin{tabular}{@{}lcc@{}}
\toprule
\textbf{Compound} 
    & \textbf{Training configs} 
    & \textbf{Total time (h)} \\
\midrule
Benzoic acid    & 6,195  & $\sim$34  \\
Benzamide       & 8,880  & $\sim$55  \\
Coumarin        & 23,985 & $\sim$115 \\
Durene          & 6,045  & $\sim$28  \\
Isonicotinamide & 3,465  & $\sim$25  \\
Nicotinic acid  & 4,395  & $\sim$22  \\
Niacinamide     & 8,250  & $\sim$35  \\
Pyrazinamide    & 25,845 & $\sim$107 \\
Resorcinol      & 4,035  & $\sim$22  \\
\bottomrule
\end{tabular}
\end{table}

\begin{figure}[htbp]
    \centering
    \includegraphics[width=0.8\textwidth]{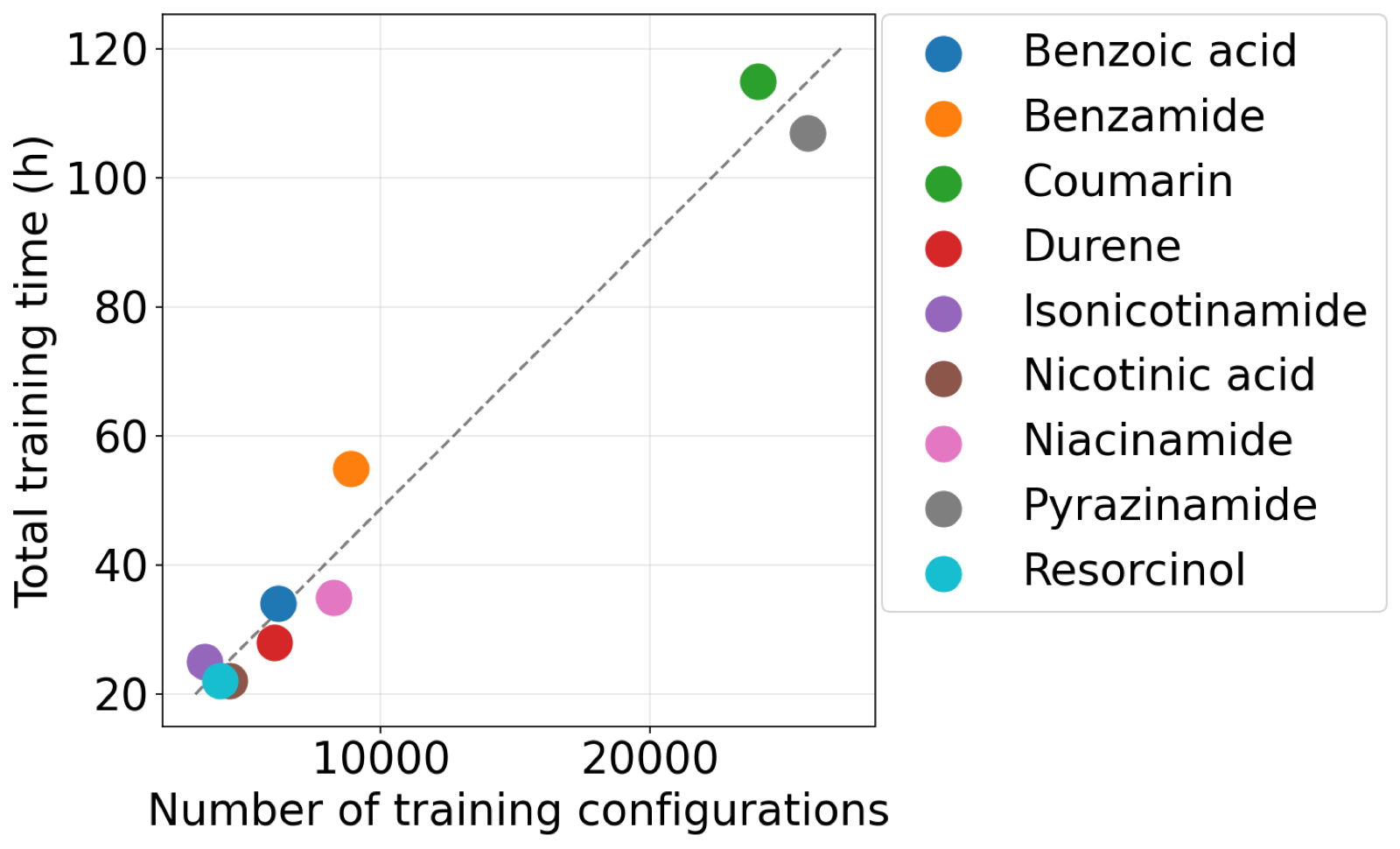}
    \caption{Number of training configurations versus total wall-clock training time for the nine MC-MLIP models. The dashed line shows a linear fit to the data. All models were trained on a single NVIDIA A100 GPU.}
    \label{fig:training_time}
\end{figure}

\section{DFT-Optimized Structures for All Polymorphs}

This section provides comprehensive structural data for all polymorphs studied in this work. For each polymorph, we present both the DFT-optimized structure (at 0~K) and the experimental structure (typically measured near 298~K) for comparison.

\subsection{MACE-Optimized Polymorph Energetics}

\begin{table}[H]
\centering
\caption{Correspondence between Roman numeral labels used in 
Figure~\ref{fig:elatt_mace_omol_dft} and CSD reference codes for each 
polymorph. Polymorphs are ordered by increasing DFT relative lattice energy 
($\Delta E_{\mathrm{latt}}$).}
\label{tab:roman_to_csd}
\begin{tabular}{clclclcl}
\toprule
\multicolumn{2}{c}{\textbf{Coumarin}} &
\multicolumn{2}{c}{\textbf{Isonicotinamide}} &
\multicolumn{2}{c}{\textbf{Niacinamide}} &
\multicolumn{2}{c}{\textbf{Pyrazinamide}} \\
\cmidrule(lr){1-2}\cmidrule(lr){3-4}\cmidrule(lr){5-6}\cmidrule(lr){7-8}
Label & CSD Code & Label & CSD Code & Label & CSD Code & Label & CSD Code \\
\midrule
I     & COUMAR20 & I   & EHOWIH05 & I   & NICOAM01 & I    & PYRZIN01 \\
II    & COUMAR13 & II  & EHOWIH07 & II  & NICOAM13 & II   & PYRZIN14 \\
III   & COUMAR21 & III & EHOWIH01 & III & NICOAM05 & III  & PYRZIN23 \\
IV    & COUMAR22 & IV  & EHOWIH   & IV  & NICOAM08 & IV   & PYRZIN18 \\
V     & COUMAR12 &     &          & V   & NICOAM07 & V    & PYRZIN29 \\
VI    & COUMAR01 &     &          & VI  & NICOAM18 & VI   & PYRZIN02 \\
VII   & COUMAR02 &     &          & VII & NICOAM17 & VII  & PYRZIN16 \\
VIII  & COUMAR19 &     &          &     &          & VIII & PYRZIN25 \\
IX    & COUMAR23 &     &          &     &          & IX   & PYRZIN21 \\
X     & COUMAR11 &     &          &     &          & X    & PYRZIN22 \\
XI    & COUMAR18 &     &          &     &          & XI   & PYRZIN15 \\
      &          &     &          &     &          & XII  & PYRZIN27 \\
\bottomrule
\end{tabular}
\end{table}

\begin{figure}[H]
\centering
\includegraphics[width=0.85\textwidth]{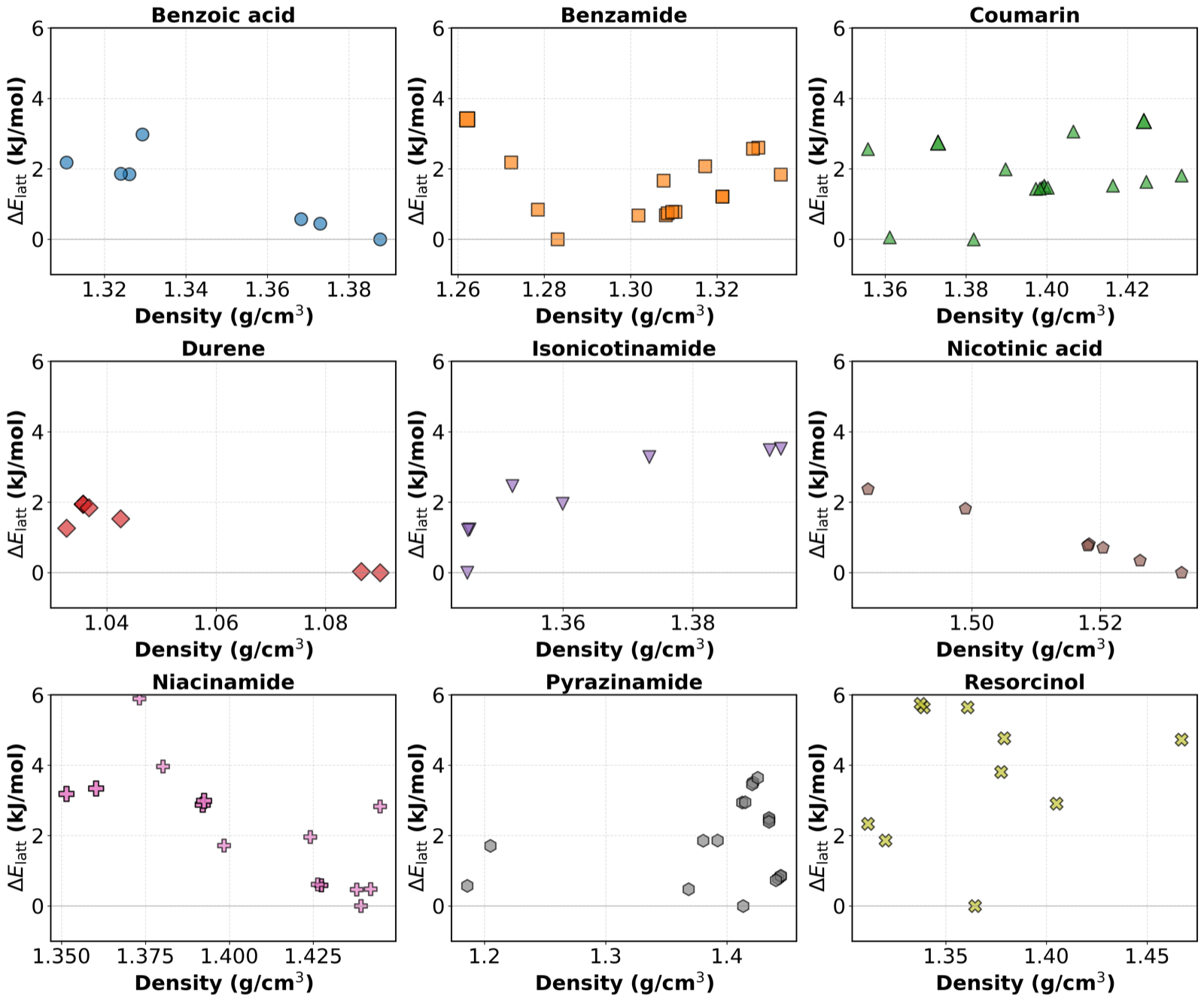}
\caption{Relative lattice energy ($\Delta E_{\mathrm{latt}}$) versus density for MACE-optimized polymorphs. Each point represents a distinct polymorph, with the most stable structure (lowest energy) set as the reference ($\Delta E = 0$).}
\label{fig:delta_elatt_density_single}
\end{figure}

To validate the transferability of the fine-tuned MACE models beyond the training data, we performed geometry optimizations on all experimental polymorphs from the Cambridge Structural Database using the trained potentials. Figure~\ref{fig:delta_elatt_density_single} shows the relative lattice energy ($\Delta E_{\mathrm{latt}}$) versus density for each system, where energies are referenced to the most stable polymorph (set to zero).

\newpage
\newgeometry{margin=0.5in}
\begin{table}[t]
\centering
\small
\caption{Per-system accuracy of the relative lattice energy $\Delta E_\mathrm{latt}$ against the PBE+D4 reference, referenced to the DFT-most-stable polymorph. Spread: DFT energy range of the system (kJ\,mol$^{-1}$); MAE and RMSE in kJ\,mol$^{-1}$ per molecule; $\tau$, $\rho$, and $r$: Kendall, Spearman, and Pearson correlations with the DFT energies/ordering; penalty: $\Delta E_\mathrm{latt}$ of the predicted most stable polymorph above the DFT minimum (kJ\,mol$^{-1}$); min: most stable polymorph correctly identified.}
\label{tab:deltaElatt_all9}
\begin{tabular}{llccccccccc}
\toprule
System & Model & spread & MAE & RMSE & MAE$_\mathrm{pair}$ & $\tau$ & $\rho$ & $r$ & Penalty & min \\
\midrule
Benzoic acid & \textbf{MolCryst} & 2.53 & 1.29 & 1.57 & 1.26 & 0.33 & 0.40 & 0.34 & 0.00 & \checkmark \\
 & \texttt{omol} & 2.53 & 1.62 & 1.88 & 1.27 & $-0.80$ & $-0.90$ & $-0.85$ & 2.53 & $\times$ \\
 & Orb & 2.53 & 1.60 & 1.89 & 1.37 & $-0.40$ & $-0.50$ & $-0.65$ & 2.53 & $\times$ \\
 & UMA & 2.53 & 1.29 & 1.56 & 1.20 & $-0.40$ & $-0.40$ & $-0.21$ & 2.53 & $\times$ \\
\midrule
Benzamide & \textbf{MolCryst} & 2.60 & 0.79 & 0.90 & 0.52 & 0.64 & 0.81 & 0.96 & 0.77 & \checkmark \\
 & \texttt{omol} & 2.60 & 1.04 & 1.27 & 0.93 & 0.50 & 0.67 & 0.85 & 0.00 & \checkmark \\
 & UMA & 2.60 & 1.22 & 1.46 & 1.02 & 0.71 & 0.88 & 0.97 & 0.00 & \checkmark \\
\midrule
Coumarin & \textbf{MolCryst} & 1.99 & 0.31 & 0.34 & 0.16 & 0.72 & 0.86 & 0.98 & 0.00 & \checkmark \\
 & \texttt{omol} & 1.99 & 1.27 & 1.43 & 1.13 & 0.72 & 0.87 & 0.39 & 0.06 & \checkmark \\
 & Orb & 1.99 & 8.00 & 8.50 & 2.75 & $-0.46$ & $-0.57$ & $-0.84$ & 1.99 & \checkmark \\
 & UMA & 1.99 & 1.02 & 1.43 & 1.21 & 0.76 & 0.88 & 0.46 & 0.00 & \checkmark \\
\midrule
Durene & \textbf{MolCryst} & 1.50 & 0.94 & 1.15 & 1.01 & $-0.33$ & $-0.50$ & $-0.45$ & 1.23 & \checkmark \\
 & \texttt{omol} & 1.50 & 0.90 & 1.11 & 0.96 & 0.33 & 0.50 & 0.26 & 1.23 & \checkmark \\
 & UMA & 1.50 & 0.92 & 1.13 & 1.01 & $-1.00$ & $-1.00$ & $-1.00$ & 1.50 & \checkmark \\
\midrule
Isonicotinamide & \textbf{MolCryst} & 3.52 & 0.23 & 0.28 & 0.23 & 0.67 & 0.80 & 1.00 & 0.00 & \checkmark \\
 & \texttt{omol} & 3.52 & 4.95 & 5.71 & 3.37 & 0.00 & $-0.20$ & $-1.00$ & 3.28 & $\times$ \\
 & Orb & 3.52 & 4.51 & 5.20 & 3.08 & $-0.33$ & $-0.40$ & $-1.00$ & 3.28 & $\times$ \\
 & UMA & 3.52 & 1.22 & 1.41 & 0.89 & 0.67 & 0.80 & 1.00 & 0.00 & \checkmark \\
\midrule
Nicotinic acid & \textbf{MolCryst} & 1.94 & 1.97 & 2.32 & 1.70 & $-0.33$ & $-0.40$ & $-0.74$ & 0.76 & \checkmark \\
 & \texttt{omol} & 1.94 & 0.98 & 1.21 & 1.06 & $-0.67$ & $-0.80$ & $-0.11$ & 0.87 & \checkmark \\
 & Orb & 1.94 & 0.95 & 1.18 & 1.02 & $-0.33$ & $-0.40$ & $-0.11$ & 0.87 & \checkmark \\
 & UMA & 1.94 & 0.91 & 1.14 & 1.01 & $-0.33$ & $-0.60$ & $-0.02$ & 0.87 & \checkmark \\
\midrule
Niacinamide & \textbf{MolCryst} & 5.42 & 0.64 & 0.95 & 1.14 & 0.78 & 0.92 & 0.87 & 0.00 & \checkmark \\
 & \texttt{omol} & 5.42 & 3.43 & 5.78 & 5.41 & $-0.39$ & $-0.52$ & $-0.79$ & 5.42 & $\times$ \\
 & Orb & 5.42 & 4.03 & 5.97 & 5.33 & $-0.59$ & $-0.77$ & $-0.90$ & 5.42 & $\times$ \\
 & UMA & 5.42 & 1.07 & 1.80 & 1.76 & 0.49 & 0.68 & 0.61 & 0.00 & \checkmark \\
\midrule
Pyrazinamide & \textbf{MolCryst} & 3.64 & 0.95 & 1.52 & 1.53 & 0.43 & 0.45 & 0.54 & 0.00 & \checkmark \\
 & \texttt{omol} & 3.64 & 2.19 & 2.62 & 2.72 & $-0.10$ & $-0.20$ & $-0.25$ & 0.48 & \checkmark \\
 & Orb & 3.64 & 1.38 & 2.04 & 2.12 & 0.36 & 0.28 & 0.28 & 0.00 & \checkmark \\
 & UMA & 3.64 & 1.24 & 1.43 & 1.17 & 0.41 & 0.53 & 0.58 & 2.50 & $\times$ \\
\midrule
Resorcinol & \textbf{MolCryst} & 5.64 & 1.68 & 1.96 & 1.37 & 0.67 & 0.80 & 0.91 & 1.86 & \checkmark \\
 & \texttt{omol} & 5.64 & 1.55 & 1.82 & 1.29 & 0.00 & 0.20 & 0.89 & 2.33 & $\times$ \\
 & Orb & 5.64 & 1.23 & 2.21 & 2.43 & 0.00 & 0.20 & 0.31 & 0.00 & \checkmark \\
 & UMA & 5.64 & 2.05 & 2.50 & 2.07 & 0.00 & 0.20 & 0.89 & 2.33 & $\times$ \\
\bottomrule
\multicolumn{11}{l}{\footnotesize Orb rows for durene and benzamide are omitted: only two durene polymorphs converged, for which}\\
\multicolumn{11}{l}{\footnotesize rank correlations are undefined; three benzamide relaxations failed to reach the convergence criterion.}
\end{tabular}
\end{table}
\restoregeometry

\begin{landscape}
\begin{table}[htbp]
\centering
\caption{Crystal polymorph data: density, lattice energy relative to the most stable form, and number of molecules in the unit cell. Bold entries indicate the most stable polymorph per system ($\Delta E_\mathrm{latt} = 0$). Green rows indicate polymorphs not included in the training set.}
\label{tab:polymorphs}

\begin{minipage}[t]{0.48\linewidth}
\centering
\begin{tabular}{lccc}
\toprule
CSD Code & $\rho$ (g cm$^{-3}$) & $\Delta E_\mathrm{latt}$ (kJ mol$^{-1}$) & $Z$ \\
\midrule
% BENZAC
BENZAC01 & 1.3293 & 2.98 & 4 \\
BENZAC02 & 1.3261 & 1.84 & 4 \\
BENZAC12 & 1.3877 & 0.00 & 4 \\
BENZAC20 & 1.3730 & 0.45 & 4 \\
\rowcolor{green!20} BENZAC21 & 1.3683 & 0.57 & 4 \\
BENZAC22 & 1.3240 & 1.86 & 4 \\
BENZAC23 & 1.3107 & 2.18 & 4 \\
\midrule
% BZAMID
BZAMID01 & 1.2831 & 0.00 & 4 \\
\rowcolor{green!20} BZAMID02 & 1.3348 & 1.84 & 4 \\
\rowcolor{green!20} BZAMID03 & 1.3213 & 1.21 & 4 \\
\rowcolor{green!20} BZAMID04 & 1.3213 & 1.21 & 4 \\
\rowcolor{green!20} BZAMID05 & 1.3104 & 0.78 & 4 \\
BZAMID07 & 1.2723 & 2.18 & 4 \\
BZAMID08 & 1.2785 & 0.84 & 4 \\
BZAMID11 & 1.3296 & 2.60 & 4 \\
BZAMID12 & 1.3077 & 1.66 & 4 \\
\rowcolor{green!20} BZAMID13 & 1.2622 & 3.41 & 32 \\
BZAMID14 & 1.3081 & 0.68 & 4 \\
BZAMID15 & 1.3086 & 0.74 & 4 \\
BZAMID16 & 1.3172 & 2.07 & 4 \\
BZAMID17 & 1.3283 & 2.57 & 4 \\
BZAMID18 & 1.3096 & 0.77 & 4 \\
\rowcolor{green!20} BZAMID   & 1.3018 & 0.68 & 4 \\
\bottomrule
\end{tabular}
\end{minipage}
\hfill
\begin{minipage}[t]{0.48\linewidth}
\centering
\begin{tabular}{lccc}
\toprule
CSD Code & $\rho$ (g cm$^{-3}$) & $\Delta E_\mathrm{latt}$ (kJ mol$^{-1}$) & $Z$ \\
\midrule
% COUMAR
COUMAR01 & 1.3996 & 1.47 & 4 \\
COUMAR02 & 1.3988 & 1.47 & 4 \\
\rowcolor{green!20} COUMAR10 & 1.3993 & 1.52 & 4 \\
COUMAR11 & 1.4334 & 1.81 & 4 \\
COUMAR12 & 1.4003 & 1.46 & 4 \\
COUMAR13 & 1.3611 & 0.06 & 2 \\
COUMAR14 & 1.4066 & 3.06 & 8 \\
\rowcolor{green!20} COUMAR15 & 1.3557 & 2.56 & 8 \\
\rowcolor{green!20} COUMAR16 & 1.4240 & 3.36 & 12 \\
\rowcolor{green!20} COUMAR17 & 1.3731 & 2.75 & 12 \\
COUMAR18 & 1.3898 & 1.99 & 4 \\
COUMAR19 & 1.4163 & 1.52 & 4 \\
COUMAR20 & 1.3819 & 0.00 & 2 \\
COUMAR21 & 1.3973 & 1.44 & 4 \\
COUMAR22 & 1.3983 & 1.44 & 4 \\
COUMAR23 & 1.4246 & 1.63 & 4 \\
\midrule
% DURENE
DURENE01 & 1.0326 & 1.26 & 2 \\
\rowcolor{green!20} DURENE02 & 1.0357 & 1.94 & 2 \\
\rowcolor{green!20} DURENE03 & 1.0357 & 1.94 & 2 \\
\rowcolor{green!20} DURENE04 & 1.0357 & 1.94 & 2 \\
DURENE05 & 1.0425 & 1.53 & 2 \\
DURENE06 & 1.0865 & 0.03 & 2 \\
\rowcolor{green!20} DURENE07 & 1.0900 & 0.00 & 2 \\
\rowcolor{green!20} DURENE   & 1.0368 & 1.84 & 2 \\
\bottomrule
\end{tabular}
\end{minipage}

\end{table}
\end{landscape}

% ============================================================
% PAGE 2:  Left = EHOWIH + NICOAC + NICOAM   |   Right = PYRZIN + RESORA
% ============================================================
\begin{landscape}
\begin{table}[htbp]
\centering
\addtocounter{table}{-1}
\caption*{Table~\thetable{} -- \textit{continued}}

\begin{minipage}[t]{0.45\linewidth}
\centering
\begin{tabular}{lccc}
\toprule
CSD Code & $\rho$ (g cm$^{-3}$) & $\Delta E_\mathrm{latt}$ (kJ mol$^{-1}$) & $Z$ \\
\midrule
% EHOWIH
EHOWIH01 & 1.3919 & 3.48 & 4 \\
\rowcolor{green!20} EHOWIH02 & 1.3455 & 1.23 & 8 \\
EHOWIH03 & 1.3522 & 2.46 & 8 \\
\rowcolor{green!20} EHOWIH04 & 1.3599 & 1.96 & 6 \\
EHOWIH05 & 1.3452 & 0.00 & 4 \\
\rowcolor{green!20} EHOWIH06 & 1.3453 & 1.21 & 8 \\
EHOWIH07 & 1.3733 & 3.28 & 4 \\
EHOWIH   & 1.3936 & 3.52 & 4 \\
\midrule
% NICOAC
NICOAC01 & 1.5182 & 0.87 & 4 \\
\rowcolor{green!20} NICOAC02 & 1.5180 & 0.83 & 4 \\
NICOAC05 & 1.4990 & 1.94 & 4 \\
\rowcolor{green!20} NICOAC06 & 1.4838 & 2.54 & 4 \\
NICOAC07 & 1.5326 & 0.00 & 4 \\
NICOAC08 & 1.5204 & 0.76 & 4 \\
\rowcolor{green!20} NICOAC09 & 1.5261 & 0.37 & 4 \\
\midrule
% NICOAM
NICOAM01 & 1.4421 & 0.47 & 4 \\
\rowcolor{green!20} NICOAM02 & 1.4274 & 0.59 & 4 \\
\rowcolor{green!20} NICOAM03 & 1.4274 & 0.59 & 4 \\
\rowcolor{green!20} NICOAM04 & 1.3515 & 3.18 & 16 \\
NICOAM05 & 1.4263 & 0.61 & 4 \\
\rowcolor{green!20} NICOAM06 & 1.4392 & 0.00 & 4 \\
NICOAM07 & 1.4449 & 2.82 & 2 \\
NICOAM08 & 1.3984 & 1.71 & 4 \\
\rowcolor{green!20} NICOAM09 & 1.3921 & 2.88 & 40 \\
NICOAM13 & 1.4380 & 0.47 & 4 \\
\rowcolor{green!20} NICOAM14 & 1.3603 & 3.34 & 16 \\
\rowcolor{green!20} NICOAM15 & 1.3925 & 2.98 & 16 \\
\rowcolor{green!20} NICOAM16 & 1.4241 & 1.96 & 8 \\
NICOAM17 & 1.3732 & 5.89 & 4 \\
NICOAM18 & 1.3803 & 3.96 & 4 \\
\bottomrule
\end{tabular}
\end{minipage}
\hfill
\begin{minipage}[t]{0.48\linewidth}
\centering
\begin{tabular}{lccc}
\toprule
CSD Code & $\rho$ (g cm$^{-3}$) & $\Delta E_\mathrm{latt}$ (kJ mol$^{-1}$) & $Z$ \\
\midrule
% PYRZIN
PYRZIN01 & 1.4135 & 0.00 & 4 \\
PYRZIN02 & 1.3922 & 1.86 & 2 \\
PYRZIN14 & 1.3685 & 0.48 & 4 \\
PYRZIN15 & 1.4215 & 3.51 & 4 \\
PYRZIN16 & 1.4349 & 2.45 & 2 \\
PYRZIN17 & 1.1858 & 0.58 & 2 \\
PYRZIN18 & 1.4442 & 0.83 & 4 \\
PYRZIN20 & 1.2049 & 1.71 & 2 \\
PYRZIN21 & 1.4126 & 2.94 & 4 \\
PYRZIN22 & 1.4207 & 3.45 & 4 \\
PYRZIN23 & 1.4422 & 0.78 & 4 \\
PYRZIN25 & 1.4347 & 2.50 & 2 \\
\rowcolor{green!20} PYRZIN26 & 1.4347 & 2.39 & 2 \\
PYRZIN27 & 1.4255 & 3.64 & 4 \\
\rowcolor{green!20} PYRZIN28 & 1.4151 & 2.94 & 4 \\
PYRZIN29 & 1.4445 & 0.86 & 4 \\
\rowcolor{green!20} PYRZIN30 & 1.4405 & 0.73 & 4 \\
\rowcolor{green!20} PYRZIN   & 1.3805 & 1.85 & 4 \\
\midrule
% RESORA
RESORA03 & 1.3201 & 1.86 & 4 \\
RESORA08 & 1.3391 & 5.64 & 4 \\
\rowcolor{green!20} RESORA09 & 1.3775 & 3.81 & 4 \\
RESORA13 & 1.3113 & 2.33 & 4 \\
RESORA15 & 1.3608 & 5.64 & 8 \\
RESORA16 & 1.3645 & 0.00 & 4 \\
RESORA24 & 1.3375 & 5.75 & 4 \\
\rowcolor{green!20} RESORA25 & 1.4050 & 2.90 & 4 \\
RESORA27 & 1.3790 & 4.76 & 8 \\
\rowcolor{green!20} RESORA31 & 1.4671 & 4.73 & 8 \\
\bottomrule
\end{tabular}
\end{minipage}

\end{table}
\end{landscape}

\section{Structural Integrity for All Polymorphs}
\subsection{P2 Orientational Order Parameter}
%\begin{figure}[H]
%    \centering
%    \includegraphics[width=1.0\linewidth]{SI-plots/p2/plot_p2_BENZAC.png}
%    \caption{P2 orientational order parameter as a function of simulation time for all BENZAC polymorphs at 300, 500, and 600~K.}
%    \label{fig:p2_benzac}
%\end{figure}
\begin{figure}[H]
    \centering
    \includegraphics[width=1.0\linewidth]{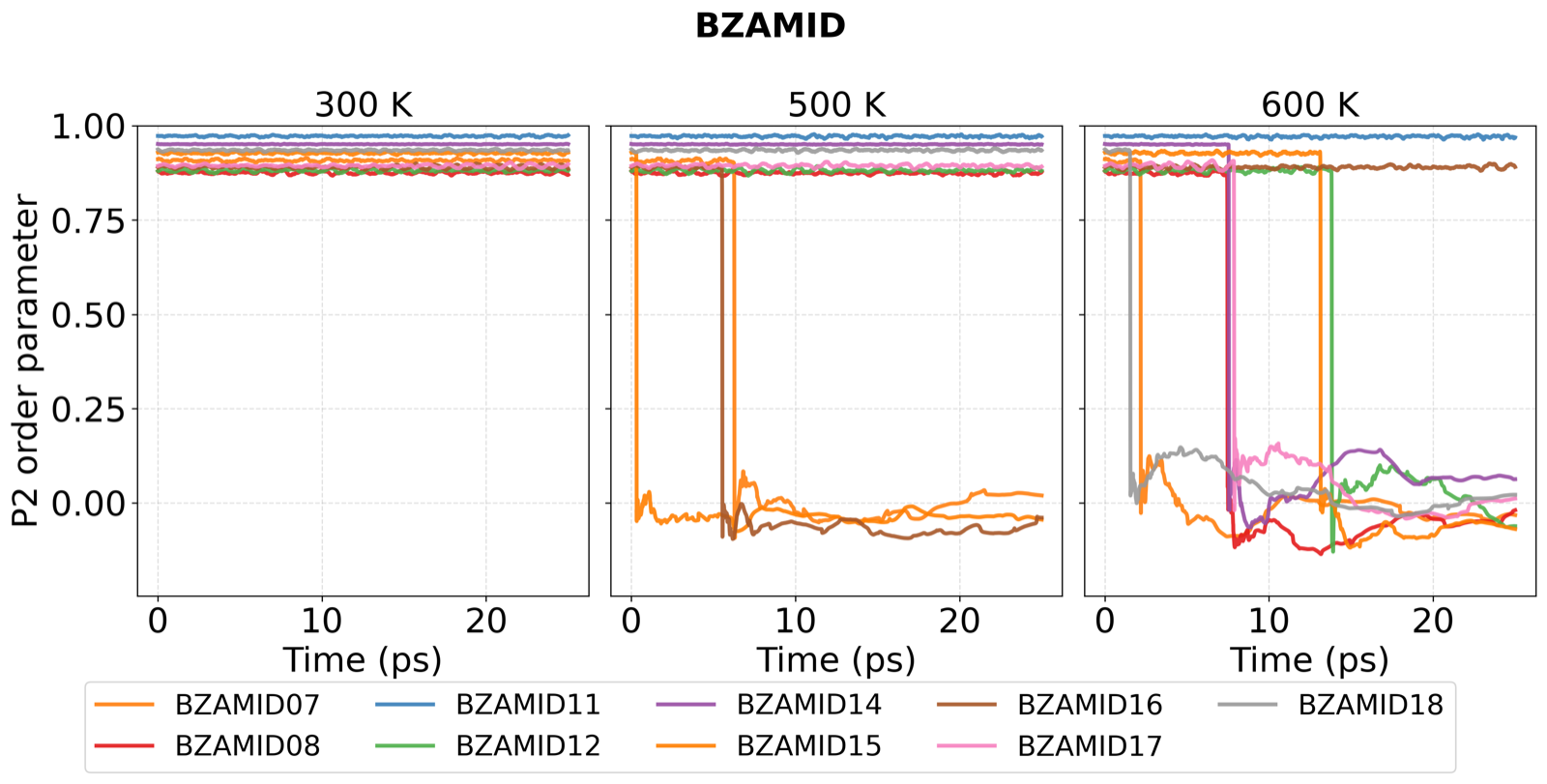}
    \caption{P2 orientational order parameter for all BZAMID polymorphs at 300, 500, and 600~K.}
    \label{fig:p2_bzamid}
\end{figure}
\begin{figure}[H]
    \centering
    \includegraphics[width=1.0\linewidth]{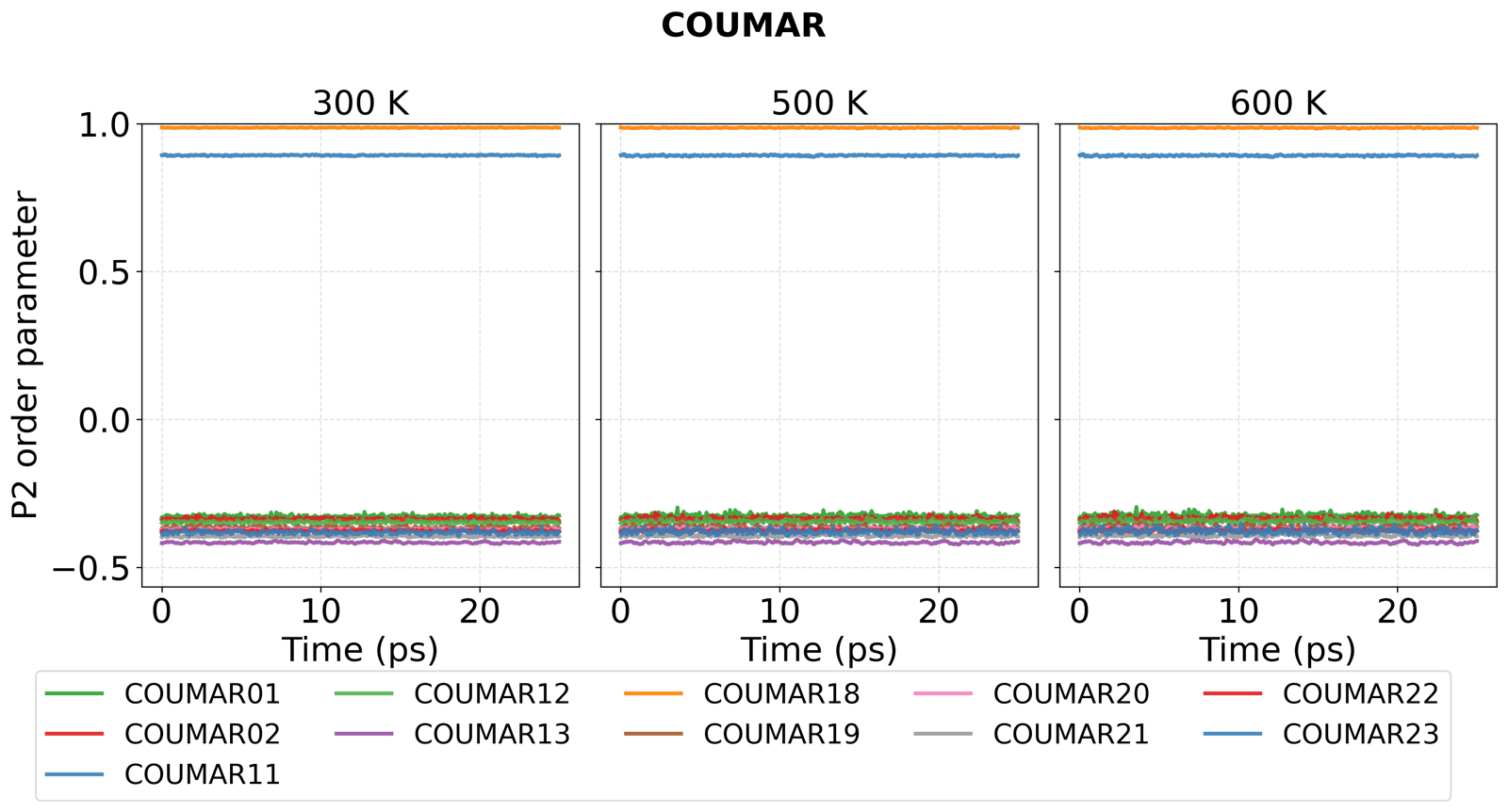}
    \caption{P2 orientational order parameter for all COUMAR polymorphs at 300, 500, and 600~K.}
    \label{fig:p2_coumar}
\end{figure}
\begin{figure}[H]
    \centering
    \includegraphics[width=1.0\linewidth]{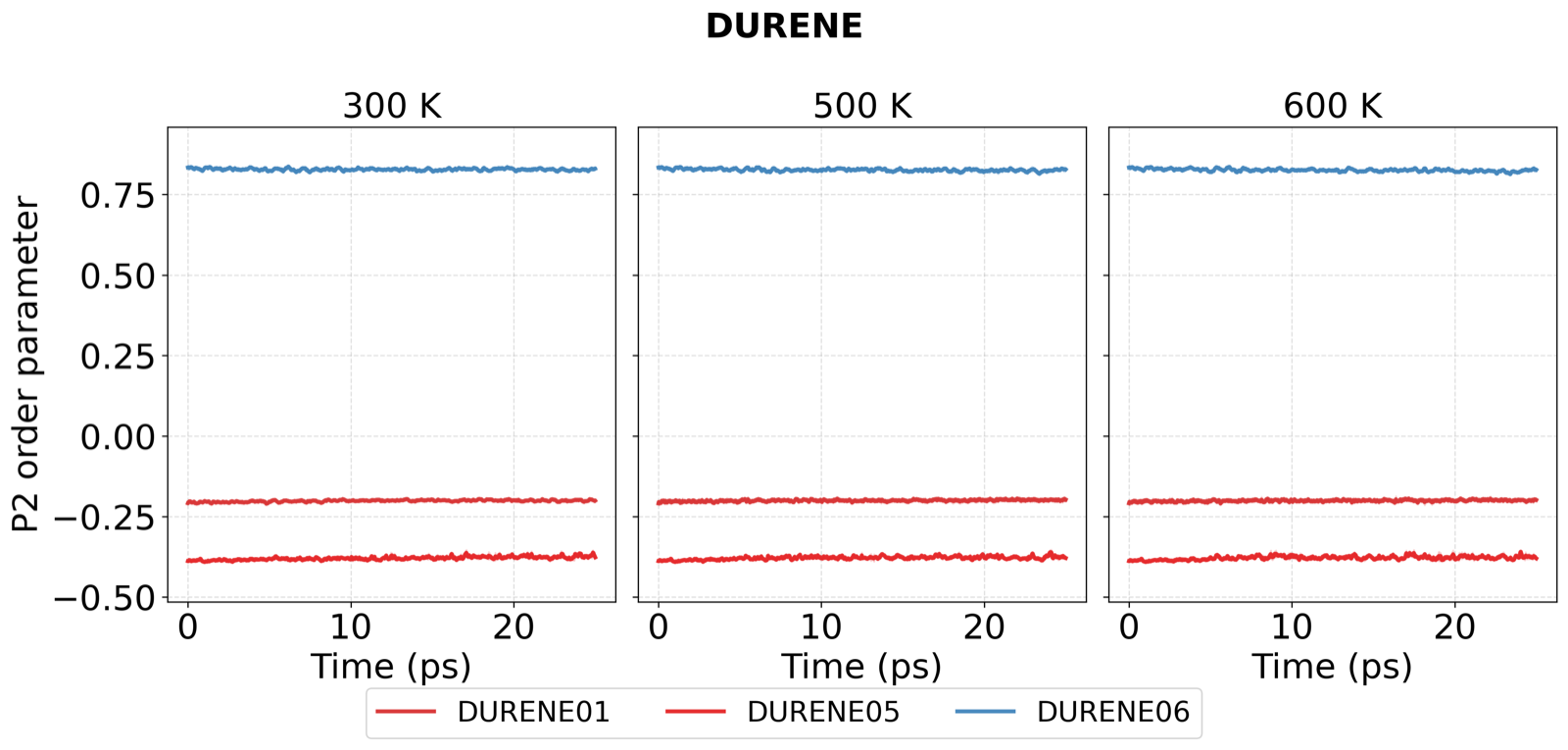}
    \caption{P2 orientational order parameter for all DURENE polymorphs at 300, 500, and 600~K.}
    \label{fig:p2_durene}
\end{figure}
\begin{figure}[H]
    \centering
    \includegraphics[width=1.0\linewidth]{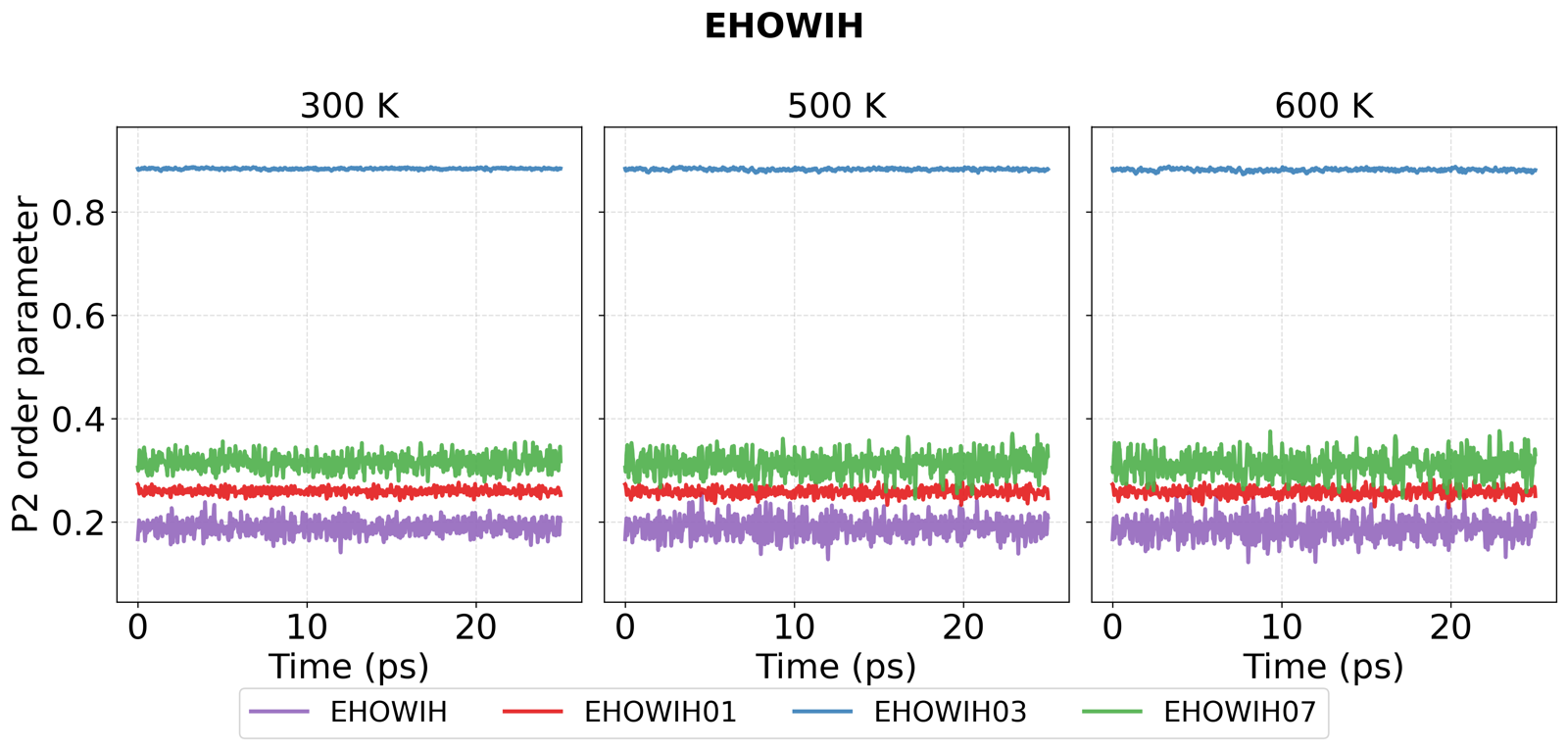}
    \caption{P2 orientational order parameter for all EHOWIH polymorphs at 300, 500, and 600~K.}
    \label{fig:p2_ehowih}
\end{figure}
\begin{figure}[H]
    \centering
    \includegraphics[width=1.0\linewidth]{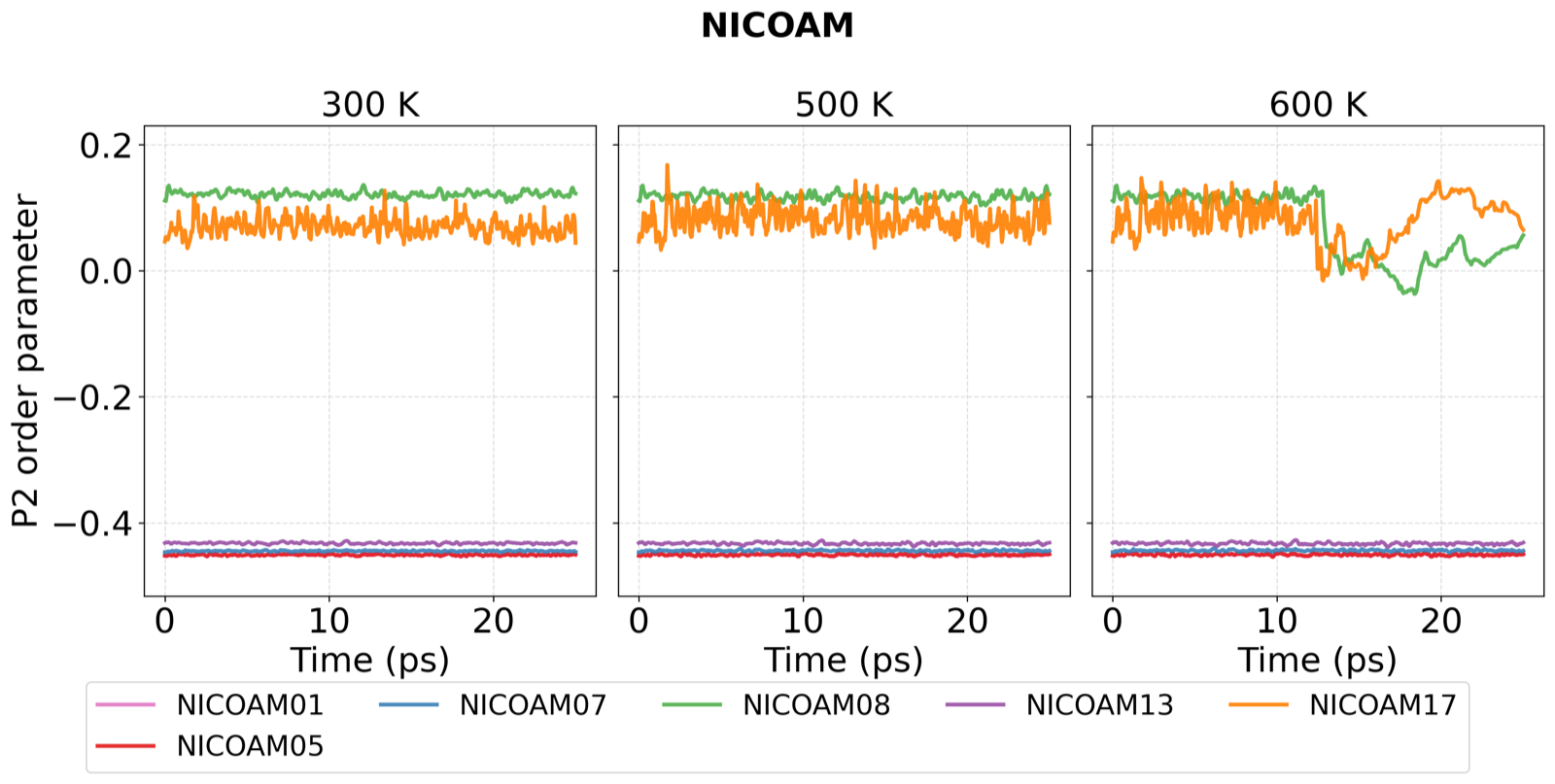}
    \caption{P2 orientational order parameter for all NICOAM polymorphs at 300, 500, and 600~K.}
    \label{fig:p2_nicoam}
\end{figure}

\begin{figure}[H]
    \centering
    \includegraphics[width=1.0\linewidth]{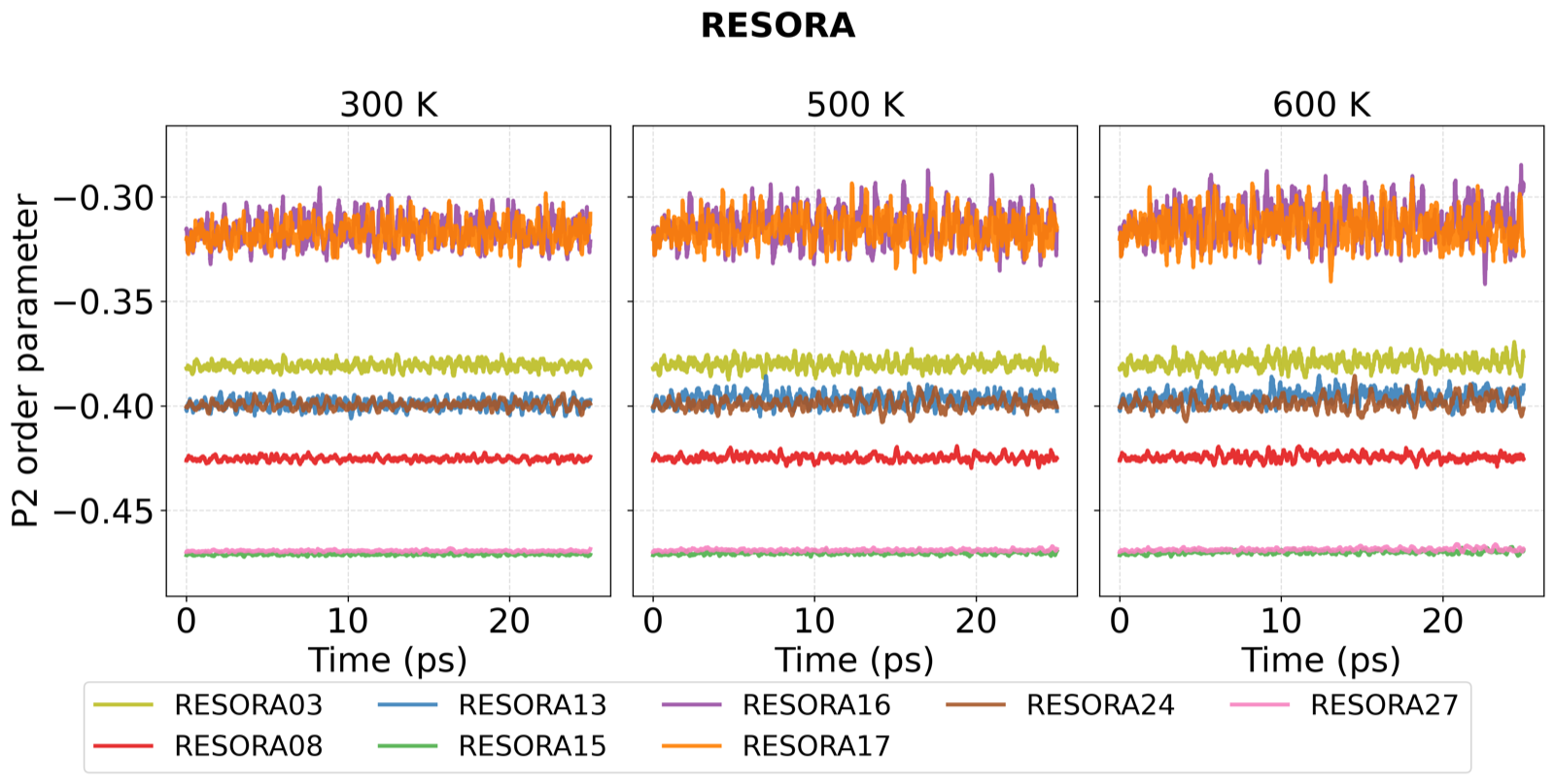}
    \caption{P2 orientational order parameter for all RESORA polymorphs at 300, 500, and 600~K.}
    \label{fig:p2_resora}
\end{figure}

\subsection{Radial Distribution Functions}
\begin{figure}[H]
    \centering
    \includegraphics[width=1.0\linewidth]{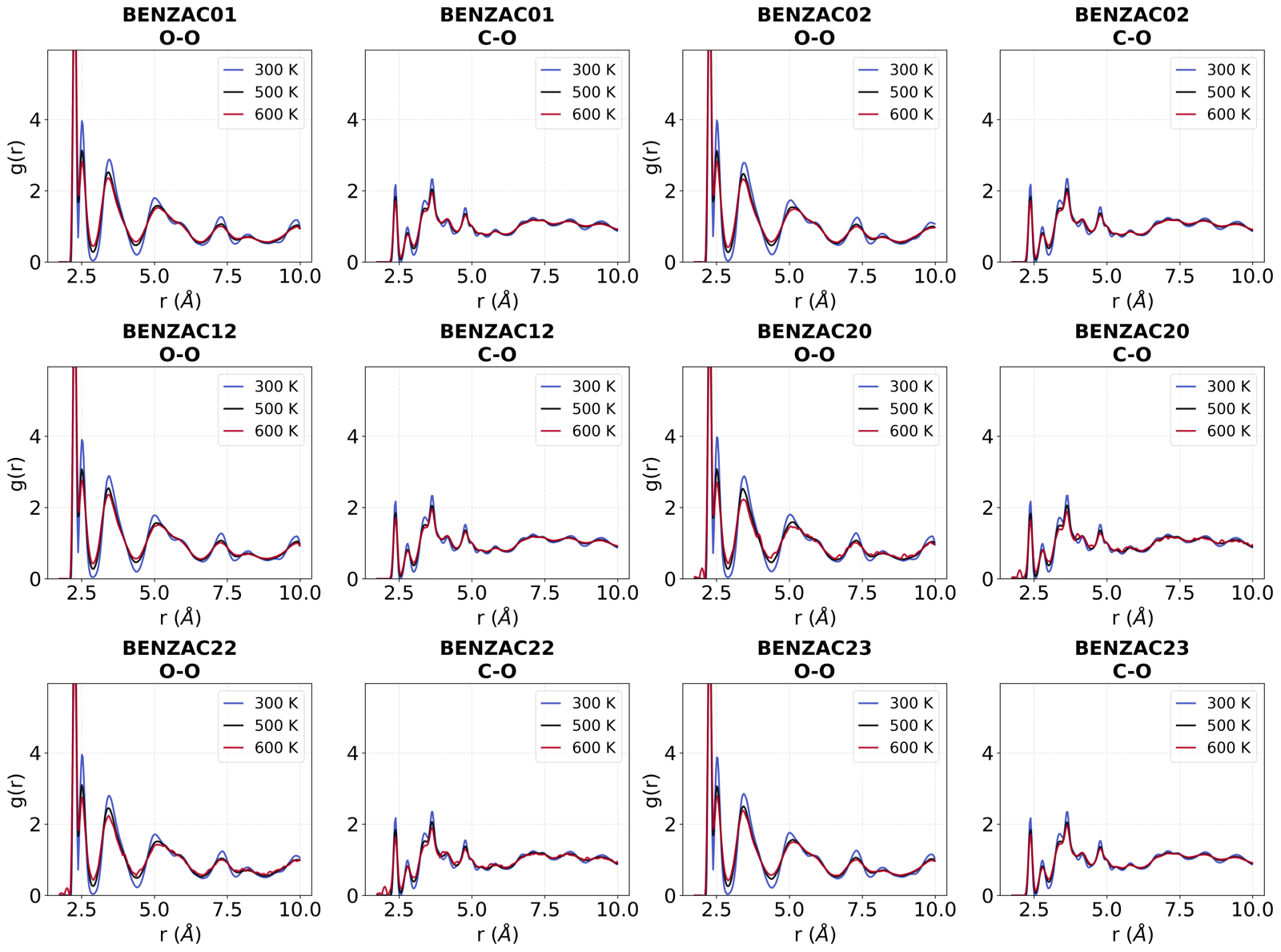}
    \caption{Radial distribution functions for all BENZAC polymorphs at 300, 500, and 600~K. Left panels show the intramolecular O--O pair; right panels show the intermolecular C--O pair.}
    \label{fig:rdf_benzac}
\end{figure}
\begin{figure}[H]
    \centering
    \includegraphics[width=1.0\linewidth]{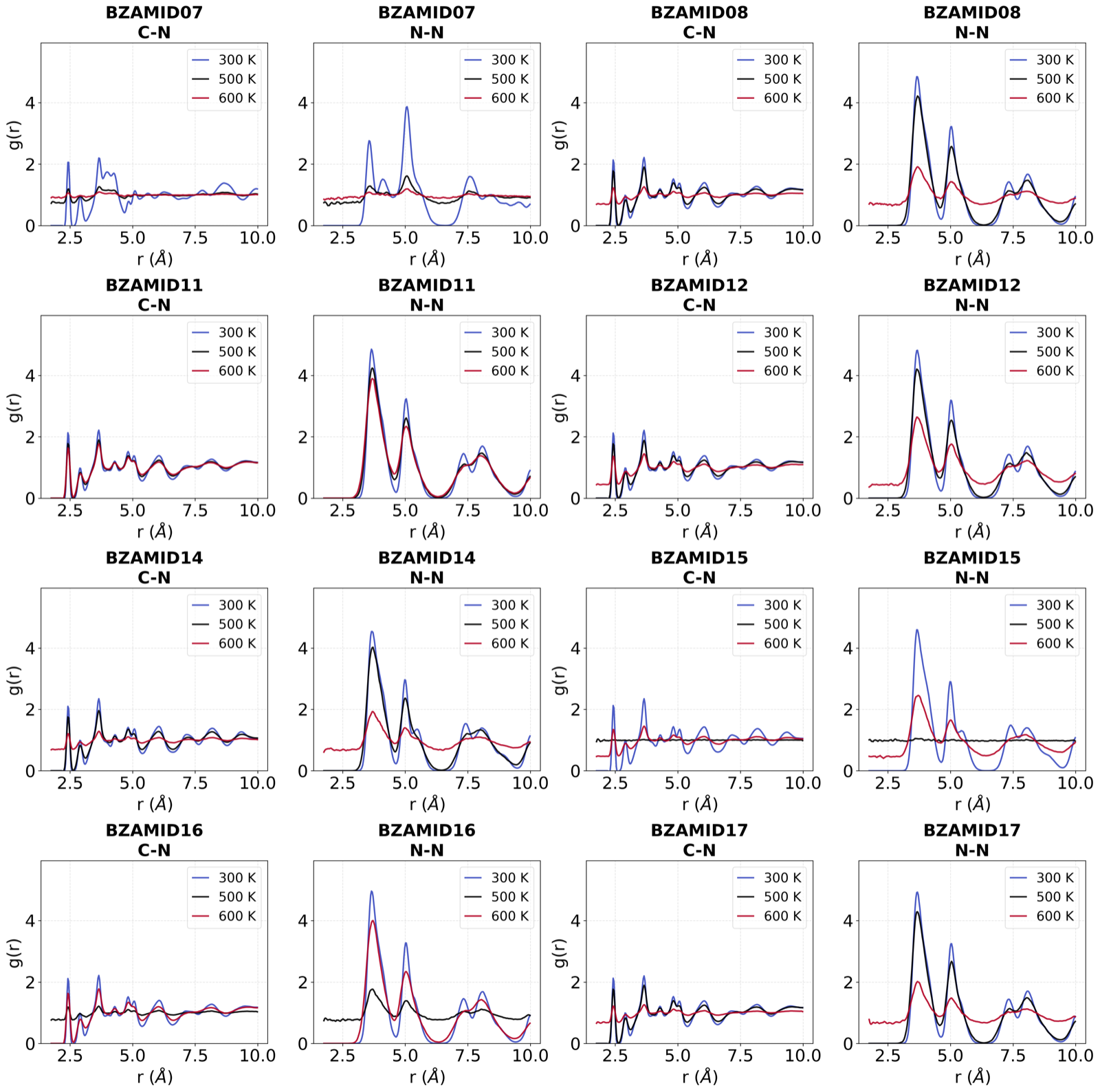}
    \caption{Radial distribution functions for BZAMID polymorphs (part 1) at 300, 500, and 600~K.}
    \label{fig:rdf_bzamid_1}
\end{figure}
\begin{figure}[H]
    \centering
    \includegraphics[width=0.5\linewidth]{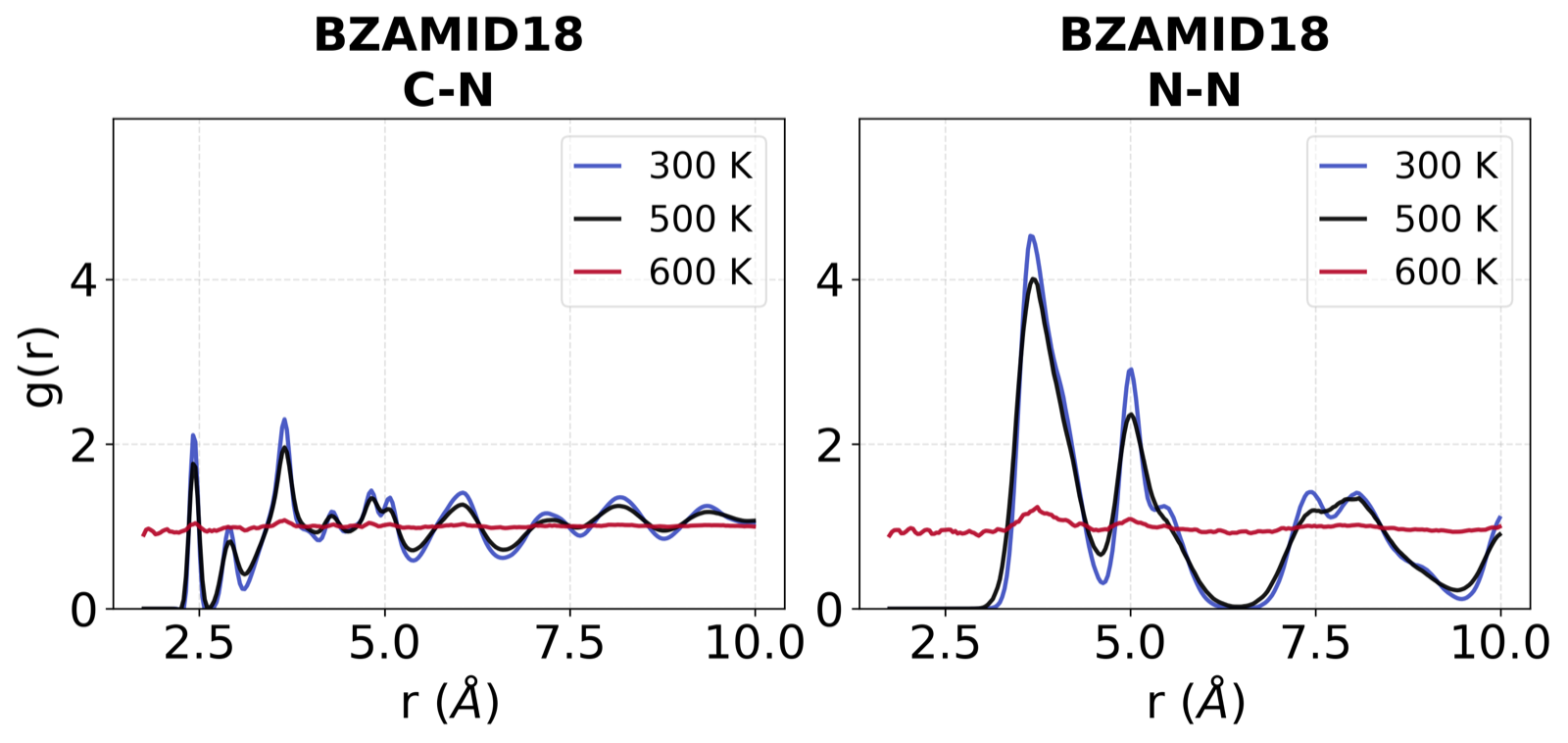}
    \caption{Radial distribution functions for BZAMID polymorphs (part 2) at 300, 500, and 600~K.}
    \label{fig:rdf_bzamid_2}
\end{figure}
\begin{figure}[H]
    \centering
    \includegraphics[width=1.0\linewidth]{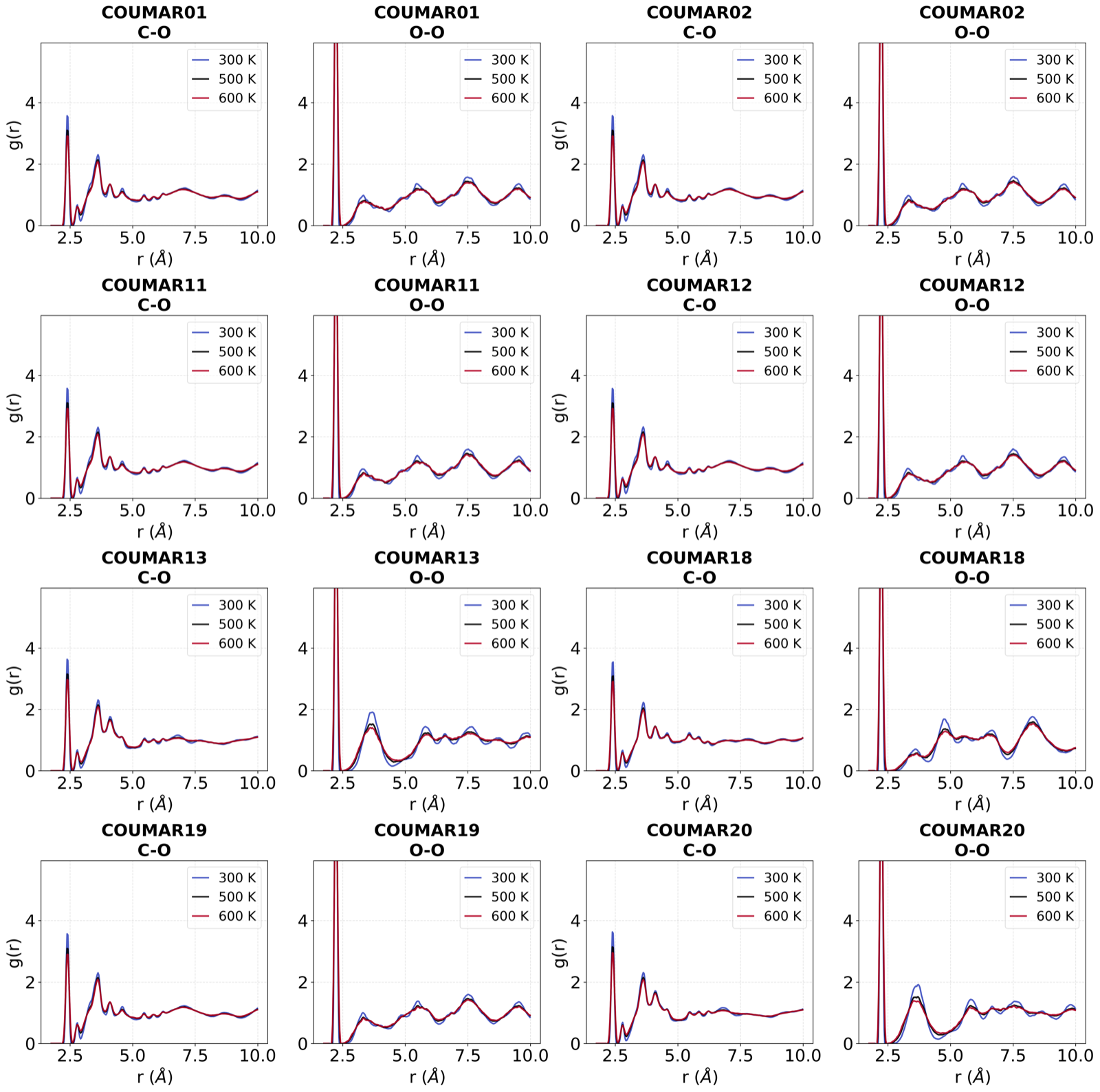}
    \caption{Radial distribution functions for COUMAR polymorphs (part 1) at 300, 500, and 600~K.}
    \label{fig:rdf_coumar_1}
\end{figure}
\begin{figure}[H]
    \centering
    \includegraphics[width=1.0\linewidth]{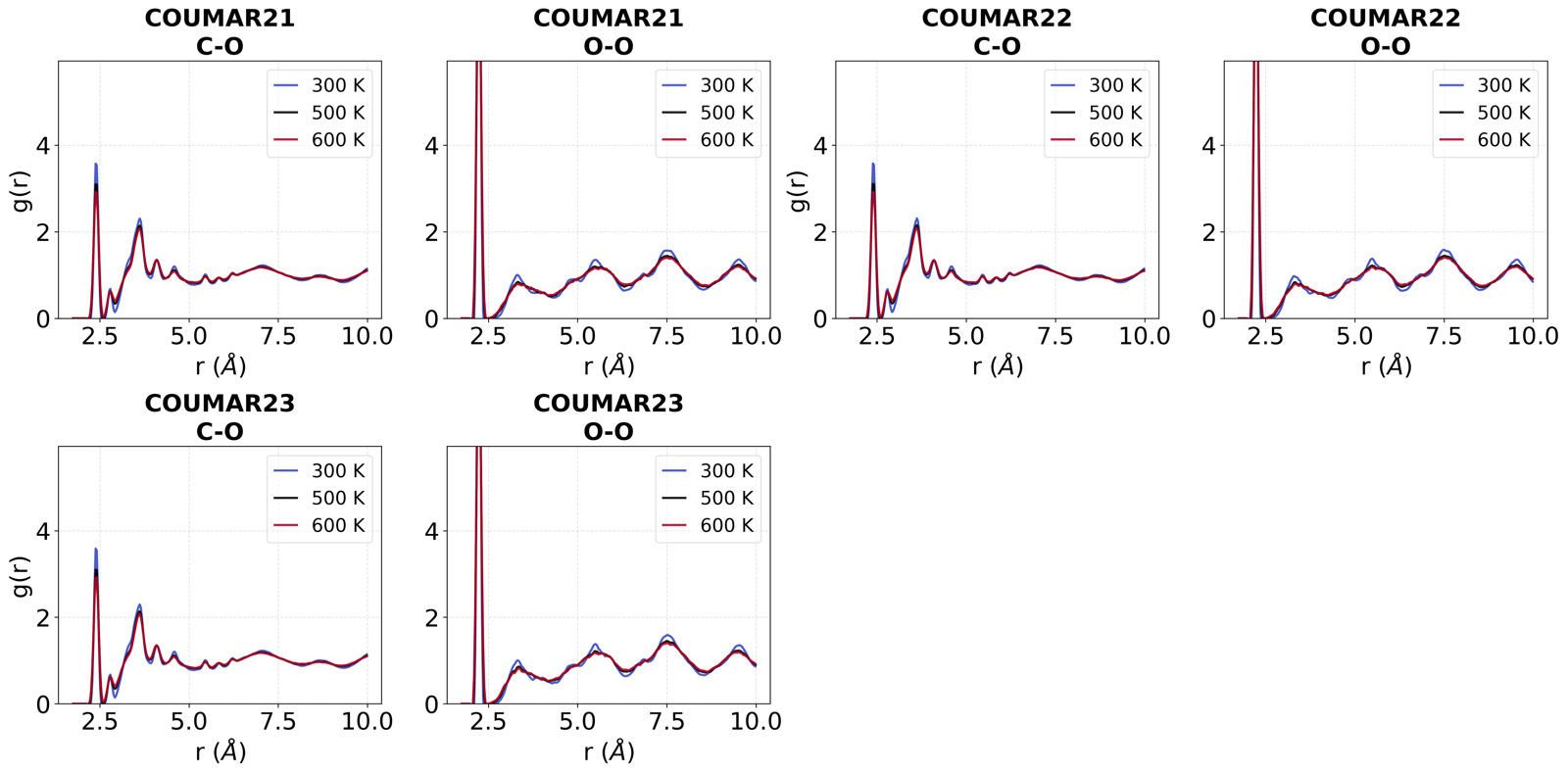}
    \caption{Radial distribution functions for COUMAR polymorphs (part 2) at 300, 500, and 600~K.}
    \label{fig:rdf_coumar_2}
\end{figure}
\begin{figure}[H]
    \centering
    \includegraphics[width=1.0\linewidth]{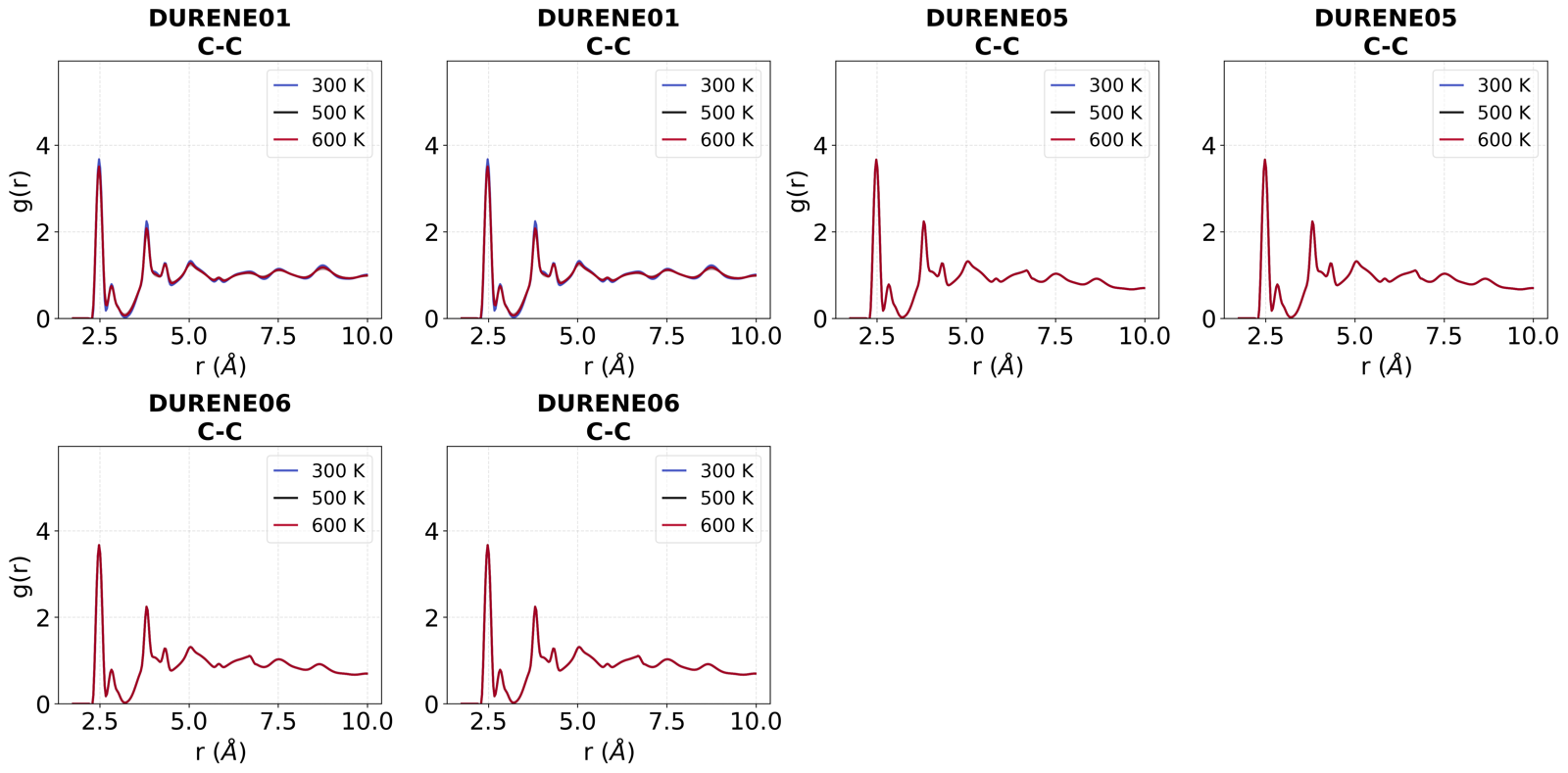}
    \caption{Radial distribution functions for all DURENE polymorphs at 300, 500, and 600~K.}
    \label{fig:rdf_durene}
\end{figure}
\begin{figure}[H]
    \centering
    \includegraphics[width=1.0\linewidth]{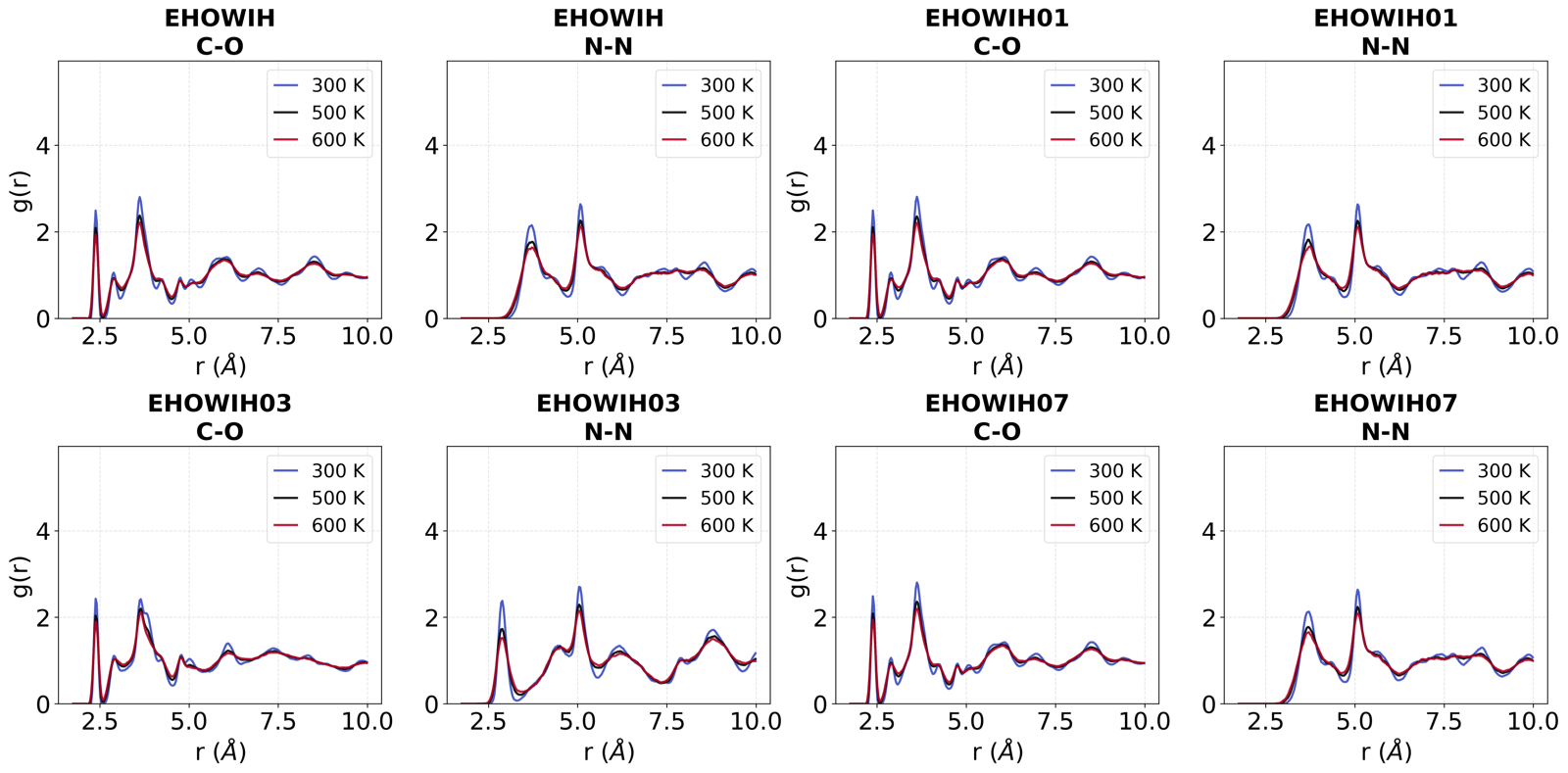}
    \caption{Radial distribution functions for all EHOWIH polymorphs at 300, 500, and 600~K.}
    \label{fig:rdf_ehowih}
\end{figure}
\begin{figure}[H]
    \centering
    \includegraphics[width=0.5\linewidth]{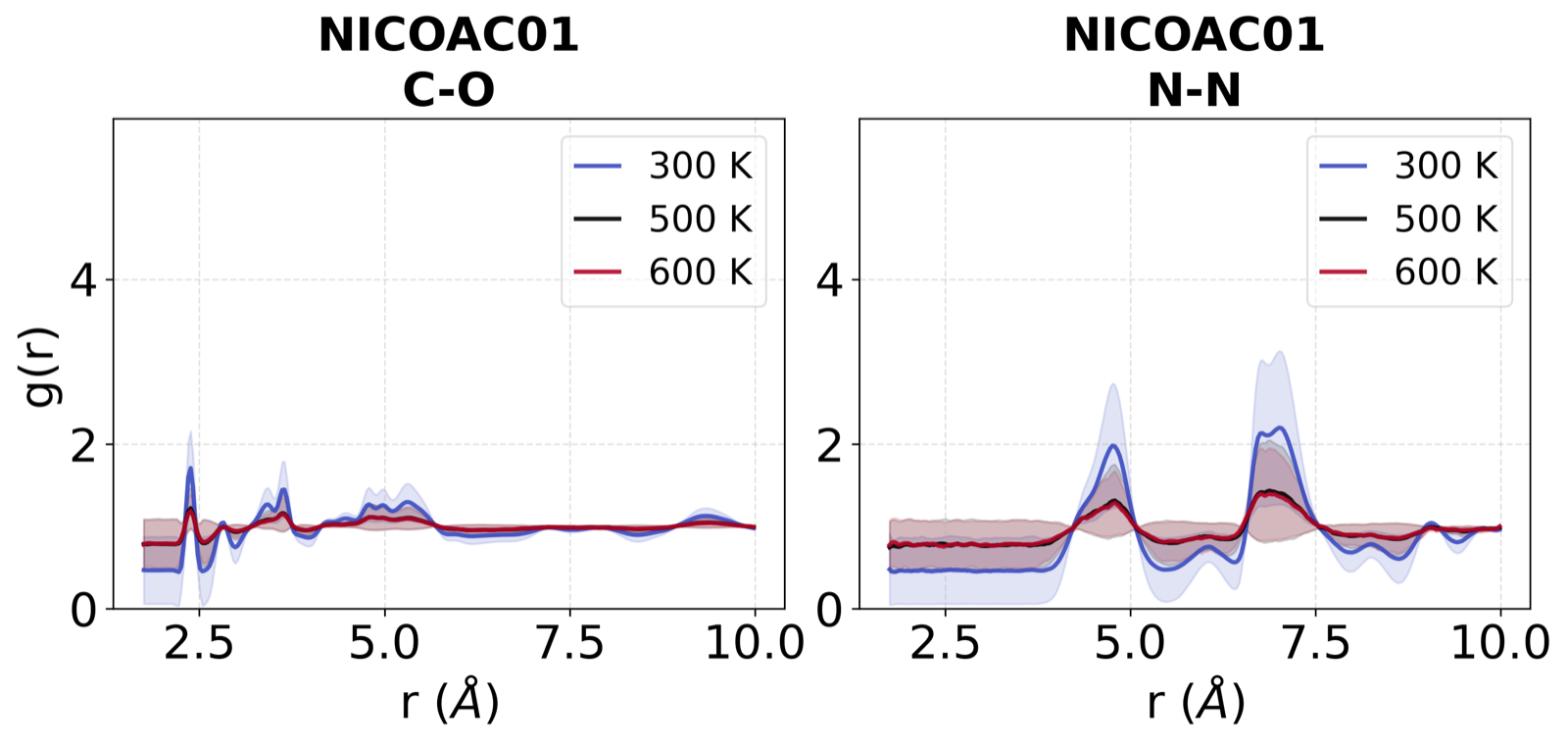}
    \caption{Radial distribution functions for all NICOAC polymorphs at 300, 500, and 600~K.}
    \label{fig:rdf_nicoac}
\end{figure}
\begin{figure}[H]
    \centering
    \includegraphics[width=1.0\linewidth]{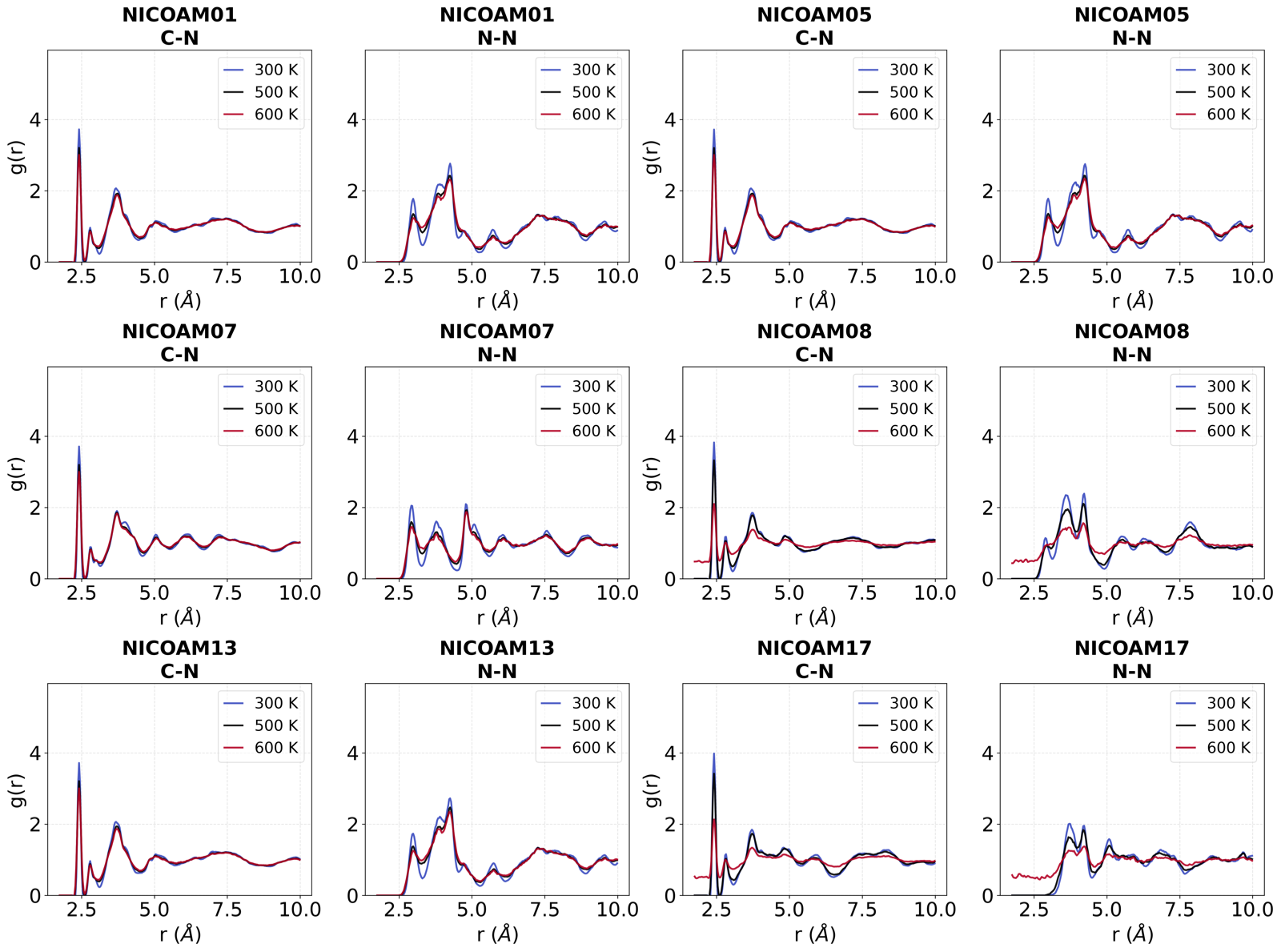}
    \caption{Radial distribution functions for all NICOAM polymorphs at 300, 500, and 600~K.}
    \label{fig:rdf_nicoam}
\end{figure}
\begin{figure}[H]
    \centering
    \includegraphics[width=1.0\linewidth]{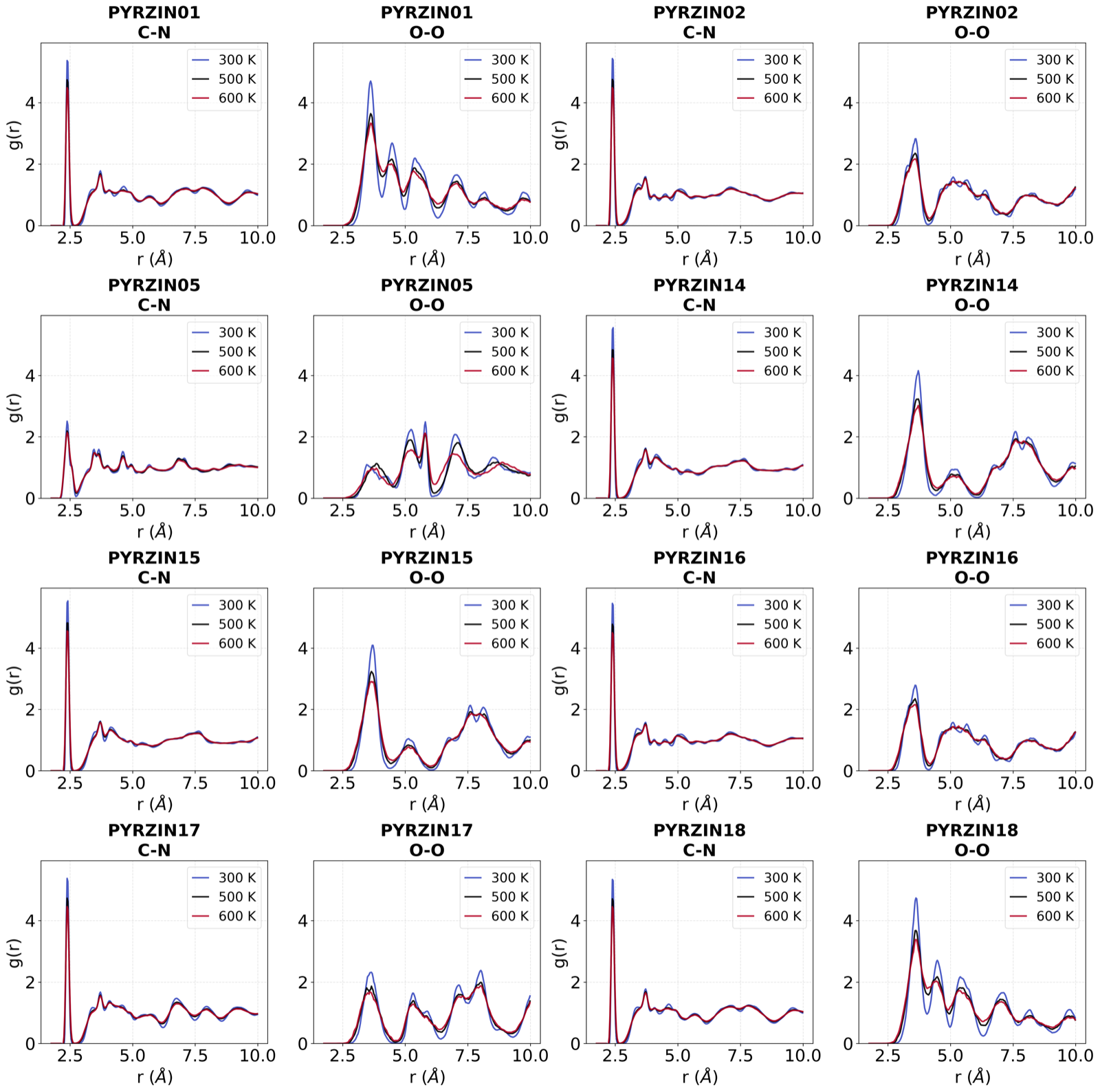}
    \caption{Radial distribution functions for PYRZIN polymorphs (part 1) at 300, 500, and 600~K.}
    \label{fig:rdf_pyrzin_1}
\end{figure}
\begin{figure}[H]
    \centering
    \includegraphics[width=1.0\linewidth]{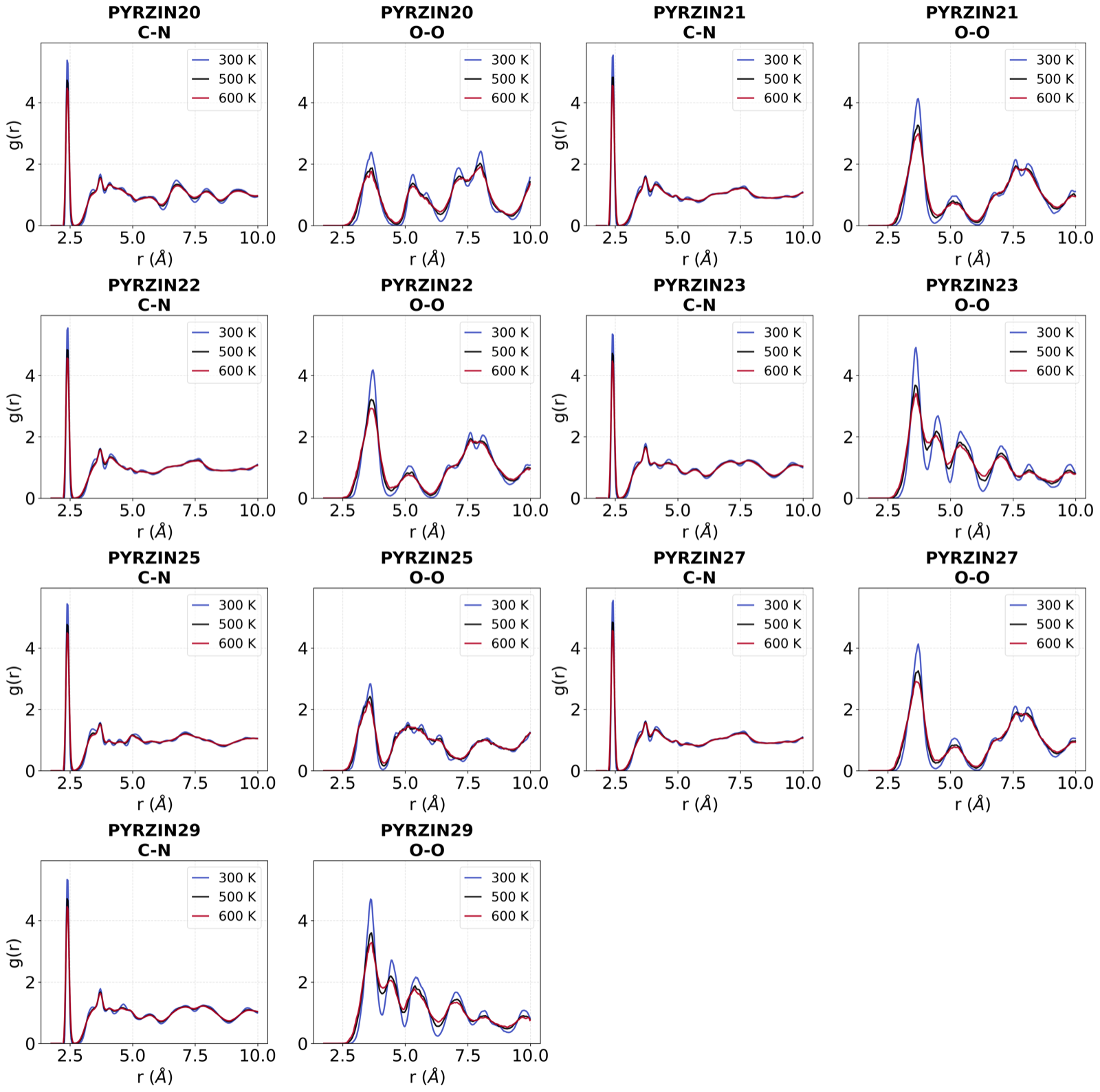}
    \caption{Radial distribution functions for PYRZIN polymorphs (part 2) at 300, 500, and 600~K.}
    \label{fig:rdf_pyrzin_2}
\end{figure}
\begin{figure}[H]
    \centering
    \includegraphics[width=1.0\linewidth]{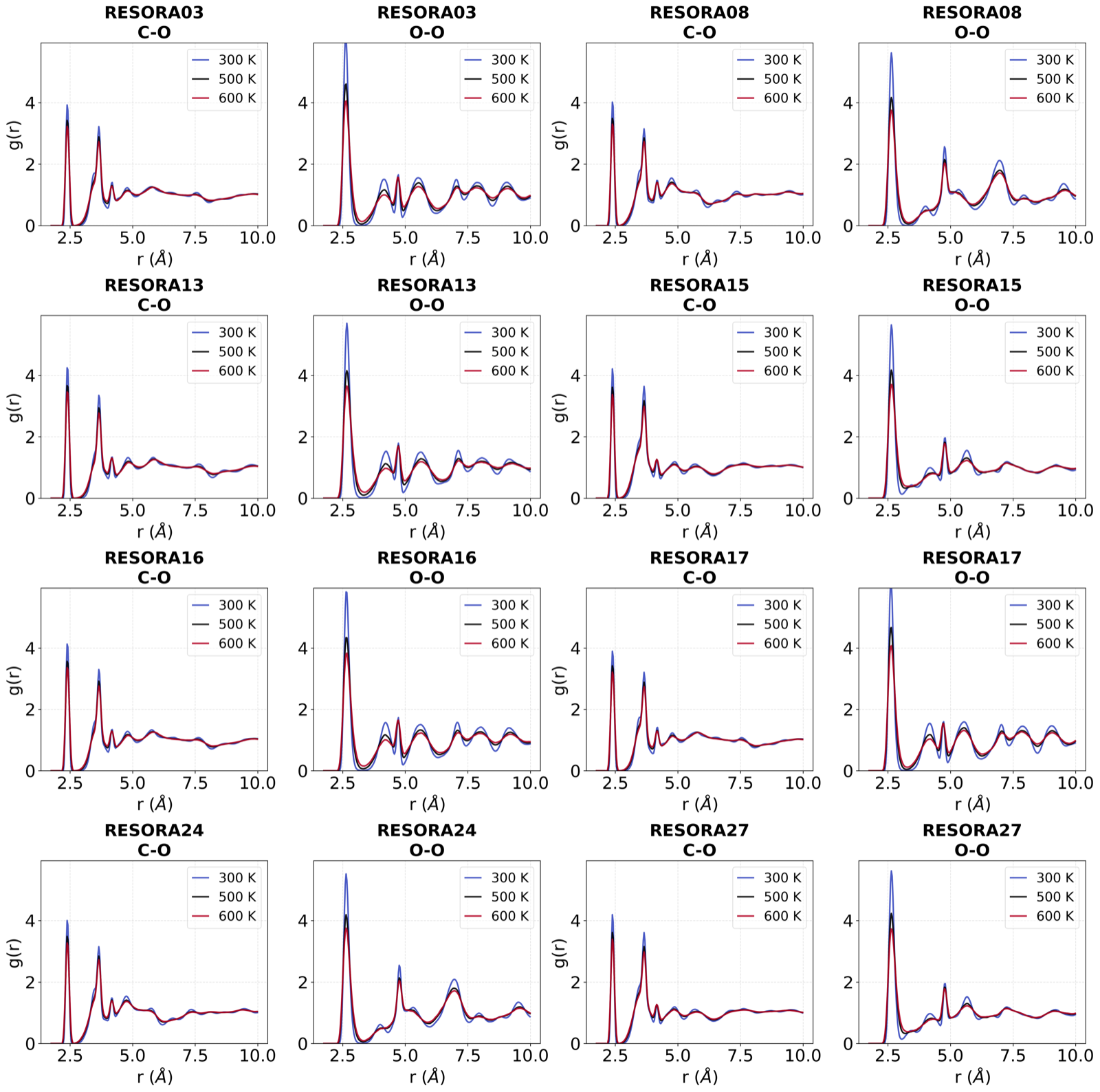}
    \caption{Radial distribution functions for all RESORA polymorphs at 300, 500, and 600~K.}
    \label{fig:rdf_resora}
\end{figure}

\end{document}